\documentclass{article} 
\usepackage{iclr2026_conference,times}

\usepackage{amsmath,amsfonts,bm}

\def\eqref#1{equation~\ref{#1}}

\def\1{\bm{1}}

\DeclareMathAlphabet{\mathsfit}{\encodingdefault}{\sfdefault}{m}{sl}
\SetMathAlphabet{\mathsfit}{bold}{\encodingdefault}{\sfdefault}{bx}{n}

\usepackage[utf8]{inputenc} 
\usepackage[T1]{fontenc}    
\usepackage{hyperref}       
\usepackage{url}            
\usepackage{booktabs}       
\usepackage{amsfonts}       
\usepackage{nicefrac}       
\usepackage{microtype}      
\usepackage[table,xcdraw]{xcolor}
\usepackage{amsmath}

\usepackage{algorithm}
\usepackage{algorithmicx}
\usepackage{algpseudocode}
\usepackage{xspace}
\usepackage{mathrsfs}
\usepackage{multirow}
\usepackage{enumitem}
\usepackage{dutchcal}
\usepackage{graphicx} 
\usepackage{subcaption}
\usepackage{pifont}
\usepackage{svg}
\usepackage{threeparttable}
\usepackage{color}
\usepackage{diagbox}
\usepackage{balance} 
\usepackage{amssymb}
\usepackage{bm}
\usepackage{caption}
\usepackage{bbding}
\usepackage{makecell}
\usepackage[most]{tcolorbox}
\usepackage{listings}
\usepackage{tabularray}

\usepackage{tocloft}

\usepackage{wrapfig}

\newtheorem{proposition}{Proposition}
\usepackage{framed}

\usepackage{rotating}

\definecolor{myblue}{RGB}{220,235,250}

\title{LLM as an Algorithmist: Enhancing Anomaly Detectors via Programmatic Synthesis
}

\author{Hangting Ye\textsuperscript{\normalfont 1 2}, Jinmeng Li\textsuperscript{\normalfont 1}, He Zhao\textsuperscript{\normalfont 3 4}, Mingchen Zhuge\textsuperscript{\normalfont 5}, Dandan Guo\textsuperscript{\normalfont 1 5}\footnotemark[1]\ \ , \\ \textbf{Yi Chang\textsuperscript{\normalfont 1 6 7}\thanks{Corresponding authors.}\ \ , Hongyuan Zha\textsuperscript{\normalfont 2}}\\
School of Artificial Intelligence, Jilin University\textsuperscript{\normalfont 1}; CUHK-Shenzhen\textsuperscript{\normalfont 2}\\
CSIRO’s Data61\textsuperscript{\normalfont 3}; Department of DSAI, Monash University\textsuperscript{\normalfont 4};
KAUST\textsuperscript{\normalfont 5}\\
International Center of Future Science, Jilin University\textsuperscript{\normalfont 6}\\
Engineering Research Center of Knowledge-Driven Human-Machine Intelligence, MOE, China\textsuperscript{\normalfont 7}
\\
\texttt{\{yeht22,lijm9921\}@mails.jlu.edu.cn, }\\
\texttt{he.zhao@data61.csiro.au, mingchen.zhuge@kaust.edu.sa, } \\
\texttt{\{guodandan,yichang\}@jlu.edu.cn, zhahy@cuhk.edu.cn}
}

\tcbset{
  promptstyle/.style={
    colframe=gray, 
    colback=white, 
    title={An example of the prompt template}, 
    fonttitle=\bfseries, 
    coltitle=black, 
    boxrule=0.5pt, 
    arc=4pt, 
    enlarge top by=3mm,
    enlarge bottom by=3mm,
  }
}

\iclrfinalcopy 
\begin{document}

\maketitle

\begin{abstract}

Existing anomaly detection (AD) methods for tabular data usually rely on some assumptions about anomaly patterns, leading to inconsistent performance in real-world scenarios. While Large Language Models (LLMs) show remarkable reasoning capabilities, their direct application to tabular AD is impeded by fundamental challenges, including difficulties in processing heterogeneous data and significant privacy risks. To address these limitations, we propose LLM-DAS, a novel framework that repositions the LLM from a ``data processor'' to an ``algorithmist''. Instead of being exposed to raw data, our framework leverages the LLM's ability to reason about algorithms. It analyzes a high-level description of a given detector to understand its intrinsic weaknesses and then generates detector-specific, data-agnostic Python code to synthesize ``hard-to-detect'' anomalies that exploit these vulnerabilities. This generated synthesis program, which is reusable across diverse datasets, is then instantiated to augment training data, systematically enhancing the detector's robustness by transforming the problem into a more discriminative two-class classification task. Extensive experiments on 36 TAD benchmarks show that LLM-DAS consistently boosts the performance of mainstream detectors. By bridging LLM reasoning with classic AD algorithms via programmatic synthesis, LLM-DAS offers a scalable, effective, and privacy-preserving approach to patching the logical blind spots of existing detectors. The source code is available at \url{https://github.com/HangtingYe/LLM_DAS#}.

\end{abstract}

\section{Introduction}

Anomaly detection (AD) is a fundamental machine learning task that identifies instances significantly deviating from normal data. 
Tabular data, typically represented as heterogeneous feature vectors (e.g. numerical and categorical features)~\citep{gorishniy2021revisiting, borisov2022deep, ye2024ptarl}, is a crucial modality for AD~\citep{han2022adbench, yin2024mcm}, with applications spanning cyber-security~\citep{ahmad2021network}, rare disease diagnosis~\citep{fernando2021deep, ye2023web}, and financial fraud detection~\citep{al2021financial}.
However, labeled anomalies are scarce due to the costly and time-consuming annotation by domain experts~\citep{chandola2009anomaly, ye2023uadb,10179265}.
In tabular anomaly detection, many recent works adopt a one-class classification paradigm, 
where only normal samples are available for training~\citep{scholkopf1999support, sohn2021learning, yin2024mcm, ye2025drl}. These TAD methods usually model how anomalies differ from normal data~\citep{ahmed2016survey} by making specific assumptions, e.g., reconstruction-based approaches like Principal Component Analysis (PCA)~\citep{shyu2003novel} assume anomalies are harder to reconstruct. 
Despite the success of these methods, it has been shown that assumptions made by a method may not hold when facing to heterogeneous tabular data in real-world scenarios~\citep{wolpert1997no, han2022adbench}.

\begin{figure}[t]
    \centering
    \includegraphics[width=0.6\linewidth]{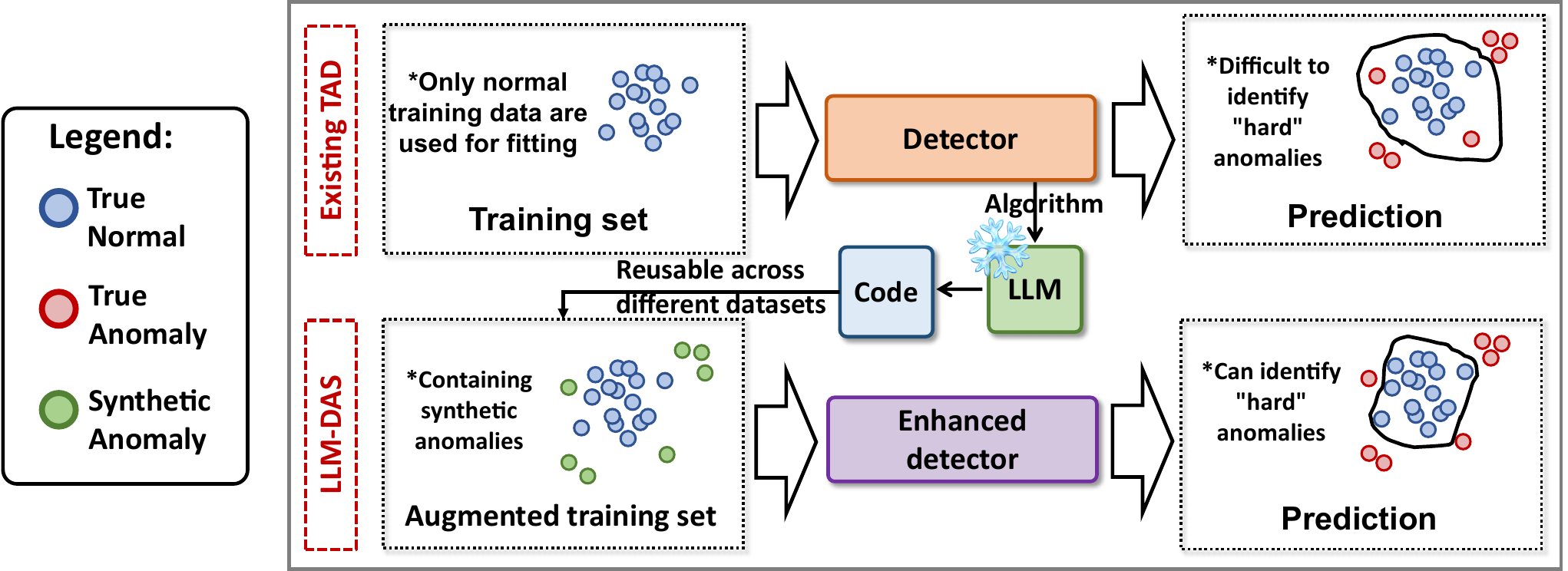}
    \captionsetup{font=small}
    \captionsetup{skip=0pt}
    \caption{Comparison between traditional TAD methods and our LLM-DAS. By synthesizing ``hard'' anomalies, LLM-DAS effectively transforms the original one-class classification problem into a more discriminative two-class classification problem, thereby strengthening the detector and yielding a more nuanced decision boundary.}
    \label{fig:introduction}
    \vspace{-1em}
\end{figure}

Rather than designing yet another detector with its own new set of fragile assumptions, we ask a more fundamental question: can we systematically enhance existing detectors to be more robust against the violation of their own core logic?
Our core idea is to strengthen detectors by exposing them to their ``blind spots'' through synthesized ``hard'' anomalies. 
Here, ``hard'' anomalies are synthetic anomalous samples that the detector struggles to tell apart from normal data., i.e., cases where the detector is most likely to fail.
As illustrated in Fig.~\ref{fig:introduction}, this targeted augmentation effectively transforms the learning problem from a limited one-class setting into a more powerful and discriminative two-class setting, allowing a model to learn a more nuanced and robust decision boundary. The pivotal challenge, however, lies in generating anomalies of sufficient quality to be truly ``hard''.
This requires targeting the detector's core algorithmic logic, demanding a paradigm shift from data-level optimization to a logic-level strategic attack.
This is a task for which the unique algorithmic reasoning capabilities of Large Language Models (LLMs) are exceptionally well-suited.
With their remarkable capabilities in language understanding~\citep{brown2020language}, complex planning~\citep{qin2023toolllm}, and code generation~\citep{zheng2024opencodeinterpreter}, LLMs possess an unparalleled ability to reason about abstract algorithmic mechanisms.

In this paper, we introduce LLM-DAS (\underline{LLM}-Guided \underline{D}etector-Aware \underline{A}nomaly \underline{S}ynthesis), a framework that positions the LLM as an ``algorithm strategist'' rather than a ``data processor''. 
In the first stage, LLM-DAS utilizes an LLM's reasoning capabilities to analyze the underlying mechanism of a given detector and generate detector-specific, data-agnostic Python code for anomaly synthesis. 
As no data is input into the LLM, this process elegantly sidesteps challenges related to tabular data processing and privacy concerns.
In the second stage, the generated code is instantiated on a specific dataset to synthesize ``hard-to-detect'' anomalies.
These anomalies are then used to augment the training data, enhancing the detector's robustness against violated assumptions.
Crucially, the detector-specific synthesis code is reusable across different TAD datasets, enabling both scalability and high effectiveness of LLM utilization. In brief, our contributions are summarized as follows: 
\textbf{1)} We identify the fragile assumptions of existing TAD methods as a core limitation and propose a novel direction of using LLM-generated code for targeted anomaly synthesis to enhance detector robustness. This approach simultaneously addresses inherent issues related to data privacy and processing for LLMs in the tabular domain.
\textbf{2)} We propose the LLM-DAS framework, which leverages an LLM to reason about detectors and generate reusable, data-agnostic synthesis code. This code is then instantiated to create synthetic hard anomalies that enhance detection performance.
\textbf{3)} We conduct extensive experiments on numerous benchmarks, where results demonstrate that LLM-DAS consistently and significantly boosts the performance of mainstream TAD methods.

\section{Related Work}

\textbf{Tabular Anomaly Detection.}
One of the main settings for tabular anomaly detection is one-class classification, where a model learns from only normal samples~\citep{sohn2021learning, yin2024mcm}. The success of existing methods, however, relies on specific and often fragile assumptions. \textit{(1) Reconstruction-based methods} like PCA~\citep{shyu2003novel} and autoencoders~\citep{principi2017acoustic, kim2019rapp} assume anomalies yield high reconstruction error~\citep{shyu2003novel, kim2019rapp, ye2025drl}. This premise fails for anomalies that lie deceptively close to the learned manifold of normal data. \textit{(2) Density estimation methods} posit that anomalies occupy low-density regions~\citep{parzen1962estimation, li2022ecod}. This makes them vulnerable to local anomalies hidden within otherwise dense data clusters. \textit{(3) Classification-based approaches} such as OCSVM learn a boundary around normal data~\citep{scholkopf1999support, tax2004support}, an assumption that is easily violated by complex, non-convex data distributions. \textit{(4) Isolation-based methods} like IForest assume anomalies are easily isolated~\citep{liu2008isolation, cao2025anomaly}. Their effectiveness diminishes when anomalies form small clusters or are adjacent to normal ones. In these cases, the isolation paths of anomalies are no longer much shorter than those of normal data, compromising the core principle of isolation. The inherent brittleness of these varied assumptions motivates our work to directly enhance detectors against the violation of their own core logic. For a comprehensive overview, we refer readers to several surveys~\citep{pang2021deep, ruff2021unifying, chandola2009anomaly}.

\textbf{Large Language Models for Tabular Data Learning.}
Recent advances in LLMs have sparked growing interest in their application to tabular data tasks, including prediction~\citep{dinh2022lift, hegselmann2023tabllm}, synthesis~\citep{borisov2022language, zhang2023generative}, and feature engineering~\citep{han2024large, nam2024optimized}. A comprehensive overview can be found in the survey by~\citet{fang2024large}. However, these approaches mostly treat the LLM as a \textit{data processor}, requiring direct access to raw feature values and treating each row as natural language descriptions. This paradigm faces fundamental challenges: LLMs inherently struggle with heterogeneous numerical features~\citep{fang2024large,yan2024making}, and direct data exposure creates significant privacy risks. Even recent attempts~\citep{tsai2025anollm} to apply LLMs for tabular anomaly detection---by fine-tuning them on normal samples---are still bound by these data-access and cost limitations, hindering their practicality. In this work, we depart from this paradigm by proposing a new role for the LLM: an \textit{algorithmist}. Instead of processing data, the LLM reasons about the high-level, data-agnostic logic of a given anomaly detector. Its task is to understand the detector's intrinsic weaknesses and programmatically generate a synthesis strategy in the form of executable code. This ``logic-level'' approach inherently preserves data privacy, leverages the LLM's core strengths in reasoning and code generation, and produces a reusable synthesis logic that is both detector-aware and data-agnostic.

\section{Method}

\subsection{Problem Formulation and Motivation}

Following previous works~\citep{yin2024mcm, ye2025drl}, we formulate tabular anomaly detection (TAD) in a one-class classification setting, where training set $D_{\text{train}}$ only contains normal samples with $\mathbf{x} \in \mathcal{X}$ and test set $D_{\text{test}}$ contains both normal and anomalous samples. The goal of TAD is to learn a score function $f: \mathcal{X} \to \mathbb{R}$ from $D_{\text{train}}$, which assigns an anomaly score $f(\mathbf{x})$ to a given test sample, with a higher score indicating a higher likelihood of being an anomaly. A variety of methods have been developed for this task, such as reconstruction-based (e.g., PCA~\citep{shyu2003novel}), isolation-based (e.g., IForest~\citep{liu2008isolation}), and density-based (e.g., ECOD~\citep{li2022ecod}) approaches. To make our exposition precise, we denote $T=\{\text{``IForest'', ``PCA'', ...}\}$ as the set of algorithm names, where $t\in T$ is a symbolic label specifying the chosen algorithm type. Each $t$ corresponds to an abstract, parameter-free algorithmic description, $A_t$. Applying the algorithm $A_t$ to the training set $D_{\text{train}}$ yields a concrete score function $f_t$ with learned parameters. 

The efficacy of these classical TAD algorithms, however, hinges on their underlying assumptions about the nature of anomalies. For instance, IForest~\citep{liu2008isolation} assumes anomalies are more easily isolated, while PCA~\citep{shyu2003novel} assumes they have large reconstruction errors from a learned low-dimensional subspace. Details of these detectors are in Appendix~\ref{appendix:baseline details}. These assumptions are often brittle and fail when confronted with heterogeneous, real-world data, leading to degraded performance~\citep{wolpert1997no, han2022adbench}. It raises a critical question: how can we systematically enhance existing detectors to be more robust against the violation of their own core logic?

Our core approach is to strengthen detectors by exposing them to their own blind spots through the synthesis of ``hard'' anomalies---samples specifically crafted to challenge the algorithmic assumptions of a given detector. By augmenting the training data with these targeted negative examples, we {transform the learning problem from a limited one-class setting into a more discriminative two-class setting}. The success of this approach, however, hinges on generating anomalies of sufficient quality to be truly ``hard''. Naive, data-level methods, such as applying small perturbations to input features, often fail to target the core algorithmic logic of a detector. To overcome this, we introduce a paradigm shift from data-level optimization to a {logic-level strategic attack}, leveraging the unique ability of Large Language Models (LLMs) to reason about abstract algorithmic mechanisms. The LLM's role is not to process data, but to analyze a detector's high-level logic, infer its weaknesses, and then {generate data-agnostic Python code} that programmatically synthesizes anomalies tailored to exploit these specific vulnerabilities.

\subsection{Proposed Method: LLM-DAS}
\begin{figure}[!t]
\centering
\includegraphics[width=\linewidth]{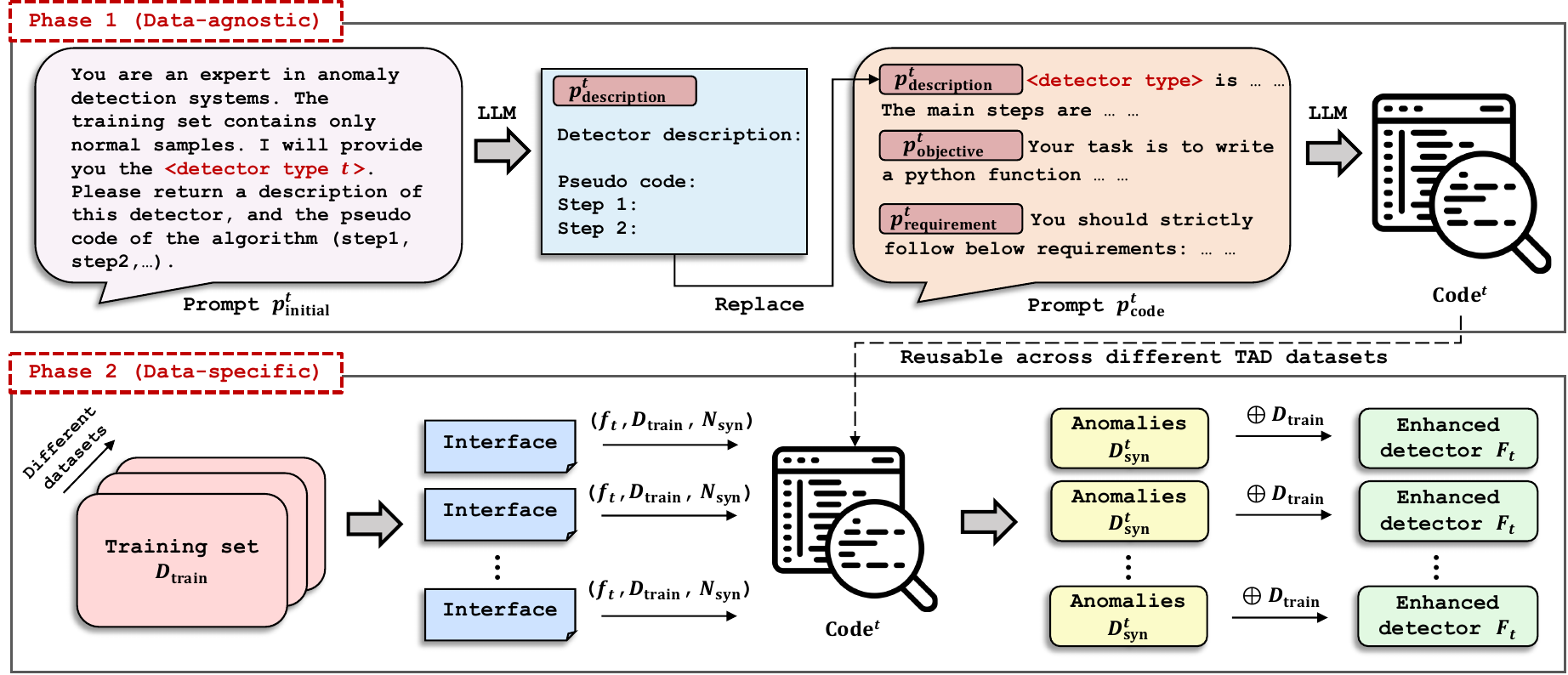}
\captionsetup{font=small}
\captionsetup{skip=0pt}
\caption{The LLM-DAS framework consists of two phases: (1) a data-agnostic reasoning phase, where an LLM generates a reusable anomaly synthesis code for one type of detector, and (2) a data-specific phase, where this code is applied to generate challenging anomalies for detector enhancement.} 
\label{fig:framework}
\vspace{-1em}
\end{figure}

This work proposes LLM-DAS (\underline{LLM}-Guided \underline{D}etector-Aware \underline{A}nomaly \underline{S}ynthesis), a two-stage framework. The core of LLM-DAS is to first leverage an LLM's reasoning to generate a data-agnostic synthesis code tailored for a detector algorithm $A_t$ (Stage 1), and then execute this code to create the ``hard'' anomalies needed to enhance $f_t$ on a specific dataset (Stage 2). Notably, the detector-specific synthesis code is reusable across different TAD datasets. Fig.~\ref{fig:framework} illustrates the overall pipeline.


\subsubsection{Detector-Aware Code Generation via LLM}
This stage aims to translate the LLM's abstract algorithmic knowledge regarding a detector type $t$ (represented by its algorithm $A_t$) into a concrete, executable synthesis code. As the LLM analyzes the detector not the data, this process is entirely data-agnostic to preserve privacy and ensure the reusability of the generated code. Specifically, given a detector type $t$, we construct a prompt $p_{\text{code}}^t$ that guides the LLM to capture the underlying assumptions and mechanisms of algorithm $A_t$, and then produce anomalies synthesis code that captures the kinds of anomalies that are inherently difficult for $A_t$ to detect. To fully exploit the reasoning capability of LLMs, the prompt $p_{\text{code}}^t$ is defined as:
 

\begin{equation}
\label{equ:prompt}
\begin{aligned}
    p_{\text{code}}^t = p_{\text{description}}^t + & \ p_{\text{objective}}^t + p_{\text{requirements}}^t.
\end{aligned}
\end{equation}
Below we introduce the three key components (see Appendix~\ref{appendix:prompt_IF} for details).  %

\textbf{(i)} \textbf{Detector description $p_{\text{description}}^t$},a summary and pseudo-code of the detector algorithm $A_t$. This information is generated by the LLM itself:

\begin{equation}
\label{equ:description}
\begin{aligned}
    p_{\text{description}}^t& = \text{LLM}(p_{\text{initial}}^t),
\end{aligned}
\end{equation}
where prompt $p_{\text{initial}}^t$ is the input for LLM with details provided in Appendix~\ref{appendix:prompt_IF}. This step provides a structured context to prime the LLM for generating detector-aware synthesis code. %


\textbf{(ii)} \textbf{Objective specification $p_{\text{objective}}^t$}, which instructs the LLM to generate code that synthesizes ``hard'' anomalies. A core tenet of our framework is the use of symbolic interfaces, which are defined in this prompt section. For example, we state: \textit{``... After the Python function is completed, users can provide the function with: A trained PCA model (\texttt{model}) that exposes \texttt{predict\_score()}, The training samples (\texttt{X\_train}).''} This interaction paradigm is crucial: by defining interfaces as symbolic placeholders rather than concrete values, the LLM can write a general-purpose program without ever accessing the real data. It effectively provides the LLM with a standard API to program against, positioning it as an ``algorithm strategist'' rather than a ``data processor.'' This enables the generated code to be both powerful--accessing training set statistics and querying the detector's scoring behavior--and universally applicable across different datasets.

\textbf{(iii)} \textbf{Generation requirements $p_{\text{requirements}}^t$}, which impose constraints on the code format and functionality. Notably, we encourage the code to identify ``borderline'' normal training samples lying near the decision boundary of the detector, and transform them into anomalies leveraging LLM reasoning.
This approach is effective for producing informative anomalous samples. Nevertheless, if the LLM discovers a superior strategy, it is allowed to adopt it.

The LLM takes the prompt $p_{\text{code}}^t$ as input and leverages its prior knowledge to understand the mechanism of algorithm $A_t$ (summarized as $p_{\text{description}}^t$ in Eq.~\ref{equ:description}). Based on this understanding, it reasons about which types of anomalies are most challenging for $A_t$ to detect while being most informative for improving detection performance. The output strictly follows  a Python code format:
\begin{equation}
\label{equ:code}
\begin{aligned}
    \text{Code}^t &= \text{LLM}(p_{\text{code}}^t), \\
       \{ \mathcal{S}_{\text{policy}}^t,  
       \mathcal{S}_{\text{program}}^t&,
       \mathcal{S}_{\text{explanation}}^t\} \subseteq \text{Code}^t,
\end{aligned}
\end{equation}
where the code consists of three structured components: the policy $\mathcal{S}_{\text{policy}}^t$ that specifies how anomalies should be synthesized in detail, the executable program $\mathcal{S}_{\text{program}}^t$ that implements the policy $\mathcal{S}_{\text{policy}}^t$ in Python, and the explanation $\mathcal{S}_{\text{explanation}}^t$ with clear comments explaining key steps in program $\mathcal{S}_{\text{program}}^t$.  This structured output, comprising not just a program but also its underlying policy and explanation, significantly enhances the interpretability and trustworthiness of our approach, allowing users to verify the logic behind the LLM's synthesis strategy. We provide an example template of the generated code in List.~\ref{lst:gen_anom}, with detailed generated code provided in Appendix~\ref{appendix:generated_code}.

\begin{minipage}{0.48\textwidth}
\scriptsize
\begin{tcolorbox}[
                  colback=gray!10,
                  colframe=gray!50!black,
                  title=The Generated Python Code Template,
                  fonttitle=\footnotesize\bfseries,
                  coltitle=white,
                  colbacktitle=black,
                  top=0mm, bottom=0.5mm, left=0.2mm, right=0.2mm,
                  arc=2mm,           
                  height=4.3cm,
                   boxrule=0.5pt,
                  label={lst:gen_anom}]
\begin{lstlisting}[language=Python, basicstyle=\tiny\ttfamily, 
    commentstyle=\itshape\color{green!50!black},backgroundcolor=\color{gray!10},
    aboveskip=0pt,     % 移除上方间距
    belowskip=0pt,     % 移除下方间距
    lineskip=0.5pt,    % 减小行距
    xleftmargin=2pt,   % 减小左边距
    xrightmargin=2pt]
def generate_hard_anomalies(n_samples: int, model, X_train: np.ndarray) -> np.ndarray:
    """
    # === Policy placeholder ===
    # The anomaly synthesis logic is applied here
    # (details omitted for brevity)
    """
    # === Program and explanation placeholder ===
    # (details omitted for brevity)
    # Explanation for step 1
    Step 1 xxx
    ...
    return np.array(generated_anomalies)
\end{lstlisting}
\end{tcolorbox}
\end{minipage}
\hfill
\begin{minipage}{0.48\textwidth}
Crucially, this LLM-driven code generation is data-agnostic. The LLM reasons about a detector's algorithmic weaknesses based solely on its description—never observing raw data or model parameters, thus preserving privacy. This process yields a reusable, detector-specific synthesis code without requiring any LLM fine-tuning. The resulting structured code is instantiated with dataset-specific values in the next stage to synthesize anomalies.
\end{minipage}

\subsubsection{Dataset-Wise Hard Anomaly Instantiation}
Once the detector-aware code is generated, the next step is to instantiate it on a specific dataset. To do so, the generated code, denoted as $\text{Code}^t$, requires a well-defined set of input interfaces to execute. We define this complete set of interfaces as $\mathcal{I}$, which comprises three core elements presented in their logical order: (1) Training set $D_{\text{train}}$, which serves as the foundation for the instantiation process. The code requires this data not only to train the score function in the next step but also to access its distribution and statistical properties (such as the decision boundary) during the synthesis phase to generate more challenging anomalies. (2) A fitted score function $f_t$: Using the training set $D_{\text{train}}$  described above, we obtain a fitted score function $f_t$ by applying algorithm $A_t$ to $D_{\text{train}}$. This score function provides the key standardized interface that maps any given input sample to an anomaly score. (3) The number of anomalies to synthesize $N_{\text{syn}}$, a hyperparameter specifying the desired number of anomaly samples to be generated. Collectively, these three components constitute the complete set of information required for the code's execution. The complete interface set $\mathcal{I}$ is therefore defined as their union:
\begin{equation}
\label{equ:interfaces_corrected}
    \mathcal{I}_t \;=\; D_{\text{train}}  \;\cup\; f_t \;\cup\; N_{\text{syn}}.
\end{equation}
 With this set of interfaces $\mathcal{I}$ available for a specific dataset, we can execute the LLM-generated code. The code operates as a function that accepts $\mathcal{I}$ as input and returns a set of synthesized anomalies,
 \begin{equation}
\label{equ:code_instantiation}
    \text{Code}^t: \ \mathcal{I}_t \;\mapsto\; D_{\text{syn}}^t, D_{\text{syn}}^t = \{\mathbf{x}_{\text{syn}}^t\}.
\end{equation}
This formulation makes explicit that the code consumes both the score function and the training set at instantiation time, while the LLM that produced the code itself remained data-agnostic during generation. The resulting $\text{Code}^t$, generated in a single forward pass for a specific detector type $t$, can be applied universally across diverse datasets.

\textbf{Case Study}. To make this instantiation process concrete, we now present a case study for the IForest detector~\citep{liu2008isolation}. The LLM-generated code, $\text{Code}^{\text{IForest} }$, operates based on a specific synthesis policy designed to exploit the core mechanics of IForest. This policy $\mathcal{S}_{\text{policy}}^{\text{IForest}}$, which the LLM articulates as part of its output (Eq.~\ref{equ:code}), is detailed below in Policy~\ref{lst:policy}.

\begin{tcolorbox}[
    colframe=black,       
    colback=gray!10,       
    coltitle=white,        
    colbacktitle=black,    
    title=Policy $\mathcal{S}_{\text{policy}}^t$ contained in $\text{Code}^t$ (Eq.~\ref{equ:code}) for Generating Hard Anomalies for IForest,    
    fonttitle=\footnotesize\bfseries,   
    fontupper=\scriptsize,
    arc=2mm,               
    boxrule=0.5pt,         
    top=1mm, bottom=1mm, left=2mm, right=2mm, 
    label={lst:policy},
    enhanced          
    ]
The \textcolor{blue}{\textbf{core weakness}} of IForest is that it relies on axis-aligned splits to isolate points.
A sample is considered anomalous if it can be isolated in a few splits (short path length).
Conversely, a sample is normal if it is "deep" within the data cloud, requiring many splits to be isolated (long path length).
Our strategy exploits this by creating new points that are conceptually anomalous but are geometrically located in a way that maximizes their path length, thereby fooling the IForest.
The policy consists of two main steps:

\textcolor{blue}{\textbf{1. Identify 'Borderline' Normal Samples:}}
These are the normal training samples
that lie on the fringe or edge of the dense data cloud. The IForest model is already
less certain about these points, assigning them the highest anomaly scores among all
normal samples. They serve as perfect "seeds" for our hard anomalies because they
are already close to the decision boundary. We identify these by finding the training
samples in the top percentile of anomaly scores.

\textcolor{blue}{\textbf{2. Transform Seeds into Hard Anomalies via 'Controlled Extrapolation':}}
We transform these borderline seeds into anomalies. A naive transformation (e.g., adding
large random noise) would create an obvious outlier with a short path length, which is an
*easy* anomaly. To create a *hard* anomaly, we must move the seed point in a way that
doesn't significantly shorten its path length.
\end{tcolorbox}

In practice, the implementation of this policy relies on the interfaces provided to the code. Specifically, to identify "Borderline" Normal Samples (Step 1), the code uses the fitted $f_t$ to compute anomaly scores for all samples in the training set. It then selects a subset of these samples as ``seeds'' from the top percentile of scores. For Step 2, the code applies targeted modifications to these seed points. The objective is to push them into a sparser region of the feature space, making them factually anomalous, while algorithmically ensuring the transformation does not significantly reduce their path length as computed by the IForest score function $f_t$. This process effectively creates anomalies that are ``camouflaged'' to appear normal to IForest, thus representing a significant challenge for the detector.

\textbf{Detector Enhancement with Hard Anomalies.}
Finally, the synthesized anomalies are integrated with the original training set:
\begin{equation}
\begin{aligned}
\label{equ:integration}
    D^t_{\text{aug}} = D_{\text{train}} \cup D_{\text{syn}}^t,
\end{aligned}
\end{equation}
and a binary classifier $\tilde{f}_t$ (e.g., a decision tree) is trained on $D^t_{\text{aug}}$ to distinguish between normal and anomalous samples. 
By explicitly framing the task as a supervised classification problem and exposing the classifier to these challenging synthetic anomalies, the learning process encourages the $\tilde{f}_t$ to capture more discriminative patterns between the two classes, thereby improving its ability to generalize and detect real-world anomalies.
We fuse the original ($f_t$) and enhancement ($\tilde{f}_{t}$) score functions by summing their min-max normalized scores with the following  final score:
\begin{equation}
\begin{aligned}
\label{equ:ensemble}
F_{t}(x) = \text{Norm}_{\text{min-max}}(f_{t}(x)) + \text{Norm}_{\text{min-max}}(\tilde{f}_{t}(x)). 
\end{aligned}
\end{equation}

This approach preserves the original detector's biases while adding the nuanced patterns learned from synthetic data.
A key aspect of our method is its adherence to the one-class classification framework since training set $D_{\text{train}}$ only have ground-truth normal data. We leverage LLM-generated code to synthesize anomalies tailored to the detector and the training dataset.
This approach allows the detector to adapt to edge cases that it would otherwise fail to recognize, effectively broadening its coverage of the abnormal space while preserving its ability to model the normal distribution (\textit{mitigating the challenge of fragile assumptions in TAD}). 
Overall, this straightforward yet principled strategy ensures that the benefits of LLM-guided anomaly synthesis are fully incorporated into the detector, enhancing both its discriminative power and generalization ability.

\subsubsection{Summary and Discussion}
The LLM-DAS framework operates in a two-phase process that decouples data-agnostic reasoning from data-specific synthesis. In Phase 1, an LLM analyzes a high-level description of a detector and generates a universal, reusable Python code that encapsulates a strategy for synthesizing hard anomalies. Crucially, this phase is data-agnostic, preserving privacy. In Phase 2, this code is instantiated on a specific training set $D_{\text{train}}$ and its fitted detector $f_t$ to generate challenging anomalies. These anomalies are then used to train an enhancement classifier $\tilde{f}_t$, which is fused with $f_t$ to form the final detector $F_t$. Please the detailed Algorithm~\ref{algorithm} in Appendix~\ref{appendix:LLM-DAS details}.
The core rationale of this framework is to improve detector robustness. Traditional one-class detectors rely on fragile assumptions that are often violated by real-world anomalies. By leveraging an LLM's unique capability for algorithmic reasoning, our method programmatically generates samples that specifically target these assumptions, exposing the detector to edge cases it would otherwise miss. This targeted augmentation leads to a more nuanced decision boundary and better generalization(a theoretical intuition is in Appendix~\ref{appendix:theoretical_analysis}). Overall, our method offers three key advantages: (i) \textbf{Data privacy}, as the LLM never observes the data; (ii) \textbf{Model-specificity}, leading to highly effective synthesis; and (iii) \textbf{Generalizability}, as the framework is a plug-and-play module for various detectors.

\section{Experiments}
\subsection{Experimental Setup}
\textbf{Datasets.}
We adopt an extensive benchmark with 36 datasets selected from Outlier Detection DataSets (ODDS)~\citep{rayana2016odds} and Anomaly Detection Benchmark (ADBench)~\citep{han2022adbench} following previous works \citep{yin2024mcm, ye2025drl}.
These datasets span diverse domains, including healthcare, science, and social sciences.
The dataset properties are summarized in Appendix~\ref{appendix:datasets details}.
Per the literature~\citep{zong2018deep, bergman2020classification, yin2024mcm, thimonier2024beyond, ye2025drl}, we construct the training set by subsampling 50\% of the normal samples. The remaining 50\% of the normal samples are then combined with the entire set of anomalies to form the test set.
We employ Area Under the Precision-Recall Curve (AUC-PR) and Area Under the Receiver Operating Characteristic Curve (AUC-ROC) as our evaluation criteria.

\textbf{Implementation and Baselines Details.}
For the code generation phase, we utilized the Gemini-2.5-Pro API. This represents a modest, two-time cost per detector type (Eq.~\ref{equ:description} and Eq.~\ref{equ:code}), as the resulting synthesis  $\text{Code}^t$ is reusable across any dataset. The subsequent anomaly instantiation and binary $\tilde{f}_t$ classifier training phases were performed locally and incurred no further monetary cost. 
The number of synthesized samples is set to 10\% of the training set size by default. Following recent studies~\citep{yin2024mcm, thimonier2024beyond, ye2025drl}, we implement the hyperparameters of all detector models according to their original papers. In our experiments, we use PCA~\citep{shyu2003novel}, IForest~\citep{liu2008isolation}, OCSVM~\citep{scholkopf1999support}, ECOD~\citep{li2022ecod}, and DRL~\citep{ye2025drl} as five source  detectors (based on different assumptions) to evaluate the effectiveness of LLM-DAS. In addition to these detector-based integrations, we also conduct direct comparisons against conventional and advanced TAD baselines to further evaluate benchmark performance. Details of these detectors are provided in Appendix~\ref{appendix:baseline details}.


\begin{table}[t!]
\scriptsize
\centering
\captionsetup{font=small}
\caption{The statistical information of detection performance improvement achieved by LLM-DAS ($F_t$) over trained source detectors ($f_t$) on 36 datasets. ``Baseline'' means the average performance achieved by the corresponding source detector ($f_t$) on all datasets. ``Improv. ($\Delta$)'' indicates the average absolute performance improvement by the LLM-DAS ($F_t$) over $f_t$, ``Improv. (\%)'' indicates the average relative performance improvement in percentage. ``Win Count'' represents the number of datasets that LLM-DAS ($F_t$) made improvements over $f_t$. $p$-value measures the improvement significance across the 36 datasets ($< 5\%$ is significant).}
\label{table:main summary}
\setlength{\tabcolsep}{1.2mm}{
\begin{tabular}{c|ccccc|ccccc}
\toprule
\multirow{2}{*}{Detector} & \multicolumn{5}{c|}{AUC-PR} & \multicolumn{5}{c}{AUC-ROC}        \\ \cline{2-6} \cline{7-11}
         & Baseline & Improv. ($\Delta$) & Improv. (\%) & Win Count  & $p$-value       & Baseline & Improv. ($\Delta$) & Improv. (\%) & Win Count  & $p$-value \\ \midrule
PCA      & .5975    & .0402       & 21.50            & 30 / 36  & .0271         & .7349    & .0325       & 7.64             & 27 / 36  & .0155 \\
IForest  & .5724    & .0617       & 23.60            & 26 / 36  & .0010         & .7555    & .0385       & 10.90            & 28 / 36  & .0034 \\
OCSVM    & .5295    & .0723       & 84.49            & 25 / 36  & .0239         & .7141    & .0437       & 10.02            & 25 / 36  & .0170 \\
ECOD     & .5376    & .0512       & 23.05            & 28 / 36  & .0014         & .7056    & .0398       & 8.20             & 26 / 36  & .0183 \\
DRL      & .7437    & .0412       & 15.81            & 25 / 36  & .0238         & .8509    & .0272       & 3.79             & 21 / 36  & .0035 \\
\bottomrule
\end{tabular}}
\end{table}

\subsection{Main Results}
\textbf{Improvement over Source Detectors.} 
We first evaluate the effectiveness of LLM-DAS by measuring its improvements over the original source detectors across 36 benchmark datasets. As summarized in Table~\ref{table:main summary}, overall, LLM-DAS consistently enhances all source detectors across both AUC-PR and AUC-ROC metrics. For instance, when enhancing OCSVM, LLM-DAS improves the AUC-PR by an average of 0.0723, corresponding to an 84.49\% relative improvement, and achieves gains on 25 out of 36 datasets. Similar trends can be observed for PCA, IForest, ECOD, and DRL, confirming that the ours is broadly effective across diverse detection algorithms.
To assess whether the observed improvements are statistically significant, we conduct a paired one-tailed t-test across the 36 datasets. The results show that most improvements achieve $p$-values below 0.05, indicating that the performance gains are statistically significant rather than random fluctuations. These results demonstrate that LLM-DAS can reliably enhance existing detectors by leveraging synthesized anomalies. Complete results can be found in Table~\ref{appendix:main PR} and Table~\ref{appendix:main ROC} located in the Appendix~\ref{appendix: full main}.

\paragraph{Comparison with Other Baseline Methods.}
We further compare the performance of LLM-DAS with conventional and advanced TAD baselines. Since LLM-DAS and  the recent finetune-based AnoLLM~\citep{tsai2025anollm} have some shared datasets, 
\begin{wrapfigure}{r}{0.42\textwidth}
\vspace{-3mm}
\centering
\includegraphics[width=\linewidth]{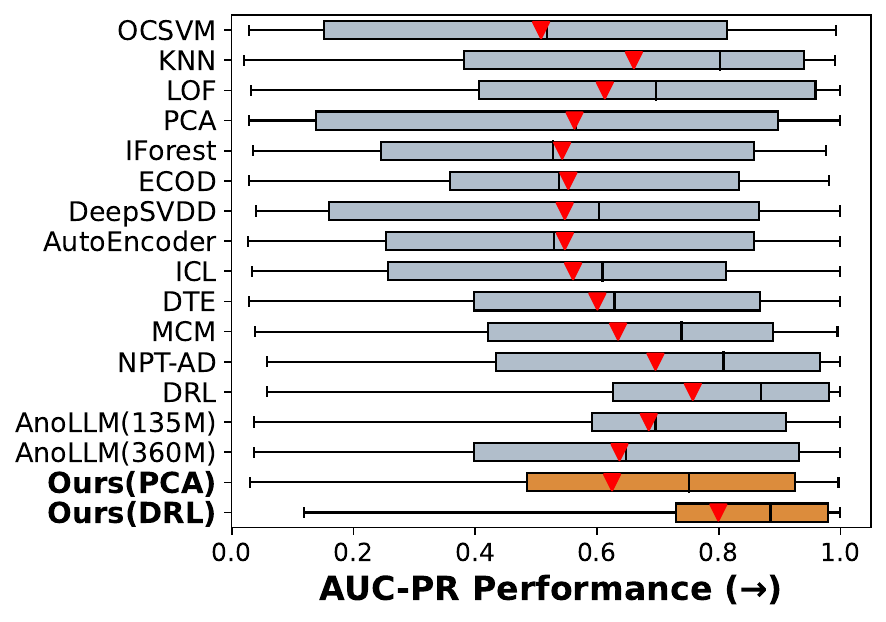}
\captionsetup{font=small}
\captionsetup{skip=0pt}
\caption{Comparison of all models’ performance across different datasets (in AUC-PR). The red triangles represent the average value. AnoLLM has two versions (135M and 360M parameters).}
\label{fig:baseline comparison}
\vspace{-3mm}
\end{wrapfigure}
we cite the results of AnoLLM due to its high computational cost, and perform other methods on the shared subsets. 
As shown in Fig.~\ref{fig:baseline comparison}, LLM-DAS, when coupled with DRL, achieves the highest average performance across these datasets. Notably, although both AnoLLM and ours leverage LLMs, the ways they are used differ fundamentally: AnoLLM fine-tunes a small LLM with open-weights on each target dataset, learning a specific model for each dataset. In contrast, we view the LLM with the reasoning ability as an algorithmic analyst: it examines the logic of detection methods and generates corresponding code, making our approach dataset-agnostic and independent of further LLM fine-tuning.
Even with a significantly lower computational cost than AnoLLM, our method still outperforms it. Full results are provided in Table~\ref{appendix:methodsPR} and Table~\ref{appendix:methodsROC} of Appendix~\ref{appendix:full comparison with other baseline}.

\subsection{Further Analysis}

\begin{figure}[b!]
     \includegraphics[width=\linewidth]{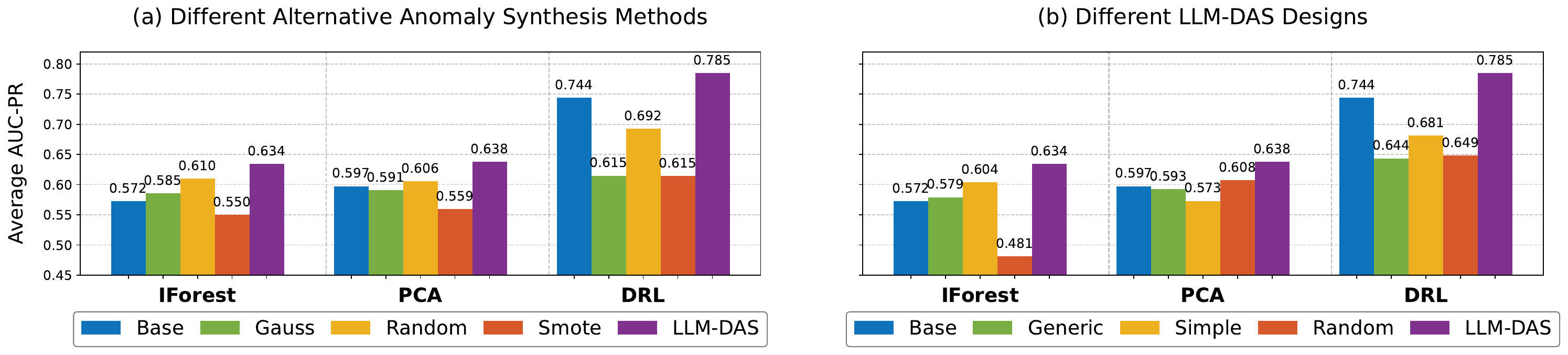}
    \captionsetup{font=small}
    \captionsetup{skip=0pt}
    \captionof{figure}{\label{fig:ablationfig4} \small{Performance comparison of anomaly synthesis methods  (left) and LLM-DAS designs across different datasets (right) in terms of AUC-PR.}}
    \vspace{-1em}
\end{figure}

\textbf{Comparison with Other Anomaly  Synthesis Methods.} 
We compare LLM-DAS with several representative anomaly synthesis strategies, including Gaussian noise injection, random outliers, and SMOTE-based~\citep{chawla2002smote} sample generation. Fig.~\ref{fig:ablationfig4} (a) reports the average AUC-PR across datasets (detailed results and base synthesis strategy details in Appendix~\ref{appendix: Alternative Anomaly Synthesis}). We can find that, naive methods usually yield inconsistent results. While methods like random outlier generation may offer marginal gains for simple detectors like IForest, they are not universally effective and can be detrimental. For instance, random synthesis severely degrades the performance of sophisticated models like DRL, highlighting the risk of a generic policy conflicting with a detector's intrinsic logic. In contrast, LLM-DAS delivers consistent and substantial improvements across all models. Its detector-aware strategy generates hard anomalies tailored to each model's specific vulnerabilities, ensuring the augmented data is always informative and leads to a more robust decision boundary.



\textbf{Ablation Studies on LLM-DAS Design.} Our ablation studies (Fig.~\ref{fig:ablationfig4} (b)) validate the critical role of each core component in LLM-DAS by changing our designed prompt.  First, to test our core detector-aware premise, a Generic variant was created by removing the detector's principles from the prompt. Second, to confirm the necessity of generating hard anomalies, a Simple variant was prompted without access to the model.predict\_score() function, preventing the LLM from assessing anomaly difficulty. Finally, to evaluate our borderline heuristic, a Random variant was instructed to transform randomly selected normal samples rather than borderline normal samples into anomalies. The inferior performance of each variant confirms that detector-awareness, the capability to generate challenging anomalies, and a focus on borderline samples are all critical to the success of LLM-DAS. Please find details in Appendix~\ref{appendix:  Ablation Studies on LLM-DAS Design}.

\textbf{Cross-Detector Synthesis Results.} To validate the critical role of our detector-aware design, we conducted a cross-detector experiment to test if a synthesis strategy for one detector could enhance another. We set the base detector $f_t$ as OCSVM and applied synthesis codes generated specifically for IForest, PCA, and OCSVM itself. The results in Table~\ref{tab:cross_detector_synthesis}, which report the absolute AUC-PR change, are revealing. The ``mismatched'' scenarios (e.g., using PCA-specific code to enhance OCSVM) yielded inconsistent  outcomes. While gains were observed on some datasets, this approach introduced a significant risk of performance degradation, evidenced by a substantial drop of 0.1899 on the Vowels dataset. In contrast, the ``matched'' synthesis (using OCSVM-specific code for the OCSVM detector) provided the most stable and significant average improvement. This experiment strongly demonstrates that the success of LLM-DAS stems not from a generic augmentation effect, but from the LLM's ability to programmatically exploit the unique mechanics of the target detector. It confirms that detector-awareness is the critical factor for achieving robust performance enhancement.

\begin{table}[t]
\centering
\scriptsize
\captionsetup{font=small}
\captionsetup{skip=0pt}
\caption{Performance comparison in AUC-PR when enhancing OCSVM with the anomaly generation code designed for IForest, PCA, OCSVM, respectively. The average relative improvement is provided ($\uparrow$ \%).}
\label{tab:cross_detector_synthesis}  
\resizebox{\linewidth}{!}{
\begin{tabular}{ccrrrrrrr|c}  
\toprule  
Baseline  & Synthesis code & Vertebral & Glass  & Imgseg & Shuttle & Speech  & Cardio  & Vowels  & Average  \\
\midrule  
\multirow{3}{*}{OCSVM}   
& IForest        & \textcolor{red!60}{+0.1025}    & \textcolor{red!60}{+0.0615} & \textcolor{red!60}{+0.0546} & \textcolor{red!60}{+0.0414}  & \textcolor{blue!60}{-0.0015} & \textcolor{blue!60}{-0.1148} & \textcolor{blue!60}{-0.0905} & \textcolor{red!60}{+0.0076($\uparrow$ 1.68\%)}   \\
& PCA            & \textcolor{blue!60}{-0.0037}   & \textcolor{red!60}{+0.1108} & \textcolor{red!60}{+0.0378} & \textbf{\textcolor{red!60}{+0.0457}}  & \textcolor{blue!60}{-0.0025} & \textcolor{blue!60}{-0.1476} & \textcolor{blue!60}{-0.1899} & \textcolor{blue!60}{-0.0213($\downarrow$ 4.72\%)}  \\
& OCSVM          & \textbf{\textcolor{red}{+0.1216}}    & \textbf{\textcolor{red}{+0.111}}  & \textbf{\textcolor{red}{+0.0709}} & \textcolor{red!60}{+0.0418}  & \textbf{\textcolor{red}{+0.0015}}  & \textbf{\textcolor{blue}{-0.0531}} & \textbf{\textcolor{blue}{-0.0871}} & \textbf{\textcolor{red}{+0.0295($\uparrow$ 6.53\%)}}   \\
\bottomrule   
\end{tabular}
}
\vspace{-1em}
\end{table}


\begin{figure}[t!]
    \centering
    \includegraphics[width=\linewidth,height=4cm]{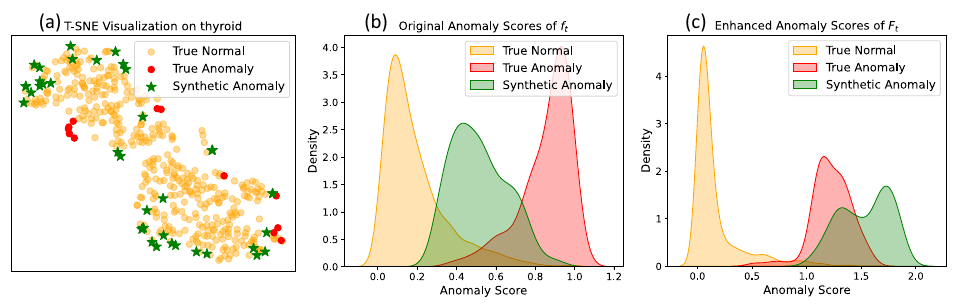}
    \captionsetup{font=small}
    \captionsetup{skip=0pt}
    \caption{\small{Visualization of synthetic hard anomalies and score distributions  on Thyroid test dataset. (a) T-SNE plots of normal, real anomaly, and LLM-DAS–generated anomaly samples. (b) Kernel density estimation (KDE) of anomaly scores from the source detector $f_t$. (c) KDE of anomaly scores from the enhanced detector $F_t$.}}
    \label{fig:visualization}
    \vspace{-2em}
\end{figure}

\textbf{Visualization and KDE Analysis of Synthetic Anomalies.}
To further examine the quality of anomalies synthesized by LLM-DAS, we utilize T-SNE  and Kernel Density Estimation (KDE) plots of anomaly scores on the Thyroid. Fig.~\ref{fig:visualization} (a) shows that the Synthetic Anomalies (green stars) are strategically placed near the boundary of the True Normal manifold, confirming they are hard anomalies that intentionally avoid the sparse, obvious outlier regions. This ``hardness'' is quantitatively proven by the KDE plots of the original score distribution in Fig.~\ref{fig:visualization} (b). The synthetic anomaly distribution severely overlaps with both true normal and anomaly distribution. This indicates the base detector struggles to distinguish the synthetic samples from normal data. Crucially, after training with LLM-DAS, the enhanced detector's KDE in Fig.~\ref{fig:visualization} (c) shows a significant rightward shift in the synthetic anomaly scores. This boundary refinement not only elevates the synthetic anomaly scores but also pushes the true anomaly score distribution further away from the normal score distribution. This confirms that exposure to LLM-DAS generated hard samples results in a significantly tighter and more robust decision boundary, improving overall generalization.

\textbf{Additional Analysis.}
We incorporate Appendix~\ref{appendix:anomaly_types} to demonstrate how LLM-DAS inherently targets detector-specific weaknesses across diverse anomaly types.
We introduce a controllable hardness-level prompting interface and a comprehensive failure-mode analysis in Appendix~\ref{appendix:hardness_control}.

\section{Conclusion}
This paper introduces LLM-DAS, a novel framework that {reimagines the role of LLMs} in tabular anomaly detection. Instead of treating LLMs as direct data processors, we have successfully repositioned them as ``algorithmists'' that reason about the intrinsic logic of detectors. This paradigm shift can address two challenges: the LLM's inherent difficulty with raw tabular data and the fragile assumptions of conventional detectors. Our core contribution is a fully programmatic, privacy-preserving approach where the LLM generates detector-specific, data-agnostic, and reusable synthesis code to produce ``hard'' anomalies, enhancing model robustness. Extensive experiments on 36 benchmarks demonstrate that LLM-DAS significantly enhances a wide array of detectors. Looking forward, the ``LLM as an algorithmist'' paradigm opens up exciting new avenues, such as enhancing algorithms in other domains or even automating algorithmic discovery. Our work paves the way for more adaptive and robust machine learning systems by integrating the algorithmic reasoning of LLMs.

\paragraph{Ethics Statement}
We confirm that this research complies with all applicable ethical guidelines and does not present any ethical issues.

\paragraph{Reproducibility Statement} To ensure reproducibility, we provide source code through the link in the abstract. Complete details regarding datasets, experimental settings, and implementation are documented in Appendix~\ref{appendix:datasets details}, Appendix~\ref{appendix:LLM-DAS details} and Appendix~\ref{appendix:baseline details}.


\subsubsection*{Acknowledgments}
The authors would like to thank the anonymous referees for their valuable comments. In this work, Hangting Ye, Jinmeng Li, Dandan Guo and Yi Chang are supported by the National Key R\&D Program of China under Grant (No. 2023YFF0905400), the National Natural Science Foundation of China (No. 623B2043, No. U2341229, No. 62306125), the Fundamental and Interdisciplinary Disciplines Breakthrough Plan of the Ministry of Education of China (No. JYB2025XDXM903), and the New Cornerstone Science Foundation through the XPLORER PRIZE.
Hongyuan Zha's research is supported in part by Shenzhen Stability Science Program 2023, and National Natural Science Foundation of China (No. 72495131).

Part of this work was carried out during Dandan Guo's visit to King Abdullah University of Science and Technology (KAUST).

\bibliography{iclr2026_conference}
\bibliographystyle{iclr2026_conference}

\newpage
\section{Appendix}


\vspace{1.2em}

{\large\bfseries \MakeUppercase{Table of Contents}}
\par\vspace{0.5em}
\hrule height 1pt
\vspace{1em}

\renewcommand{\contentsname}{}  
\renewcommand{\cftsecfont}{\bfseries}  
\renewcommand{\cftsecpagefont}{\bfseries} 
\renewcommand{\cftsubsecfont}{} 
\renewcommand{\cftsubsecpagefont}{} 

\setlength{\cftbeforesecskip}{0.6em}
\setlength{\cftbeforesubsecskip}{0.15em}

\tableofcontents

\vspace{1em}
\hrule height 0.8pt
\vspace{2em}

\subsection{Use of LLMs~\label{app:llmstate}}
Large Language Models were used as assistive tools in the preparation of this manuscript. We employed LLMs for grammar checking, LaTeX formatting, and improving the clarity of technical descriptions. 
The core scientific contributions and conclusions presented in this paper originate from the authors' work.

\subsection{Datasets details}
\label{appendix:datasets details}

We adopt an extensive benchmark with 36 datasets selected from Outlier Detection DataSets (ODDS)~\citep{rayana2016odds} and Anomaly Detection Benchmark (ADBench)~\citep{han2022adbench} following previous works \citep{yin2024mcm, ye2025drl}.
These datasets span diverse domains, including healthcare, science, and social sciences.
The dataset properties are summarized in Table~\ref{appendix:table_dataset_properties}.

\begin{table}[h!]
\centering
\caption{Dataset properties. We use 36 commonly used tabular anomaly detection datasets in this paper.}
\label{appendix:table_dataset_properties}
\begin{tabular}{lccc}
\toprule
                 & Samples & Dims & Anomalies  \\ \midrule
Abalone          & 4177    & 7    & 2081       \\
Amazon           & 10000   & 768  & 500        \\
Annthyroid       & 7200    & 6    & 534        \\
Arrhythmia       & 452     & 274  & 66         \\
Breastw          & 683     & 9    & 239        \\
Campaign         & 41188   & 62   & 4640       \\
Cardio           & 1831    & 21   & 176        \\
Cardiotocography & 2114    & 21   & 466        \\
Comm.and.crime   & 1994    & 101  & 993        \\
Fault            & 1941    & 27   & 673        \\
Glass            & 214     & 9    & 9          \\
Hepatitis        & 80      & 19   & 13         \\
Imgseg           & 2310    & 18   & 990        \\
Ionosphere       & 351     & 33   & 126        \\
Lympho           & 148     & 18   & 6          \\
Mammography      & 11183   & 6    & 260        \\
Mnist            & 7603    & 100  & 700        \\
Musk             & 3062    & 166  & 97         \\
Optdigits        & 5216    & 64   & 150        \\
Parkinson        & 195     & 22   & 147        \\
Pendigits        & 6870    & 16   & 156        \\
Pima             & 768     & 8    & 268        \\
Satellite        & 6435    & 36   & 2036       \\
Satimage-2       & 5803    & 36   & 71         \\
Shuttle          & 49097   & 9    & 3511       \\
SpamBase         & 4207    & 57   & 1679       \\
Speech           & 3686    & 400  & 61         \\
Thyroid          & 3772    & 6    & 93         \\
Vertebral        & 240     & 6    & 30         \\
Vowels           & 1456    & 12   & 50         \\
WDBC             & 367     & 30   & 10         \\
WPBC             & 198     & 33   & 47         \\
Wbc              & 378     & 30   & 21         \\
Wilt             & 4819    & 5    & 257        \\
Wine             & 129     & 13   & 10         \\
Yeast            & 1484    & 8    & 507        \\
\bottomrule
\end{tabular}
\end{table}

\subsection{LLM-DAS details}
\label{appendix:LLM-DAS details}

For the code generation phase, we utilized the Gemini-2.5-Pro API. This represents a modest, two-time cost per detector type (Eq.~\ref{equ:description} and Eq.~\ref{equ:code}), as the resulting synthesis  $\text{Code}^t$ is reusable across any dataset. The subsequent anomaly instantiation and binary $\tilde{f}_t$ classifier training phases were performed locally and incurred no further monetary cost. 
The number of synthesized samples is set to 10\% of the training set size by default. The detailed algorithm is as follows.

\begin{algorithm}[h]
\caption{LLM-DAS training and inference workflow.}
\label{algorithm}
\begin{algorithmic}[1]
    \State {\bfseries Input:} Normal-only training set $D_{\text{train}}$, test set $D_{\text{test}}$ (labels not accessed), LLM API, detector type;
    \State Obtaining $p_{\text{description}}^t$ that summarizes the algorithm  $A_t$ of $t$ and provides its pseudo code describing the concrete steps (Eq.~\ref{equ:description});
    \State Formulate a prompt $p_{\text{code}}^t$ describing the detector mechanism, synthesis objectives and requirements (Eq.~\ref{equ:prompt});
    \State Query LLM with prompt $p_{\text{code}}^t$ to generate a detector-aware anomaly synthesis code $\text{Code}^t$ (Eq.~\ref{equ:code});
    \State Train the initial score function $f_t$ on $D_{\text{train}}$;
    \State Extract interfaces $\mathcal{I}_t \;=\; D_{\text{train}}  \;\cup\; f_t \;\cup\; N_{\text{syn}}$ (Eq.~\ref{equ:interfaces_corrected});
    \State Instantiate the LLM-generated code $\text{Code}^t$ using the interfaces $\mathcal{I}_t$ to synthesize hard anomalies $D_{\text{syn}}^t$ (Eq.~\ref{equ:code_instantiation});
    \State Augment the training set: $D_{\text{aug}}^t = D_{\text{train}} \cup D_{\text{syn}}^t$ (Eq.~\ref{equ:integration});
    \State Train a binary classifier on $D_{\text{aug}}^t$ to obtain the enhanced score function $\tilde{f}_t$;
    \State Integrate the source function $f_t$ with  $\tilde{f}_t$ to form the final function $F_t$ (Eq.~\ref{equ:ensemble});
    \State {\bfseries Output:} Final anomaly scores $\{F(\mathbf{x}) \mid \mathbf{x} \in D_{\text{test}}\}$.
\end{algorithmic}
\end{algorithm}

\subsection{Baseline details}
\label{appendix:baseline details}

\subsubsection{Base Detectors Coupled with LLM-DAS}
Following recent studies~\citep{yin2024mcm, thimonier2024beyond, ye2025drl}, we implement the hyperparameters of all detector models according to their original papers. In our experiments, we use PCA~\citep{shyu2003novel}, IForest~\citep{liu2008isolation}, OCSVM~\citep{scholkopf1999support}, ECOD~\citep{li2022ecod}, and DRL~\citep{ye2025drl} as five source  detectors (based on different assumptions) to evaluate the effectiveness of LLM-DAS. The details of these detectors and LLM's opinion towards their weakness (included in the generated code $\text{Code}^t$) are as follows.

\textbf{PCA.} Principal Component Analysis (PCA) is a commonly used method for dimensionality reduction and anomaly detection. It identifies directions in the data (called principal components) along which the variance is maximized. In anomaly detection, PCA can be used to reconstruct each sample from the top principal components and compute the reconstruction error as the anomaly score. Samples with higher reconstruction error are considered more anomalous.

\textbf{IForest.} Isolation Forest (IForest) is an ensemble-based anomaly detection method that isolates anomalies instead of profiling normal data points. The key intuition is that anomalies are few and different, so they are easier to isolate using random partitions. IForest builds multiple random binary trees (isolation trees) by recursively splitting data based on randomly chosen features and split values. Data points that require fewer splits to be isolated are more likely to be anomalies, while normal points typically require more splits. The average path length across trees is used to compute an anomaly score, where shorter paths indicate higher anomaly likelihood.

\textbf{OCSVM.} One-Class Support Vector Machine (OCSVM) is a kernel-based anomaly detection method that learns a decision boundary around normal data. It maps the input data into a high-dimensional feature space using a kernel function, then finds the smallest region (a hyperplane or hypersphere) that encloses most of the normal samples. The assumption is that normal data lies in a dense region, while anomalies fall outside this boundary. OCSVM is particularly effective when the training set contains only normal samples, as it explicitly models the support of the normal data distribution.

\textbf{ECOD.} ECOD (Empirical Cumulative Distribution-based Outlier Detection) is a lightweight, univariate anomaly detection method that leverages the empirical cumulative distribution (ECDF) of each feature independently. It computes how extreme a value is relative to the distribution of that feature by evaluating its position in the ECDF. The core idea is that anomalous values often lie in the tails of the distribution. ECOD combines the tail probabilities across all features to produce an overall anomaly score, without assuming any parametric distribution. This makes it simple, interpretable, and effective for high-dimensional tabular data with only normal samples for training.

\textbf{DRL.} DRL (Decomposed Representation Learning) is an anomaly detection method designed for tabular data that learns only from normal samples. Instead of relying purely on reconstruction, DRL enforces a structured latent space where each normal representation is expressed as a linear combination of fixed, orthogonal basis vectors. This decomposition constraint encourages normal samples to occupy a well-defined subspace, while anomalies naturally fall outside of it. By amplifying discrepancies in this constrained representation space, DRL improves separation between normal and anomalous patterns. The anomaly score is derived from how poorly a test sample aligns with the decomposed latent space, optionally combined with reconstruction error.

\textbf{LLM's opinion on PCA's weakness.} 
PCA's Blind Spot: PCA-based anomaly detectors learn a low-dimensional subspace (the "principal subspace") that captures the maximum variance of the normal training data. The anomaly score is the reconstruction error, which is the distance of a sample to this subspace. Consequently, PCA is good at detecting samples that lie far from this subspace. Its primary weakness is detecting anomalies that lie *within* this learned subspace but are far from the cluster of normal data. Such points are ``anomalous'' because they are in an unobserved region, but have a low reconstruction error, fooling the detector. 


\textbf{LLM's opinion on IForest's weakness.} 
The core weakness of the Isolation Forest (IForest) lies in its reliance on axis-aligned splits to isolate data points. A sample is deemed anomalous if it can be isolated with just a few splits (resulting in a short path length). Conversely, a sample is considered normal if it is ``deep'' within the data cluster and requires numerous splits for isolation (leading to a long path length). Therefore, it exhibits insufficient sensitivity to samples that reside in the principal component subspace and are far from the data center, to synthetic anomalies within dense regions, and to samples where all feature values are normal yet their combinations are novel. This is because these anomalous samples all result in longer split paths, thereby receiving lower anomaly scores in IForest.

\textbf{LLM's opinion on OCSVM's weakness.} 
The core mechanism of One-Class SVM (OCSVM) is to learn a decision boundary that encloses normal data, and determine whether a sample is anomalous based on the sample’s relative position to this boundary (quantified by an anomaly score). The decision boundary of OCSVM is entirely defined by support vectors, while ``borderline normal samples'' themselves are the group ``closest to anomalies'' among normal data—they have the highest anomaly scores among all normal samples. When such samples are slightly pushed across the decision boundary and converted into anomalous samples, OCSVM becomes even less able to effectively distinguish between ``normal samples inside the boundary''  and ``neighboring anomalous samples outside the boundary'' : the two groups are extremely close in the feature space, and the difference in their anomaly scores is negligible. This ultimately leads to the formation of an ambiguous zone for anomaly determination near the decision boundary of OCSVM, resulting in obvious detection blind spots.

\textbf{LLM's opinion on ECOD's weakness.} 
ECOD assesses the anomaly of a sample by evaluating how extreme each feature value is in its univariate distribution,independently of other features. A hard anomaly for ECOD is therefore a sample whose feature values are individually common (i.e., not in the tails of their distributions) but whose *combination* of feature values is novel or unseen.

\textbf{LLM's opinion on DRL's weakness.} 
DRL first learns a structured latent space spanned by orthogonal basis vectors, and then uses the degree of deviation of samples from the  ``structured latent subspace defined by the decomposition process'' as the basis for anomaly scoring. Under this mechanism, normal samples form a tight alignment with this latent subspace; once a sample's latent space representation deviates, the model will classify it as an anomaly. However,DRL often struggles to effectively detect samples that are inherently anomalous yet maintain a high degree of alignment with the learned latent subspace.

\subsubsection{Additional Conventional and Advanced TAD Baselines.} In addition to these detector-based integrations, we also conduct direct comparisons against conventional and advanced TAD baselines to further evaluate benchmark performance.
Specifically, KNN~\citep{ramaswamy2000efficient}, LOF~\citep{breunig2000lof} represent classic AD approaches that continue to maintain popularity.
In addition, we compare our method to recent deep learning based methods, namely Deep SVDD~\citep{ruff2018deep}, AutoEncoder~\citep{chen2018autoencoder}, ICL~\citep{shenkar2022anomaly}, DTE~\citep{livernoche2024on}, MCM~\citep{yin2024mcm} and NPT-AD~\citep{thimonier2024beyond}.
A recent attempt AnoLLLM~\citep{tsai2025anollm} has been made to apply LLMs to tabular anomaly detection by fine-tuning them on normal samples via next-token prediction, enabling the model to act as a generative learner of the normal distribution. However, this approach still requires direct access to raw feature values and incurs the additional cost of fine-tuning, limiting its practicality.

\subsubsection{Implementation Details.} We use the popular PyOD python package~\citep{zhao2019pyod} to implement OCSVM, LOF, PCA, IForest, ECOD, Deep SVDD and AutoEncoder.
We use the DeepOD python library~\citep{xu2023deep} to implement ICL.
The implementation of the other methods is based on their official open-source code releases. 
Following latest works~\citep{yin2024mcm, ye2025drl}, We implement all baseline models' hyperparameters following their original papers.
All the methods are implemented with identical dataset partitioning and preprocessing procedures, following previous works~\citep{ye2025drl}. 

\clearpage
\subsection{Example Prompt Used in LLM-DAS}
\label{appendix:prompt_IF}
Here, we take the IForest as an example, i.e., $t=$``IForest''.
\begin{tcolorbox}[promptstyle, title={Example Prompt $p_{\text{initial}}^t$ (the input of Eq.~\ref{equ:description}) for Generating Detector Description.}]
You are an expert in anomaly detection systems. The training set contains only normal samples. I will provide you the <IForest>. Please return a description of this detector, and the pseudo code of the algorithm (step1, step2,…). The answer format for another detector is: 

\textbf{Here you should provide one paragraph for description}

The main steps of <detector type> for anomaly detection are:

\textbf{Here you should provide the main steps for this algorithm}

* xxx. 

* xxx. 

* xxx. 

...

\end{tcolorbox}

\begin{tcolorbox}[colframe=gray!40, colback=white, title={Example Prompt $p_{\text{description}}^t$ (the output of Eq.~\ref{equ:description}), Serving as One Part of $p_{\text{code}}^t$.}]
Isolation Forest (IForest) is an ensemble-based anomaly detection method that isolates anomalies instead of profiling normal data points. The key intuition is that anomalies are few and different, so they are easier to isolate using random partitions. IForest builds multiple random binary trees (isolation trees) by recursively splitting data based on randomly chosen features and split values. Data points that require fewer splits to be isolated are more likely to be anomalies, while normal points typically require more splits. The average path length across trees is used to compute an anomaly score, where shorter paths indicate higher anomaly likelihood.

The main steps of Isolation Forest (IForest) for anomaly detection are:

* Randomly select a subsample of the training data to build each tree.

* For each tree, recursively split the data by choosing a random feature and a random split value within its range, until the data is isolated or the maximum tree depth is reached.

* For each sample, compute its path length (the number of splits required for isolation) in every tree.

* Average the path lengths across all trees and normalize them to compute the anomaly score.

* Classify samples as anomalies if their score exceeds a chosen threshold (higher score → more anomalous).

\end{tcolorbox}

\begin{tcolorbox}[promptstyle, title={Example Prompt $p_{\text{code}}^t = p_{\text{description}}^t + \ p_{\text{objective}}^t + p_{\text{requirements}}^t$ (the input of Eq.~\ref{equ:code}) for Generating Synthesis Code}]
You are an expert in anomaly detection systems. The training set contains only normal samples. We use a IForest detector, where the anomaly score is computed using model.predict\_score(). The higher the score, the more anomalous the sample.  

\textbf{The description of IForest.}

Isolation Forest (IForest) is an ensemble-based anomaly detection method that isolates anomalies instead of profiling normal data points. The key intuition is that anomalies are few and different, so they are easier to isolate using random partitions. IForest builds multiple random binary trees (isolation trees) by recursively splitting data based on randomly chosen features and split values. Data points that require fewer splits to be isolated are more likely to be anomalies, while normal points typically require more splits. The average path length across trees is used to compute an anomaly score, where shorter paths indicate higher anomaly likelihood.

The main steps of Isolation Forest (IForest) for anomaly detection are:

* Randomly select a subsample of the training data to build each tree.

* For each tree, recursively split the data by choosing a random feature and a random split value within its range, until the data is isolated or the maximum tree depth is reached.

* For each sample, compute its path length (the number of splits required for isolation) in every tree.

* Average the path lengths across all trees and normalize them to compute the anomaly score.

* Classify samples as anomalies if their score exceeds a chosen threshold (higher score → more anomalous).

\textbf{Your task} is to write a Python function generate\_hard\_anomalies(...) that generates anomalies which are the most difficult for the IForest detector to detect. This means that you the generated anomalies should have relatively low anomaly score, thus they are hard to be detected. But these anomalies are helpful to build a more robust detector. After the Python function is completed, users can provide the function with:

* A trained IForest model (model) that exposes predict\_score(),

* training samples (X\_train).

\textbf{Requirements:}
Your should strictly follow below requirements:

1. You must use your expertise to give anomalies generation policies that are specific designed for IForest, not a model-agnostic policy.

2. Generated samples should have as low a score as possible from model.predict\_score(). To achieve it, you can first find the set of ‘borderline’ normal training samples based on your unique and professional understanding to IForest, not only based on the anomaly score. Then transform them to anomalies that is tailor-designed for IForest. Please note that the transformation function should be specific for IForest, which means that it is not a general transformation for other detectors.

3. For the model, you can only use the function model.predict\_score.

4. Use NumPy to generate the samples, and output an array of shape (n\_samples, d). And it should generate anomalies as much as I want.

5. The function should allow setting:

    * the number of samples (n\_samples),
    
    * the trained IForest model (model),
    
    * training samples (X\_train).

Return only the complete Python function generate\_hard\_anomalies(...), with policy you used for genenrating anomalies and clear comments explaining key steps.
\end{tcolorbox}

\definecolor{lightgreen}{RGB}{184, 219, 179}  
\definecolor{darkgreen}{RGB}{114, 176, 99}  
\definecolor{grayblue}{RGB}{113, 154, 172}  
\definecolor{darkblue}{RGB}{62, 96, 141}  
\definecolor{blue2}{RGB}{68, 133, 199}  
\definecolor{gray2}{RGB}{120,120,120}  
\renewcommand{\lstlistingname}{Code}
\lstset{
    basicstyle = \fontfamily{qcr}\selectfont\small, 
     keywordstyle = \color{magenta}, 
    commentstyle = \color{darkgreen},
    stringstyle = \color{gray2},   
    numbers = left,                  
    numberstyle=\tiny\color{red},
    firstnumber=1,
    showspaces = false,                 
    showstringspaces = false,                 
    frame = lrtb,   
    breaklines = true,                  
    columns = fixed,                 
    basewidth = 0.55em,                
    escapeinside = {},                    
    lineskip=0pt,
}

\lstdefinestyle{Python}{
    language = Python,
    basicstyle = \fontfamily{qcr}\selectfont\small, 
    numberstyle = \tiny\color{gray},
    breaklines = true,
}
\subsection{Example Generated Code by LLM}
\label{appendix:generated_code}
\begin{lstlisting}[
    style = Python,
    caption = {Generated Code (the output of Eq.~\ref{equ:code}) for IForest-Specific Hard Anomaly Generation}, 
    label = {code:IForest}
]
import numpy as np
def generate_hard_anomalies(n_samples: int, model, X_train: np.ndarray):
    """
    Generates anomalies that are difficult for a trained Isolation Forest (IForest) model to detect.

    These "hard" anomalies are designed to have relatively low anomaly scores, making them resemble normal data from the perspective of the IForest, thus challenging the detector.

    Policy for Generating IForest-Specific Hard Anomalies:
    ---------------------------------------------------------
    The core weakness of IForest is that it relies on axis-aligned splits to isolate points.
    A sample is considered anomalous if it can be isolated in a few splits (short path length).
    Conversely, a sample is normal if it is "deep" within the data cloud, requiring many splits to be isolated (long path length).

    Our strategy exploits this by creating new points that are conceptually anomalous but are geometrically located in a way that maximizes their path length, thereby fooling the IForest.

    The policy consists of two main steps:

    1.  Identify 'Borderline' Normal Samples:
        These are the normal training samples that lie on the fringe or edge of the dense data cloud. 
        The IForest model is already less certain about these points, assigning them the highest anomaly scores among all normal samples. 
        They serve as perfect "seeds" for our hard anomalies because they are already close to the decision boundary. 
        We identify these by finding the training samples in the top percentile of anomaly scores.

    2.  Transform Seeds into Hard Anomalies via 'Controlled Extrapolation':
        We transform these borderline seeds into anomalies. 
        A naive transformation (e.g., adding large random noise) would create an obvious outlier with a short path length, which is an*easy* anomaly. 
        To create a *hard* anomaly, we must move the seed point in a way that doesn't significantly shorten its path length.

        The chosen transformation is a controlled extrapolation along the vector pointing from the center of the training data to the borderline seed point.
        -  `new_anomaly = seed + magnitude * (seed - data_center)`
        This pushes the borderline point slightly further away from the data's core, making it a true anomaly. 
        However, because it's moved along an existing axis of the data's distribution, it remains "camouflaged". 
        The new point is not in a completely isolated region of the feature space but rather just outside the densest part of the manifold.
        An IForest will still require a relatively high number of splits to isolate it compared to a random, distant outlier, resulting in a deceptively low anomaly score.

    This tailored approach ensures the generated anomalies specifically target the operational logic of the Isolation Forest, making them truly "hard" to detect.

    Args:
        n_samples (int): The number of hard anomaly samples to generate.
        model: A trained Isolation Forest model with a `predict_score()` method.The method should return higher scores for more anomalous samples.
        X_train (np.ndarray): The training data (normal samples) used to train the model,with shape (n_train_samples, n_features).

    Returns:
        np.ndarray: An array of generated hard anomalies with shape (n_samples, n_features).
    """
    # 1. All package imports must be done inside the function.
    import numpy as np

    if not hasattr(model, 'predict_score'):
        raise AttributeError("The provided model must have a 'predict_score' method.")
    if not isinstance(X_train, np.ndarray) or X_train.ndim != 2:
        raise ValueError("X_train must be a 2D NumPy array.")
    if n_samples <= 0:
        raise ValueError("n_samples must be a positive integer.")

    # 2. Identify 'borderline' normal training samples.
    # These are the normal samples that the model already finds most anomalous.
    # We define them as the top 20% of training samples with the highest anomaly scores.
    # This percentile is a tunable parameter but 80 (for top 20%) is a robust choice.
    print("Step 1: Identifying borderline normal samples from the training set...")
    train_scores = model.predict_score(X_train)
    
    # Find the score threshold for the 80th percentile.
    # Samples with scores above this are our "borderline" seeds.
    score_threshold = np.percentile(train_scores, 80)
    
    borderline_indices = np.where(train_scores >= score_threshold)[0]
    
    # If no samples are above the threshold (e.g., all scores are identical),
    # fall back to using the entire training set as potential seeds.
    if len(borderline_indices) == 0:
        borderline_indices = np.arange(len(X_train))
        
    borderline_samples = X_train[borderline_indices]
    print(f"Found {len(borderline_samples)} borderline samples to use as seeds.")

    # 3. Transform borderline samples into hard anomalies.
    # We will generate each hard anomaly from a randomly chosen borderline seed.
    print(f"Step 2: Generating {n_samples} hard anomalies via controlled extrapolation...")
    
    # Calculate the geometric center (mean) of the normal data distribution.
    data_center = np.mean(X_train, axis=0)
    
    hard_anomalies = []
    for _ in range(n_samples):
        # Randomly select a borderline sample to serve as the base for our new anomaly.
        # We sample with replacement, allowing us to generate more anomalies than seeds.
        seed_idx = np.random.randint(0, len(borderline_samples))
        seed_sample = borderline_samples[seed_idx]
        
        # Calculate the direction vector pointing from the data center to the seed sample.
        # This vector represents a direction of natural variation in the data.
        direction_vector = seed_sample - data_center
        
        # Define a small, random magnitude for the extrapolation.
        # A small magnitude ensures the new point is not an obvious outlier but just
        # beyond the boundary of the normal data cloud.
        # Values between 5% and 25% of the vector length work well.
        magnitude = np.random.uniform(0.05, 0.25)
        
        # Create the new hard anomaly by pushing the seed sample along the direction vector.
        perturbation = direction_vector * magnitude
        new_anomaly = seed_sample + perturbation
        
        hard_anomalies.append(new_anomaly)

    print("Generation complete.")
    # 4. Return the generated samples as a NumPy array.
    return np.array(hard_anomalies)
\end{lstlisting} 
\subsection{Full Results of Improvement Over Source Detectors}
\label{appendix: full main}

We evaluate the effectiveness of LLM-DAS by measuring its improvements over the original source detectors across 36 benchmark datasets. The full results are provided in Table~\ref{appendix:main PR} and Table~\ref{appendix:main ROC}.

\begin{table}
\tiny
\centering
\caption{The performance improvement achieved by LLM-DAS ($F_t$) over trained source detectors ($f_t$) on 36 datasets in terms of AUC-PR ($\uparrow$).} 
\label{appendix:main PR}
\setlength{\tabcolsep}{1pt}
\resizebox{\linewidth}{!}{
\begin{tabular}{lcccccccccc}
\toprule

                 & PCA     & +LLM-DAS & IForest     & +LLM-DAS & OCSVM   & +LLM-DAS & ECOD        & +LLM-DAS & DRL         & +LLM-DAS  \\ \midrule  
Abalone          & 0.839{\scalebox{0.7}{±0}}  & \textbf{0.841{\scalebox{0.7}{±0}}}                  & 0.848{\scalebox{0.7}{±0.003}} & \textbf{0.855{\scalebox{0.7}{±0.005}}}              & 0.846{\scalebox{0.7}{±0}} & \textbf{0.849{\scalebox{0.7}{±0.005}}}              & 0.655{\scalebox{0.7}{±0}}     & 0.645{\scalebox{0.7}{±0.002}}              & 0.885{\scalebox{0.7}{±0.006}} & \textbf{0.902{\scalebox{0.7}{±0.001}}}               \\
Amazon           & 0.107{\scalebox{0.7}{±0}} & \textbf{0.108{\scalebox{0.7}{±0.001}}}              & 0.109{\scalebox{0.7}{±0.002}} & 0.109{\scalebox{0.7}{±0.002}}              & 0.105{\scalebox{0.7}{±0}} & \textbf{0.106{\scalebox{0.7}{±0}}}                  & 0.104{\scalebox{0.7}{±0}}     & 0.104{\scalebox{0.7}{±0.002}}              & 0.121{\scalebox{0.7}{±0.007}} & 0.12{\scalebox{0.7}{±0.008}}                \\
Annthyroid       & 0.566{\scalebox{0.7}{±0}} & \textbf{0.66{\scalebox{0.7}{±0.009}}}               & 0.615{\scalebox{0.7}{±0.053}} & \textbf{0.62{\scalebox{0.7}{±0.055}}}               & 0.183{\scalebox{0.7}{±0}} & \textbf{0.248{\scalebox{0.7}{±0.01}}}               & 0.4{\scalebox{0.7}{±0}}       & 0.304{\scalebox{0.7}{±0.008}}              & 0.676{\scalebox{0.7}{±0.018}} & \textbf{0.762{\scalebox{0.7}{±0.021}}}               \\
Arrhythmia       & 0.534{\scalebox{0.7}{±0}} & \textbf{0.551{\scalebox{0.7}{±0.013}}}              & 0.51{\scalebox{0.7}{±0.034}}  & \textbf{0.602{\scalebox{0.7}{±0.037}}}              & 0.534{\scalebox{0.7}{±0}} & \textbf{0.609{\scalebox{0.7}{±0.012}}}              & 0.446{\scalebox{0.7}{±0.005}} & \textbf{0.466{\scalebox{0.7}{±0.003}}}              & 0.627{\scalebox{0.7}{±0.036}} & \textbf{0.73{\scalebox{0.7}{±0.014}}}    \\
Breastw          & 0.993{\scalebox{0.7}{±0}} & \textbf{0.996{\scalebox{0.7}{±0.001}}}              & 0.945{\scalebox{0.7}{±0}}     & \textbf{0.995{\scalebox{0.7}{±0.002}}}              & 0.993{\scalebox{0.7}{±0}} & \textbf{0.995{\scalebox{0.7}{±0}}}                  & 0.952{\scalebox{0.7}{±0}}     & \textbf{0.977{\scalebox{0.7}{±0.038}}}              & 0.997{\scalebox{0.7}{±0.004}} & 0.997{\scalebox{0.7}{±0.001}}               \\
Campaign         & 0.488{\scalebox{0.7}{±0}} & 0.488{\scalebox{0.7}{±0}}                  & 0.461{\scalebox{0.7}{±0.007}} & 0.452{\scalebox{0.7}{±0.007}}              & 0.475{\scalebox{0.7}{±0}} & 0.435{\scalebox{0.7}{±0}}                  & 0.471{\scalebox{0.7}{±0}}     & \textbf{0.503{\scalebox{0.7}{±0}}}                  & 0.501{\scalebox{0.7}{±0.006}} & 0.495{\scalebox{0.7}{±0.004}}               \\
Cardio           & 0.863{\scalebox{0.7}{±0}} & 0.824{\scalebox{0.7}{±0.014}}              & 0.702{\scalebox{0.7}{±0.039}} & 0.622{\scalebox{0.7}{±0.032}}              & 0.861{\scalebox{0.7}{±0}} & 0.808{\scalebox{0.7}{±0.015}}              & 0.364{\scalebox{0.7}{±0}}     & \textbf{0.716{\scalebox{0.7}{±0.026}}}              & 0.833{\scalebox{0.7}{±0.024}} & \textbf{0.886{\scalebox{0.7}{±0.038}}}               \\
Cardiotocography & 0.697{\scalebox{0.7}{±0}} & \textbf{0.734{\scalebox{0.7}{±0.006}}}              & 0.604{\scalebox{0.7}{±0.032}} & \textbf{0.638{\scalebox{0.7}{±0.041}}}              & 0.662{\scalebox{0.7}{±0}} & \textbf{0.703{\scalebox{0.7}{±0.057}}}              & 0.697{\scalebox{0.7}{±0}}     & 0.664{\scalebox{0.7}{±0.009}}              & 0.754{\scalebox{0.7}{±0}}     & \textbf{0.805{\scalebox{0.7}{±0.026}}}               \\
Comm.and.crime   & 0.889{\scalebox{0.7}{±0}} & 0.881{\scalebox{0.7}{±0.002}}              & 0.894{\scalebox{0.7}{±0.009}} & \textbf{0.896{\scalebox{0.7}{±0.006}}}              & 0.837{\scalebox{0.7}{±0}} & \textbf{0.843{\scalebox{0.7}{±0.022}}}              & 0.685{\scalebox{0.7}{±0}}     & \textbf{0.702{\scalebox{0.7}{±0.021}}}              & 0.916{\scalebox{0.7}{±0.004}} & \textbf{0.947{\scalebox{0.7}{±0.004}}}               \\
Fault            & 0.604{\scalebox{0.7}{±0}} & 0.603{\scalebox{0.7}{±0.009}}              & 0.595{\scalebox{0.7}{±0.014}} & 0.577{\scalebox{0.7}{±0.02}}               & 0.606{\scalebox{0.7}{±0}} & 0.542{\scalebox{0.7}{±0.01}}               & 0.517{\scalebox{0.7}{±0}}     & \textbf{0.573{\scalebox{0.7}{±0.007}}}              & 0.665{\scalebox{0.7}{±0}}     & \textbf{0.705{\scalebox{0.7}{±0.043}}}               \\
Glass            & 0.09{\scalebox{0.7}{±0}}  & \textbf{0.128{\scalebox{0.7}{±0.016}}}              & 0.095{\scalebox{0.7}{±0.006}} & \textbf{0.12{\scalebox{0.7}{±0.02}}}                & 0.09{\scalebox{0.7}{±0}}  & \textbf{0.201{\scalebox{0.7}{±0.033}}}              & 0.111{\scalebox{0.7}{±0}}     & \textbf{0.125{\scalebox{0.7}{±0.045}}}              & 0.167{\scalebox{0.7}{±0}}     & \textbf{0.493{\scalebox{0.7}{±0.116}}}               \\
Hepatitis        & 0.583{\scalebox{0.7}{±0}} & \textbf{0.626{\scalebox{0.7}{±0.038}}}              & 0.418{\scalebox{0.7}{±0.022}} & \textbf{0.566{\scalebox{0.7}{±0.069}}}              & 0.282{\scalebox{0.7}{±0}} & \textbf{0.3{\scalebox{0.7}{±0.007}}}                & 0.405{\scalebox{0.7}{±0}}     & \textbf{0.477{\scalebox{0.7}{±0.024}}}              & 0.663{\scalebox{0.7}{±0.109}} & \textbf{0.696{\scalebox{0.7}{±0.107}}}               \\
Imgseg           & 0.772{\scalebox{0.7}{±0}} & \textbf{0.824{\scalebox{0.7}{±0.014}}}              & 0.756{\scalebox{0.7}{±0.012}} & \textbf{0.859{\scalebox{0.7}{±0.023}}}              & 0.788{\scalebox{0.7}{±0}} & \textbf{0.859{\scalebox{0.7}{±0.011}}}              & 0.737{\scalebox{0.7}{±0}}     & \textbf{0.844{\scalebox{0.7}{±0.02}}}               & 0.924{\scalebox{0.7}{±0.007}} & \textbf{0.944{\scalebox{0.7}{±0.008}}}               \\
Ionosphere       & 0.897{\scalebox{0.7}{±0}} & \textbf{0.926{\scalebox{0.7}{±0.026}}}              & 0.977{\scalebox{0.7}{±0.008}} & 0.869{\scalebox{0.7}{±0.035}}              & 0.897{\scalebox{0.7}{±0}} & 0.838{\scalebox{0.7}{±0.012}}              & 0.971{\scalebox{0.7}{±0}}     & 0.896{\scalebox{0.7}{±0.007}}              & 0.99{\scalebox{0.7}{±0.013}}  & 0.985{\scalebox{0.7}{±0.004}}               \\
Lympho           & 1{\scalebox{0.7}{±0}}     & 0.989{\scalebox{0.7}{±0.022}}              & 0.959{\scalebox{0.7}{±0.023}} & \textbf{0.979{\scalebox{0.7}{±0.02}}}               & 0.811{\scalebox{0.7}{±0}} & \textbf{0.934{\scalebox{0.7}{±0.03}}}               & 0.897{\scalebox{0.7}{±0}}     & \textbf{0.97{\scalebox{0.7}{±0.028}}}               & 1{\scalebox{0.7}{±0.062}}     & 0.871{\scalebox{0.7}{±0.07}}                \\
Mammography      & 0.417{\scalebox{0.7}{±0}} & \textbf{0.579{\scalebox{0.7}{±0.008}}}              & 0.333{\scalebox{0.7}{±0.032}} & \textbf{0.571{\scalebox{0.7}{±0.019}}}              & 0.418{\scalebox{0.7}{±0}} & \textbf{0.536{\scalebox{0.7}{±0.022}}}              & 0.538{\scalebox{0.7}{±0}}     & \textbf{0.543{\scalebox{0.7}{±0.007}}}              & 0.841{\scalebox{0.7}{±0.02}}  & 0.529{\scalebox{0.7}{±0.025}}               \\
Mnist            & 0.65{\scalebox{0.7}{±0}}  & 0.65{\scalebox{0.7}{±0}}                   & 0.535{\scalebox{0.7}{±0.013}} & \textbf{0.601{\scalebox{0.7}{±0.01}}}               & 0.169{\scalebox{0.7}{±0}} & \textbf{0.181{\scalebox{0.7}{±0.012}}}              & 0.302{\scalebox{0.7}{±0.007}} & \textbf{0.322{\scalebox{0.7}{±0.004}}}              & 0.887{\scalebox{0.7}{±0.017}} & \textbf{0.911{\scalebox{0.7}{±0.018}}}               \\
Musk             & 1{\scalebox{0.7}{±0}}     & 0.997{\scalebox{0.7}{±0.006}}              & 0.528{\scalebox{0.7}{±0.202}} & \textbf{1{\scalebox{0.7}{±0}}}                      & 0.061{\scalebox{0.7}{±0}} & \textbf{0.868{\scalebox{0.7}{±0.075}}}              & 0.982{\scalebox{0.7}{±0}}     & \textbf{1{\scalebox{0.7}{±0}}}                      & 1{\scalebox{0.7}{±0.009}}     & 1{\scalebox{0.7}{±0}}                       \\
Optdigits        & 0.06{\scalebox{0.7}{±0}}  & 0.06{\scalebox{0.7}{±0}}                   & 0.157{\scalebox{0.7}{±0.015}} & \textbf{0.179{\scalebox{0.7}{±0.043}}}              & 0.069{\scalebox{0.7}{±0}} & \textbf{0.928{\scalebox{0.7}{±0}}}                  & 0.067{\scalebox{0.7}{±0.006}} & \textbf{0.087{\scalebox{0.7}{±0.003}}}              & 0.936{\scalebox{0.7}{±0}}     & \textbf{0.957{\scalebox{0.7}{±0.017}}}               \\
Parkinson        & 0.93{\scalebox{0.7}{±0}}  & \textbf{0.944{\scalebox{0.7}{±0.004}}}              & 0.96{\scalebox{0.7}{±0.004}}  & 0.951{\scalebox{0.7}{±0.011}}              & 0.889{\scalebox{0.7}{±0}} & \textbf{0.92{\scalebox{0.7}{±0.021}}}               & 0.891{\scalebox{0.7}{±0}}     & 0.863{\scalebox{0.7}{±0.025}}              & 0.921{\scalebox{0.7}{±0.009}} & \textbf{0.966{\scalebox{0.7}{±0.007}}}              \\
Pendigits        & 0.386{\scalebox{0.7}{±0}} & \textbf{0.485{\scalebox{0.7}{±0.047}}}              & 0.513{\scalebox{0.7}{±0.02}}  & \textbf{0.602{\scalebox{0.7}{±0.049}}}              & 0.518{\scalebox{0.7}{±0}} & 0.491{\scalebox{0.7}{±0.044}}              & 0.415{\scalebox{0.7}{±0}}     & \textbf{0.498{\scalebox{0.7}{±0.028}}}              & 0.936{\scalebox{0.7}{±0.135}} & 0.885{\scalebox{0.7}{±0.025}}               \\
Pima             & 0.701{\scalebox{0.7}{±0}} & \textbf{0.751{\scalebox{0.7}{±0.005}}}              & 0.666{\scalebox{0.7}{±0.007}} & \textbf{0.74{\scalebox{0.7}{±0.017}}}               & 0.701{\scalebox{0.7}{±0}} & 0.651{\scalebox{0.7}{±0.036}}              & 0.588{\scalebox{0.7}{±0}}     & 0.569{\scalebox{0.7}{±0.014}}              & 0.745{\scalebox{0.7}{±0.012}} & \textbf{0.791{\scalebox{0.7}{±0.007}}}               \\
Satellite        & 0.778{\scalebox{0.7}{±0}} & \textbf{0.792{\scalebox{0.7}{±0.001}}}              & 0.858{\scalebox{0.7}{±0.006}} & 0.808{\scalebox{0.7}{±0.007}}              & 0.778{\scalebox{0.7}{±0}} & \textbf{0.832{\scalebox{0.7}{±0.005}}}              & 0.833{\scalebox{0.7}{±0}}     & 0.803{\scalebox{0.7}{±0.006}}              & 0.869{\scalebox{0.7}{±0.008}} & \textbf{0.894{\scalebox{0.7}{±0.009}}}               \\
Satimage-2       & 0.919{\scalebox{0.7}{±0}} & \textbf{0.931{\scalebox{0.7}{±0.01}}}               & 0.885{\scalebox{0.7}{±0.007}} & 0.871{\scalebox{0.7}{±0.085}}              & 0.919{\scalebox{0.7}{±0}} & 0.9{\scalebox{0.7}{±0.014}}                & 0.778{\scalebox{0.7}{±0}}     & \textbf{0.934{\scalebox{0.7}{±0.008}}}             & 0.97{\scalebox{0.7}{±0.044}}  & \textbf{0.98{\scalebox{0.7}{±0.006}}}                \\
Shuttle          & 0.963{\scalebox{0.7}{±0}} & \textbf{0.991{\scalebox{0.7}{±0.002}}}              & 0.917{\scalebox{0.7}{±0.005}} & \textbf{0.995{\scalebox{0.7}{±0.001}}}              & 0.949{\scalebox{0.7}{±0}} & \textbf{0.991{\scalebox{0.7}{±0.001}}}              & 0.982{\scalebox{0.7}{±0}}     & \textbf{0.992{\scalebox{0.7}{±0}}}                  & 0.982{\scalebox{0.7}{±0.006}} & \textbf{0.999{\scalebox{0.7}{±0}}}                  \\
SpamBase         & 0.818{\scalebox{0.7}{±0}} & \textbf{0.823{\scalebox{0.7}{±0.002}}}              & 0.89{\scalebox{0.7}{±0.006}}  & 0.889{\scalebox{0.7}{±0.007}}              & 0.814{\scalebox{0.7}{±0}} & 0.814{\scalebox{0.7}{±0}}                  & 0.713{\scalebox{0.7}{±0}}     & \textbf{0.737{\scalebox{0.7}{±0.007}}}              & 0.841{\scalebox{0.7}{±0.008}} & \textbf{0.933{\scalebox{0.7}{±0.013}}}               \\
Speech           & 0.028{\scalebox{0.7}{±0}} & \textbf{0.029{\scalebox{0.7}{±0.001}}}              & 0.035{\scalebox{0.7}{±0.009}} & 0.032{\scalebox{0.7}{±0.007}}              & 0.028{\scalebox{0.7}{±0}} & \textbf{0.029{\scalebox{0.7}{±0.002}}}              & 0.029{\scalebox{0.7}{±0}}     & \textbf{0.031{\scalebox{0.7}{±0.003}}}              & 0.058{\scalebox{0.7}{±0.011}} & \textbf{0.119{\scalebox{0.7}{±0.01}}}                \\
Thyroid          & 0.813{\scalebox{0.7}{±0}} & \textbf{0.815{\scalebox{0.7}{±0.022}}}              & 0.606{\scalebox{0.7}{±0.025}} & \textbf{0.886{\scalebox{0.7}{±0.053}}}              & 0.813{\scalebox{0.7}{±0}} & 0.633{\scalebox{0.7}{±0.014}}              & 0.681{\scalebox{0.7}{±0}}     & \textbf{0.765{\scalebox{0.7}{±0.01}}}               & 0.863{\scalebox{0.7}{±0.021}} & \textbf{0.877{\scalebox{0.7}{±0.017}}}              \\
Vertebral        & 0.138{\scalebox{0.7}{±0}} & \textbf{0.143{\scalebox{0.7}{±0.004}}}              & 0.134{\scalebox{0.7}{±0.001}} & \textbf{0.269{\scalebox{0.7}{±0.084}}}              & 0.152{\scalebox{0.7}{±0}} & \textbf{0.273{\scalebox{0.7}{±0.038}}}              & 0.192{\scalebox{0.7}{±0}}     & \textbf{0.403{\scalebox{0.7}{±0.068}}}              & 0.285{\scalebox{0.7}{±0.043}} & \textbf{0.566{\scalebox{0.7}{±0.093}}}               \\
Vowels           & 0.105{\scalebox{0.7}{±0}} & \textbf{0.109{\scalebox{0.7}{±0.01}}}               & 0.098{\scalebox{0.7}{±0.013}} & \textbf{0.178{\scalebox{0.7}{±0.011}}}              & 0.297{\scalebox{0.7}{±0}} & 0.21{\scalebox{0.7}{±0.02}}                & 0.177{\scalebox{0.7}{±0}}     & \textbf{0.274{\scalebox{0.7}{±0.015}}}              & 0.451{\scalebox{0.7}{±0.025}} & \textbf{0.892{\scalebox{0.7}{±0.016}}}               \\
Wbc              & 0.839{\scalebox{0.7}{±0}} & \textbf{0.84{\scalebox{0.7}{±0.013}}}               & 0.857{\scalebox{0.7}{±0.015}} & 0.823{\scalebox{0.7}{±0.033}}              & 0.839{\scalebox{0.7}{±0}} & 0.793{\scalebox{0.7}{±0.034}}              & 0.722{\scalebox{0.7}{±0}}     & \textbf{0.881{\scalebox{0.7}{±0.016}}}              & 0.974{\scalebox{0.7}{±0.032}} & 0.934{\scalebox{0.7}{±0.018}}               \\
WDBC             & 0.983{\scalebox{0.7}{±0}} & 0.98{\scalebox{0.7}{±0.01}}                & 0.975{\scalebox{0.7}{±0.029}} & \textbf{0.994{\scalebox{0.7}{±0.009}}}              & 0.435{\scalebox{0.7}{±0}} & \textbf{0.874{\scalebox{0.7}{±0.059}}}              & 0.773{\scalebox{0.7}{±0}}     & \textbf{0.93{\scalebox{0.7}{±0.046}}}               & 1{\scalebox{0.7}{±0.016}}     & 1{\scalebox{0.7}{±0}}                       \\
Wilt             & 0.064{\scalebox{0.7}{±0}} & \textbf{0.097{\scalebox{0.7}{±0.011}}}              & 0.085{\scalebox{0.7}{±0.004}} & \textbf{0.278{\scalebox{0.7}{±0.049}}}              & 0.225{\scalebox{0.7}{±0}} & 0.219{\scalebox{0.7}{±0.06}}               & 0.077{\scalebox{0.7}{±0}}     & \textbf{0.375{\scalebox{0.7}{±0.043}}}              & 0.454{\scalebox{0.7}{±0.058}} & \textbf{0.593{\scalebox{0.7}{±0.05}}}                \\
Wine             & 0.133{\scalebox{0.7}{±0}} & \textbf{0.846{\scalebox{0.7}{±0.044}}}              & 0.246{\scalebox{0.7}{±0.026}} & \textbf{0.456{\scalebox{0.7}{±0.076}}}              & 0.142{\scalebox{0.7}{±0}} & \textbf{0.334{\scalebox{0.7}{±0.172}}}              & 0.358{\scalebox{0.7}{±0}}     & 0.289{\scalebox{0.7}{±0.098}}              & 1{\scalebox{0.7}{±0.03}}      & 1{\scalebox{0.7}{±0}}                       \\
WPBC             & 0.394{\scalebox{0.7}{±0}} & \textbf{0.408{\scalebox{0.7}{±0.015}}}              & 0.376{\scalebox{0.7}{±0.012}} & \textbf{0.399{\scalebox{0.7}{±0.017}}}              & 0.397{\scalebox{0.7}{±0}} & \textbf{0.407{\scalebox{0.7}{±0.015}}}              & 0.353{\scalebox{0.7}{±0}}     & \textbf{0.393{\scalebox{0.7}{±0.013}}}              & 0.502{\scalebox{0.7}{±0.057}} & \textbf{0.567{\scalebox{0.7}{±0.044}}}              \\
Yeast            & 0.468{\scalebox{0.7}{±0}} & \textbf{0.505{\scalebox{0.7}{±0.005}}}              & 0.465{\scalebox{0.7}{±0.004}} & \textbf{0.546{\scalebox{0.7}{±0.019}}}              & 0.48{\scalebox{0.7}{±0}}  & \textbf{0.515{\scalebox{0.7}{±0.007}}}              & 0.494{\scalebox{0.7}{±0}}     & \textbf{0.543{\scalebox{0.7}{±0.017}}}              & 0.542{\scalebox{0.7}{±0.008}} & 0.528{\scalebox{0.7}{±0.009}}               \\ \midrule  
Average AUC-PR   & 0.597{\scalebox{0.7}{±0}} & \textbf{0.638{\scalebox{0.7}{±0.011}}}              & 0.572{\scalebox{0.7}{±0.02}}  & \textbf{0.634{\scalebox{0.7}{±0.027}}}              & 0.53{\scalebox{0.7}{±0}}  & \textbf{0.602{\scalebox{0.7}{±0.024}}}              & 0.538{\scalebox{0.7}{±0.001}} & \textbf{0.589{\scalebox{0.7}{±0.019}}}              & 0.744{\scalebox{0.7}{±0.024}} & \textbf{0.785{\scalebox{0.7}{±0.024}}}              
  \\ \bottomrule         
\end{tabular}}
\end{table}

\begin{table}
\tiny
\centering
\caption{The performance improvement achieved by LLM-DAS ($F_t$) over trained source detectors ($f_t$) on 36 datasets in terms of AUC-ROC ($\uparrow$).}
\label{appendix:main ROC}
\setlength{\tabcolsep}{1pt}
\resizebox{\linewidth}{!}{
\begin{tabular}{lcccccccccc}
\toprule

                 & PCA     & +LLM-DAS & IForest     & +LLM-DAS & OCSVM   & +LLM-DAS & ECOD        & +LLM-DAS & DRL         & +LLM-DAS  \\ \midrule 
Abalone          & 0.704{\scalebox{0.7}{±0}} & \textbf{0.705{\scalebox{0.7}{±0.001}}} & 0.735{\scalebox{0.7}{±0.006}} & \textbf{0.746{\scalebox{0.7}{±0.007}}} & 0.724{\scalebox{0.7}{±0}} & \textbf{0.728{\scalebox{0.7}{±0.007}}} & 0.487{\scalebox{0.7}{±0}} & 0.475{\scalebox{0.7}{±0.002}} & 0.809{\scalebox{0.7}{±0.008}} & \textbf{0.837{\scalebox{0.7}{±0.003}}}         \\
Amazon           & 0.549{\scalebox{0.7}{±0}} & 0.549{\scalebox{0.7}{±0}}              & 0.559{\scalebox{0.7}{±0.008}} & 0.559{\scalebox{0.7}{±0.008}}              & 0.542{\scalebox{0.7}{±0}} & 0.542{\scalebox{0.7}{±0}}              & 0.538{\scalebox{0.7}{±0}} & \textbf{0.539{\scalebox{0.7}{±0.004}}} & 0.57{\scalebox{0.7}{±0.016}}  & 0.553{\scalebox{0.7}{±0.025}} \\
Annthyroid       & 0.852{\scalebox{0.7}{±0}} & \textbf{0.906{\scalebox{0.7}{±0.006}}} & 0.911{\scalebox{0.7}{±0.022}} & \textbf{0.925{\scalebox{0.7}{±0.018}}} & 0.555{\scalebox{0.7}{±0}} & \textbf{0.667{\scalebox{0.7}{±0.013}}} & 0.785{\scalebox{0.7}{±0}} & 0.745{\scalebox{0.7}{±0.013}}              & 0.924{\scalebox{0.7}{±0.007}} & \textbf{0.949{\scalebox{0.7}{±0.013}}}                    \\
Arrhythmia       & 0.768{\scalebox{0.7}{±0}} & \textbf{0.776{\scalebox{0.7}{±0.015}}} & 0.773{\scalebox{0.7}{±0.01}}  & \textbf{0.795{\scalebox{0.7}{±0.022}}} & 0.769{\scalebox{0.7}{±0}} & \textbf{0.774{\scalebox{0.7}{±0.011}}} & 0.72{\scalebox{0.7}{±0.007}}  & \textbf{0.74{\scalebox{0.7}{±0.005}}}       & 0.774{\scalebox{0.7}{±0.013}} & \textbf{0.832{\scalebox{0.7}{±0.007}}}      \\
Breastw          & 0.994{\scalebox{0.7}{±0}} & \textbf{0.996{\scalebox{0.7}{±0.001}}} & 0.972{\scalebox{0.7}{±0}}     & \textbf{0.995{\scalebox{0.7}{±0.002}}} & 0.994{\scalebox{0.7}{±0}} & \textbf{0.996{\scalebox{0.7}{±0}}}         & 0.965{\scalebox{0.7}{±0}}     & \textbf{0.985{\scalebox{0.7}{±0.021}}}      & 0.997{\scalebox{0.7}{±0.003}} & 0.997{\scalebox{0.7}{±0}}                   \\
Campaign         & 0.771{\scalebox{0.7}{±0}} & 0.771{\scalebox{0.7}{±0}}              & 0.727{\scalebox{0.7}{±0.005}} & \textbf{0.729{\scalebox{0.7}{±0.005}}} & 0.763{\scalebox{0.7}{±0}} & 0.689{\scalebox{0.7}{±0}}                  & 0.756{\scalebox{0.7}{±0}}     & \textbf{0.775{\scalebox{0.7}{±0.001}}}      & 0.771{\scalebox{0.7}{±0.004}} & 0.753{\scalebox{0.7}{±0.004}}               \\
Cardio           & 0.966{\scalebox{0.7}{±0}} & 0.958{\scalebox{0.7}{±0.006}}              & 0.922{\scalebox{0.7}{±0.017}} & 0.919{\scalebox{0.7}{±0.015}}              & 0.965{\scalebox{0.7}{±0}} & 0.954{\scalebox{0.7}{±0.002}}              & 0.637{\scalebox{0.7}{±0}}     & \textbf{0.931{\scalebox{0.7}{±0.005}}}      & 0.956{\scalebox{0.7}{±0.009}} & \textbf{0.968{\scalebox{0.7}{±0.015}}}      \\
Cardiotocography & 0.789{\scalebox{0.7}{±0}} & \textbf{0.804{\scalebox{0.7}{±0.007}}} & 0.725{\scalebox{0.7}{±0.028}} & \textbf{0.768{\scalebox{0.7}{±0.036}}} & 0.752{\scalebox{0.7}{±0}} & \textbf{0.761{\scalebox{0.7}{±0.034}}}     & 0.789{\scalebox{0.7}{±0}}     & 0.778{\scalebox{0.7}{±0.008}}              & 0.84{\scalebox{0.7}{±0}}      & \textbf{0.853{\scalebox{0.7}{±0.029}}}      \\
Comm.and.crime   & 0.787{\scalebox{0.7}{±0}} & 0.779{\scalebox{0.7}{±0.002}}              & 0.804{\scalebox{0.7}{±0.014}} & \textbf{0.807{\scalebox{0.7}{±0.01}}}      & 0.705{\scalebox{0.7}{±0}} & \textbf{0.713{\scalebox{0.7}{±0.035}}}     & 0.517{\scalebox{0.7}{±0}}     & \textbf{0.525{\scalebox{0.7}{±0.031}}}      & 0.846{\scalebox{0.7}{±0.008}} & \textbf{0.894{\scalebox{0.7}{±0.007}}}      \\
Fault            & 0.559{\scalebox{0.7}{±0}} & \textbf{0.592{\scalebox{0.7}{±0.005}}} & 0.561{\scalebox{0.7}{±0.018}} & 0.558{\scalebox{0.7}{±0.018}}              & 0.568{\scalebox{0.7}{±0}} & 0.561{\scalebox{0.7}{±0.006}}              & 0.504{\scalebox{0.7}{±0}}     & \textbf{0.533{\scalebox{0.7}{±0.005}}}      & 0.628{\scalebox{0.7}{±0}}     & \textbf{0.698{\scalebox{0.7}{±0.05}}}       \\
Glass            & 0.548{\scalebox{0.7}{±0}} & \textbf{0.62{\scalebox{0.7}{±0.03}}}       & 0.577{\scalebox{0.7}{±0.014}} & \textbf{0.624{\scalebox{0.7}{±0.027}}} & 0.548{\scalebox{0.7}{±0}} & \textbf{0.644{\scalebox{0.7}{±0.015}}} & 0.624{\scalebox{0.7}{±0}}     & 0.425{\scalebox{0.7}{±0.073}}              & 0.697{\scalebox{0.7}{±0}}     & \textbf{0.906{\scalebox{0.7}{±0.044}}}      \\
Hepatitis        & 0.812{\scalebox{0.7}{±0}} & \textbf{0.835{\scalebox{0.7}{±0.022}}} & 0.726{\scalebox{0.7}{±0.02}}  & \textbf{0.778{\scalebox{0.7}{±0.021}}} & 0.496{\scalebox{0.7}{±0}} & \textbf{0.513{\scalebox{0.7}{±0.007}}} & 0.699{\scalebox{0.7}{±0}}     & \textbf{0.707{\scalebox{0.7}{±0.052}}}      & 0.812{\scalebox{0.7}{±0.038}} & \textbf{0.869{\scalebox{0.7}{±0.035}}}      \\
Imgseg           & 0.674{\scalebox{0.7}{±0}} & \textbf{0.708{\scalebox{0.7}{±0.031}}} & 0.686{\scalebox{0.7}{±0.021}} & \textbf{0.833{\scalebox{0.7}{±0.049}}} & 0.741{\scalebox{0.7}{±0}} & \textbf{0.778{\scalebox{0.7}{±0.016}}} & 0.624{\scalebox{0.7}{±0}}     & \textbf{0.813{\scalebox{0.7}{±0.024}}}      & 0.909{\scalebox{0.7}{±0.009}} & \textbf{0.915{\scalebox{0.7}{±0.011}}}      \\
Ionosphere       & 0.877{\scalebox{0.7}{±0}} & \textbf{0.911{\scalebox{0.7}{±0.026}}} & 0.968{\scalebox{0.7}{±0.006}} & 0.844{\scalebox{0.7}{±0.031}}              & 0.877{\scalebox{0.7}{±0}} & 0.839{\scalebox{0.7}{±0.005}}              & 0.957{\scalebox{0.7}{±0}}     & 0.873{\scalebox{0.7}{±0.008}}              & 0.984{\scalebox{0.7}{±0.015}} & 0.978{\scalebox{0.7}{±0.007}}               \\
Lympho           & 1{\scalebox{0.7}{±0}}     & 0.999{\scalebox{0.7}{±0.003}}              & 0.995{\scalebox{0.7}{±0.003}} & \textbf{0.998{\scalebox{0.7}{±0.003}}} & 0.981{\scalebox{0.7}{±0}} & \textbf{0.992{\scalebox{0.7}{±0.004}}} & 0.991{\scalebox{0.7}{±0}}     & \textbf{0.997{\scalebox{0.7}{±0.003}}}      & 1{\scalebox{0.7}{±0.002}}     & 0.991{\scalebox{0.7}{±0.003}}               \\
Mammography      & 0.899{\scalebox{0.7}{±0}} & \textbf{0.907{\scalebox{0.7}{±0.002}}} & 0.822{\scalebox{0.7}{±0.005}} & \textbf{0.903{\scalebox{0.7}{±0.003}}} & 0.9{\scalebox{0.7}{±0}}   & 0.897{\scalebox{0.7}{±0.005}}              & 0.825{\scalebox{0.7}{±0}}     & \textbf{0.907{\scalebox{0.7}{±0.001}}}      & 0.946{\scalebox{0.7}{±0.012}} & 0.883{\scalebox{0.7}{±0.013}}               \\
Mnist            & 0.902{\scalebox{0.7}{±0}} & 0.902{\scalebox{0.7}{±0}}                  & 0.862{\scalebox{0.7}{±0.006}} & \textbf{0.876{\scalebox{0.7}{±0.005}}} & 0.5{\scalebox{0.7}{±0}}   & \textbf{0.508{\scalebox{0.7}{±0.008}}} & 0.749{\scalebox{0.7}{±0.011}} & \textbf{0.769{\scalebox{0.7}{±0.002}}}      & 0.971{\scalebox{0.7}{±0.004}} & \textbf{0.974{\scalebox{0.7}{±0.004}}}      \\
Musk             & 1{\scalebox{0.7}{±0}}     & 1{\scalebox{0.7}{±0}}                      & 0.952{\scalebox{0.7}{±0.024}} & \textbf{1{\scalebox{0.7}{±0}}}             & 0.5{\scalebox{0.7}{±0}}   & \textbf{0.986{\scalebox{0.7}{±0.009}}} & 0.999{\scalebox{0.7}{±0}}     & \textbf{1{\scalebox{0.7}{±0}}}              & 1{\scalebox{0.7}{±0}}         & 1{\scalebox{0.7}{±0}}                            \\
Optdigits        & 0.582{\scalebox{0.7}{±0}} & 0.582{\scalebox{0.7}{±0}}                  & 0.824{\scalebox{0.7}{±0.026}} & \textbf{0.837{\scalebox{0.7}{±0.032}}} & 0.634{\scalebox{0.7}{±0}} & \textbf{0.997{\scalebox{0.7}{±0}}}         & 0.615{\scalebox{0.7}{±0.009}} & \textbf{0.635{\scalebox{0.7}{±0.001}}}      & 0.997{\scalebox{0.7}{±0}}     & 0.99{\scalebox{0.7}{±0.005}}                \\
Parkinson        & 0.693{\scalebox{0.7}{±0}} & \textbf{0.729{\scalebox{0.7}{±0.013}}} & 0.768{\scalebox{0.7}{±0.016}} & 0.74{\scalebox{0.7}{±0.035}}               & 0.604{\scalebox{0.7}{±0}} & \textbf{0.67{\scalebox{0.7}{±0.058}}}      & 0.515{\scalebox{0.7}{±0}}     & 0.472{\scalebox{0.7}{±0.071}}              & 0.673{\scalebox{0.7}{±0.035}} & \textbf{0.832{\scalebox{0.7}{±0.026}}}      \\
Pendigits        & 0.944{\scalebox{0.7}{±0}} & \textbf{0.949{\scalebox{0.7}{±0.005}}} & 0.967{\scalebox{0.7}{±0.003}} & \textbf{0.975{\scalebox{0.7}{±0.005}}} & 0.964{\scalebox{0.7}{±0}} & 0.951{\scalebox{0.7}{±0.001}}              & 0.93{\scalebox{0.7}{±0}}      & \textbf{0.945{\scalebox{0.7}{±0.006}}}      & 0.998{\scalebox{0.7}{±0.019}} & 0.994{\scalebox{0.7}{±0.002}}               \\
Pima             & 0.713{\scalebox{0.7}{±0}} & \textbf{0.739{\scalebox{0.7}{±0.004}}} & 0.674{\scalebox{0.7}{±0.006}} & \textbf{0.74{\scalebox{0.7}{±0.017}}}      & 0.713{\scalebox{0.7}{±0}} & 0.627{\scalebox{0.7}{±0.032}}              & 0.583{\scalebox{0.7}{±0}}     & 0.552{\scalebox{0.7}{±0.014}}              & 0.765{\scalebox{0.7}{±0.015}} & \textbf{0.782{\scalebox{0.7}{±0.009}}}      \\
Satellite        & 0.666{\scalebox{0.7}{±0}} & \textbf{0.678{\scalebox{0.7}{±0.001}}} & 0.803{\scalebox{0.7}{±0.007}} & 0.754{\scalebox{0.7}{±0.014}}              & 0.666{\scalebox{0.7}{±0}} & \textbf{0.799{\scalebox{0.7}{±0.007}}} & 0.788{\scalebox{0.7}{±0}}     & 0.788{\scalebox{0.7}{±0.006}}              & 0.832{\scalebox{0.7}{±0.008}} & \textbf{0.873{\scalebox{0.7}{±0.013}}}      \\
Satimage-2       & 0.982{\scalebox{0.7}{±0}} & \textbf{0.984{\scalebox{0.7}{±0.001}}} & 0.994{\scalebox{0.7}{±0.002}} & 0.989{\scalebox{0.7}{±0.002}}              & 0.982{\scalebox{0.7}{±0}} & \textbf{0.984{\scalebox{0.7}{±0.004}}} & 0.965{\scalebox{0.7}{±0}}     & \textbf{0.993{\scalebox{0.7}{±0.001}}}      & 0.997{\scalebox{0.7}{±0.004}} & \textbf{0.998{\scalebox{0.7}{±0.001}}}              \\
Shuttle          & 0.994{\scalebox{0.7}{±0}} & \textbf{0.998{\scalebox{0.7}{±0.001}}} & 0.996{\scalebox{0.7}{±0.001}} & \textbf{0.999{\scalebox{0.7}{±0.001}}} & 0.997{\scalebox{0.7}{±0}} & \textbf{0.999{\scalebox{0.7}{±0}}}         & 0.998{\scalebox{0.7}{±0}}     & 0.998{\scalebox{0.7}{±0}}                  & 0.999{\scalebox{0.7}{±0}}     & \textbf{1{\scalebox{0.7}{±0}}}              \\
SpamBase         & 0.814{\scalebox{0.7}{±0}} & \textbf{0.815{\scalebox{0.7}{±0.002}}} & 0.858{\scalebox{0.7}{±0.009}} & 0.857{\scalebox{0.7}{±0.007}}              & 0.768{\scalebox{0.7}{±0}} & 0.768{\scalebox{0.7}{±0}}                  & 0.688{\scalebox{0.7}{±0}}     & \textbf{0.703{\scalebox{0.7}{±0.008}}}      & 0.829{\scalebox{0.7}{±0.008}} & \textbf{0.912{\scalebox{0.7}{±0.019}}}      \\
Speech           & 0.364{\scalebox{0.7}{±0}} & \textbf{0.402{\scalebox{0.7}{±0.014}}} & 0.381{\scalebox{0.7}{±0.014}} & \textbf{0.4{\scalebox{0.7}{±0.008}}}       & 0.367{\scalebox{0.7}{±0}} & \textbf{0.385{\scalebox{0.7}{±0.008}}} & 0.36{\scalebox{0.7}{±0}}      & \textbf{0.394{\scalebox{0.7}{±0.037}}}      & 0.582{\scalebox{0.7}{±0.031}} & 0.577{\scalebox{0.7}{±0.035}}               \\
Thyroid          & 0.986{\scalebox{0.7}{±0}} & 0.985{\scalebox{0.7}{±0.002}}              & 0.927{\scalebox{0.7}{±0.001}} & \textbf{0.995{\scalebox{0.7}{±0.002}}} & 0.986{\scalebox{0.7}{±0}} & 0.956{\scalebox{0.7}{±0.008}}              & 0.883{\scalebox{0.7}{±0}}     & \textbf{0.972{\scalebox{0.7}{±0.009}}}      & 0.991{\scalebox{0.7}{±0.003}} & \textbf{0.987{\scalebox{0.7}{±0.003}}}      \\
Vertebral        & 0.175{\scalebox{0.7}{±0}} & \textbf{0.192{\scalebox{0.7}{±0.015}}} & 0.144{\scalebox{0.7}{±0.011}} & \textbf{0.423{\scalebox{0.7}{±0.101}}} & 0.265{\scalebox{0.7}{±0}} & \textbf{0.574{\scalebox{0.7}{±0.062}}} & 0.412{\scalebox{0.7}{±0}}     & \textbf{0.65{\scalebox{0.7}{±0.038}}}       & 0.622{\scalebox{0.7}{±0.057}} & \textbf{0.763{\scalebox{0.7}{±0.075}}}      \\
Vowels           & 0.523{\scalebox{0.7}{±0}} & \textbf{0.559{\scalebox{0.7}{±0.011}}} & 0.59{\scalebox{0.7}{±0.031}}  & \textbf{0.727{\scalebox{0.7}{±0.034}}} & 0.756{\scalebox{0.7}{±0}} & 0.716{\scalebox{0.7}{±0.017}}              & 0.615{\scalebox{0.7}{±0}}     & \textbf{0.766{\scalebox{0.7}{±0.022}}}      & 0.851{\scalebox{0.7}{±0.004}} & \textbf{0.987{\scalebox{0.7}{±0.003}}}      \\
Wbc              & 0.967{\scalebox{0.7}{±0}} & 0.966{\scalebox{0.7}{±0.001}}              & 0.972{\scalebox{0.7}{±0.003}} & 0.953{\scalebox{0.7}{±0.018}}              & 0.967{\scalebox{0.7}{±0}} & 0.952{\scalebox{0.7}{±0.009}}              & 0.875{\scalebox{0.7}{±0}}     & \textbf{0.968{\scalebox{0.7}{±0.008}}}      & 0.996{\scalebox{0.7}{±0.008}} & 0.988{\scalebox{0.7}{±0.003}}               \\
WDBC             & 0.999{\scalebox{0.7}{±0}} & 0.999{\scalebox{0.7}{±0.001}}              & 0.998{\scalebox{0.7}{±0.002}} & \textbf{1{\scalebox{0.7}{±0.001}}}         & 0.964{\scalebox{0.7}{±0}} & \textbf{0.994{\scalebox{0.7}{±0.002}}} & 0.979{\scalebox{0.7}{±0}}     & \textbf{0.993{\scalebox{0.7}{±0.006}}}      & 1{\scalebox{0.7}{±0.001}}     & 1{\scalebox{0.7}{±0}}                       \\
Wilt             & 0.261{\scalebox{0.7}{±0}} & \textbf{0.474{\scalebox{0.7}{±0.06}}}      & 0.46{\scalebox{0.7}{±0.027}}  & \textbf{0.765{\scalebox{0.7}{±0.03}}}      & 0.792{\scalebox{0.7}{±0}} & 0.762{\scalebox{0.7}{±0.022}}              & 0.375{\scalebox{0.7}{±0}}     & \textbf{0.801{\scalebox{0.7}{±0.026}}}      & 0.936{\scalebox{0.7}{±0.007}} & 0.919{\scalebox{0.7}{±0.011}}               \\
Wine             & 0.447{\scalebox{0.7}{±0}} & \textbf{0.933{\scalebox{0.7}{±0.015}}} & 0.657{\scalebox{0.7}{±0.095}} & \textbf{0.748{\scalebox{0.7}{±0.065}}} & 0.485{\scalebox{0.7}{±0}} & \textbf{0.669{\scalebox{0.7}{±0.221}}} & 0.743{\scalebox{0.7}{±0}}     & 0.648{\scalebox{0.7}{±0.11}}               & 1{\scalebox{0.7}{±0.007}}     & 1{\scalebox{0.7}{±0}}            \\
WPBC             & 0.469{\scalebox{0.7}{±0}} & \textbf{0.471{\scalebox{0.7}{±0.012}}} & 0.498{\scalebox{0.7}{±0.011}} & \textbf{0.502{\scalebox{0.7}{±0.006}}} & 0.47{\scalebox{0.7}{±0}}  & \textbf{0.472{\scalebox{0.7}{±0.018}}} & 0.471{\scalebox{0.7}{±0}}     & \textbf{0.517{\scalebox{0.7}{±0.024}}}      & 0.62{\scalebox{0.7}{±0.045}}  & \textbf{0.67{\scalebox{0.7}{±0.031}}}       \\
Yeast            & 0.432{\scalebox{0.7}{±0}} & \textbf{0.456{\scalebox{0.7}{±0.005}}} & 0.41{\scalebox{0.7}{±0.007}}  & \textbf{0.526{\scalebox{0.7}{±0.024}}} & 0.448{\scalebox{0.7}{±0}} & \textbf{0.464{\scalebox{0.7}{±0.002}}} & 0.446{\scalebox{0.7}{±0}}     & \textbf{0.526{\scalebox{0.7}{±0.031}}}      & 0.512{\scalebox{0.7}{±0.015}} & 0.49{\scalebox{0.7}{±0.017}}                \\ 
\midrule 
Average AUC-ROC  & 0.735{\scalebox{0.7}{±0}} & \textbf{0.767{\scalebox{0.7}{±0.009}}} & 0.755{\scalebox{0.7}{±0.014}} & \textbf{0.794{\scalebox{0.7}{±0.019}}} & 0.714{\scalebox{0.7}{±0}} & \textbf{0.758{\scalebox{0.7}{±0.018}}} & 0.706{\scalebox{0.7}{±0.001}} & \textbf{0.745{\scalebox{0.7}{±0.019}}}      & 0.851{\scalebox{0.7}{±0.012}} & \textbf{0.878{\scalebox{0.7}{±0.015}}}          
  \\ \bottomrule       
\end{tabular}}
\end{table}

\subsection{Full Results of Comparison with Other Baseline Methods}
\label{appendix:full comparison with other baseline}
We further compare the performance of LLM-DAS with conventional and advanced TAD baselines. The full results are detailed in Table~\ref{appendix:methodsPR} and Table~\ref{appendix:methodsROC}. Since LLM-DAS and  the recent finetune-based AnoLLM~\citep{tsai2025anollm} have some shared datasets, we cite the results of AnoLLM due to its high computational cost, and perform other methods on the shared subsets.

\begin{sidewaystable}
\tiny
\centering
\caption{Comparison of AUC-PR ($\uparrow$) results between other baseline methods and LLM-DAS.}
\label{appendix:methodsPR}
\setlength{\tabcolsep}{1pt}
\resizebox{\linewidth}{!}{
\begin{tabular}{lccccccccccccccc|cc}
\toprule

                & OCSVM   & KNN     & LOF    & PCA     & IForest & ECOD    & DeepSVDD & AutoEncoder & ICL     & DTE    & MCM    & NPT-AD & DRL    & AnoLLM(135M) & AnoLLM(360M) & \textbf{Ours(PCA)}   & \textbf{Ours(DRL)}     \\ \midrule         
Annthyroid      & 0.1831  & 0.3525  & 0.4513 & 0.5657  & 0.6149  & 0.4002  & 0.3235   & 0.5291      & 0.4114  & 0.6288 & 0.3215 & 0.62   & 0.6761 & 0.631        & 0.648        & 0.6602 & 0.7618           \\
Arrhythmia      & 0.5339  & 0.6008  & 0.5277 & 0.5336  & 0.5097  & 0.4461  & 0.6036   & 0.3029      & 0.6155  & 0.4912 & 0.6107 & 0.4345 & 0.627  & 0.636        & 0.642        & 0.5508 & 0.7298           \\
Breastw         & 0.9934  & 0.9712  & 0.9923 & 0.9934  & 0.9449  & 0.9522  & 0.9924   & 0.9896      & 0.9656  & 0.8825 & 0.9952 & 0.9813 & 0.9966 & 0.991        & 0.992        & 0.9956 & 0.9972           \\
Glass           & 0.0896  & 0.1099  & 0.0923 & 0.0896  & 0.0952  & 0.1113  & 0.0912   & 0.1079      & 0.2573  & 0.2151 & 0.1905 & 0.2204 & 0.167  & 0.247        & 0.234        & 0.1278 & 0.4934           \\
Ionosphere      & 0.8969  & 0.9297  & 0.9591 & 0.8969  & 0.9768  & 0.9713  & 0.867    & 0.7328      & 0.9777  & 0.9683 & 0.9802 & 0.9812 & 0.9895 & 0.933        & 0.932        & 0.926  & 0.9846           \\
Lympho          & 0.8107  & 0.9401  & 0.9762 & 1       & 0.9593  & 0.8972  & 0.9749   & 0.2709      & 0.6091  & 0.8677 & 0.4204 & 0.9929 & 1      & 0.856        & 0.938        & 0.9889 & 0.8706           \\
Mammography     & 0.4178  & 0.381   & 0.4063 & 0.4165  & 0.3334  & 0.538   & 0.419    & 0.253       & 0.1894  & 0.3985 & 0.4755 & 0.364  & 0.8406 & 0.592        & 0.364        & 0.5793 & 0.5288           \\
Musk            & 0.0614  & 0.9917  & 1      & 1       & 0.5279  & 0.982   & 1        & 1           & 1       & 1      & 0.639  & 1      & 1      & 1            & 1            & 0.9969 & 1                \\
Optdigits       & 0.0692  & 0.8589  & 0.4363 & 0.0602  & 0.157   & 0.0669  & 0.1159   & 0.1418      & 0.1696  & 0.1534 & 0.8885 & 0.4203 & 0.9356 & 0.75         & 0.398        & 0.0602 & 0.9566           \\
Pendigits       & 0.5178  & 0.9692  & 0.7855 & 0.3863  & 0.5133  & 0.4145  & 0.0616   & 0.8904      & 0.4039  & 0.4844 & 0.8258 & 0.9671 & 0.936  & 0.623        & 0.554        & 0.4847 & 0.8848           \\
Pima            & 0.7008  & 0.7098  & 0.697  & 0.7008  & 0.6662  & 0.5877  & 0.7165   & 0.7174      & 0.6965  & 0.6798 & 0.7389 & 0.7527 & 0.7449 & 0.677        & 0.674        & 0.7509 & 0.7912           \\
Satellite       & 0.7778  & 0.8515  & 0.8088 & 0.7778  & 0.8583  & 0.8334  & 0.8217   & 0.8218      & 0.8799  & 0.8479 & 0.8532 & 0.8576 & 0.8692 & 0.91         & 0.891        & 0.7921 & 0.8942           \\
Satimage-2      & 0.9192  & 0.9555  & 0.9692 & 0.9192  & 0.8846  & 0.7775  & 0.9427   & 0.9688      & 0.8124  & 0.6821 & 0.985  & 0.9862 & 0.9703 & 0.988        & 0.974        & 0.9311 & 0.9804           \\
Shuttle         & 0.9488  & 0.937   & 0.9601 & 0.9627  & 0.9172  & 0.9815  & 0.9818   & 0.9316      & 0.9811  & 0.9403 & 0.9479 & 0.9151 & 0.9819 & 0.997        & 0.996        & 0.9906 & 0.9991           \\
Speech          & 0.0279  & 0.0197  & 0.0315 & 0.0277  & 0.0353  & 0.0287  & 0.04     & 0.027       & 0.0335  & 0.0285 & 0.038  & 0.0584 & 0.0584 & 0.036        & 0.037        & 0.0293 & 0.1189           \\
Thyroid         & 0.8134  & 0.5903  & 0.7892 & 0.8134  & 0.6055  & 0.6807  & 0.7282   & 0.8096      & 0.6575  & 0.8167 & 0.8417 & 0.8179 & 0.8626 & 0.696        & 0.74         & 0.8146 & 0.877            \\
Vertebral       & 0.1517  & 0.1239  & 0.2063 & 0.1381  & 0.1342  & 0.1917  & 0.159    & 0.1476      & 0.1598  & 0.2514 & 0.1949 & 0.2279 & 0.2854 & 0.289        & 0.181        & 0.1433 & 0.5657           \\
Vowels          & 0.2969  & 0.3146  & 0.3277 & 0.1051  & 0.0984  & 0.1772  & 0.1717   & 0.3475      & 0.1574  & 0.381  & 0.0977 & 0.9498 & 0.4506 & 0.839        & 0.599        & 0.109  & 0.8918           \\
Wbc             & 0.8391  & 0.8022  & 0.8412 & 0.8391  & 0.8573  & 0.7217  & 0.834    & 0.8578      & 0.7218  & 0.3997 & 0.8887 & 0.8079 & 0.9742 & 0.873        & 0.753        & 0.84   & 0.9341           \\
Wine            & 0.1424  & 0.9917  & 0.1253 & 0.1325  & 0.2458  & 0.3578  & 0.1476   & 0.1605      & 0.5659  & 0.9985 & 0.9335 & 0.7746 & 1      & 0.522        & 0.529        & 0.8459 & 1                \\
Yeast           & 0.4803  & 0.4737  & 0.4866 & 0.4678  & 0.4654  & 0.4943  & 0.4953   & 0.4833      & 0.5097  & 0.4974 & 0.4631 & 0.4888 & 0.5416 & 0.301        & 0.302        & 0.5053 & 0.5281           \\ \midrule
Average AUC-PR  & 0.5082  & 0.6607  & 0.6129 & 0.5632  & 0.5429  & 0.5529  & 0.547    & 0.5472      & 0.5607  & 0.6006 & 0.6348 & 0.6961 & 0.7575 & 0.6851       & 0.637        & 0.6249 & \textbf{0.7994}  \\
Average ranking & 12.1429 & 10.7143 & 9.3810 & 11.6429 & 12.1429 & 11.7619 & 10.2381  & 11.2381     & 10.0952 & 10     & 7.9048 & 6.7619 & 3.1667 & 6.619        & 8.119        & 8.5714 & \textbf{2.5}     \\
Win             & 0       & 1       & 1      & 2       & 0       & 0       & 1        & 1           & 1       & 1      & 0      & 2      & 7      & 3            & 1            & 0      & \textbf{12}  \\ \bottomrule    
\end{tabular}}
\end{sidewaystable}

\begin{sidewaystable}
\tiny
\centering
\caption{Comparison of AUC-ROC ($\uparrow$) results between other baseline methods and LLM-DAS.}
\label{appendix:methodsROC}
\setlength{\tabcolsep}{1pt}
\resizebox{\linewidth}{!}{
\begin{tabular}{lccccccccccccccc|cc}
\toprule

                 & OCSVM       & KNN         & LOF         & PCA         & IForest     & ECOD        & DeepSVDD    & AutoEncoder & ICL         & DTE         & MCM         & NPT-AD      & DRL         & AnoLLM(135M) & AnoLLM(360M) & \textbf{Ours(PCA)}   & \textbf{Ours(DRL)}            \\ \midrule
Annthyroid  & 0.5551  & 0.6903  & 0.7216 & 0.8519  & 0.9112  & 0.7845  & 0.5678   & 0.7295      & 0.6997  & 0.909  & 0.6894 & 0.8682 & 0.9239 & 0.927        & 0.931        & 0.9065    & 0.9492           \\
Arrhythmia  & 0.7689  & 0.7933  & 0.7688 & 0.7684  & 0.7734  & 0.7199  & 0.7941   & 0.5682      & 0.8145  & 0.5912 & 0.8114 & 0.7185 & 0.7742 & 0.825        & 0.822        & 0.7757    & 0.8316           \\
Breastw     & 0.9938  & 0.9714  & 0.9937 & 0.9938  & 0.9719  & 0.9649  & 0.9925   & 0.9909      & 0.9725  & 0.9278 & 0.9955 & 0.9848 & 0.9968 & 0.992        & 0.993        & 0.6196    & 0.9972           \\
Glass       & 0.548   & 0.6141  & 0.562  & 0.548   & 0.5771  & 0.6235  & 0.5566   & 0.6088      & 0.835   & 0.6964 & 0.7225 & 0.7875 & 0.6969 & 0.819        & 0.797        & 0.9109    & 0.9061           \\
Ionosphere  & 0.8765  & 0.9167  & 0.9454 & 0.8765  & 0.9683  & 0.9569  & 0.8552   & 0.6231      & 0.971   & 0.9542 & 0.9726 & 0.9735 & 0.984  & 0.909        & 0.924        & 0.9986    & 0.9779           \\
Lympho      & 0.9812  & 0.987   & 0.9977 & 1       & 0.9945  & 0.9906  & 0.9977   & 0.7856      & 0.9546  & 0.9899 & 0.9257 & 0.9993 & 1      & 0.968        & 0.993        & 0.9074    & 0.9911           \\
Mammography & 0.9003  & 0.864   & 0.8922 & 0.8993  & 0.822   & 0.8251  & 0.8879   & 0.8472      & 0.6548  & 0.8862 & 0.9053 & 0.8873 & 0.9455 & 0.915        & 0.876        & 0.9998    & 0.8834           \\
Musk        & 0.5     & 0.9917  & 1      & 1       & 0.9521  & 0.9987  & 1        & 1           & 1       & 1      & 0.9752 & 1      & 1      & 1            & 1            & 0.5817    & 1                \\
Optdigits   & 0.6338  & 0.9862  & 0.9665 & 0.5817  & 0.8239  & 0.6145  & 0.7603   & 0.6694      & 0.787   & 0.8238 & 0.9947 & 0.9317 & 0.9972 & 0.983        & 0.939        & 0.9487    & 0.9902           \\
Pendigits   & 0.9636  & 0.9906  & 0.9905 & 0.9437  & 0.9666  & 0.9295  & 0.4563   & 0.9937      & 0.9142  & 0.9761 & 0.9919 & 0.9987 & 0.9979 & 0.971        & 0.964        & 0.7393    & 0.9941           \\
Pima        & 0.7133  & 0.6723  & 0.6913 & 0.7133  & 0.6737  & 0.5834  & 0.7348   & 0.7163      & 0.6727  & 0.6788 & 0.7639 & 0.7553 & 0.7651 & 0.663        & 0.654        & 0.6783    & 0.7818           \\
Satellite   & 0.6663  & 0.8139  & 0.7391 & 0.6663  & 0.8026  & 0.7884  & 0.7659   & 0.7233      & 0.8549  & 0.7661 & 0.7962 & 0.806  & 0.832  & 0.902        & 0.877        & 0.9837    & 0.8726           \\
Satimage-2  & 0.9817  & 0.9891  & 0.9961 & 0.9817  & 0.9938  & 0.965   & 0.9881   & 0.9979      & 0.9792  & 0.9967 & 0.9992 & 0.9995 & 0.9968 & 1            & 0.999        & 0.9977    & 0.9984           \\
Shuttle     & 0.9969  & 0.9893  & 0.9983 & 0.9936  & 0.9961  & 0.9978  & 0.9952   & 0.9944      & 0.9935  & 0.9993 & 0.9975 & 0.9931 & 0.9992 & 1            & 1            & 0.4020    & 0.9999           \\
Speech      & 0.3673  & 0.3561  & 0.3759 & 0.3638  & 0.3812  & 0.3596  & 0.5071   & 0.3633      & 0.4883  & 0.3817 & 0.4409 & 0.5925 & 0.5821 & 0.47         & 0.47         & 0.9849    & 0.5774           \\
Thyroid     & 0.9855  & 0.9525  & 0.9856 & 0.9855  & 0.9271  & 0.8827  & 0.9887   & 0.978       & 0.9518  & 0.9863 & 0.9804 & 0.9787 & 0.9911 & 0.975        & 0.983        & 0.1917    & 0.9873           \\
Vertebral   & 0.2654  & 0.1171  & 0.4317 & 0.1746  & 0.1444  & 0.4124  & 0.2706   & 0.2375      & 0.2821  & 0.543  & 0.3767 & 0.5347 & 0.6216 & 0.565        & 0.408        & 0.5591    & 0.7629           \\
Vowels      & 0.7557  & 0.8174  & 0.8564 & 0.5229  & 0.5902  & 0.6147  & 0.5734   & 0.7905      & 0.702   & 0.8142 & 0.6529 & 0.9935 & 0.8511 & 0.982        & 0.938        & 0.9656    & 0.9869           \\
Wbc         & 0.9667  & 0.9536  & 0.967  & 0.9667  & 0.9715  & 0.8747  & 0.9633   & 0.9737      & 0.908   & 0.8054 & 0.9814 & 0.9619 & 0.9963 & 0.964        & 0.952        & 0.9333    & 0.9883           \\
Wine        & 0.485   & 0.9917  & 0.4083 & 0.4467  & 0.6571  & 0.7433  & 0.5067   & 0.5356      & 0.915   & 0.9944 & 0.9538 & 0.9622 & 1      & 0.909        & 0.851        & 0.4557    & 1                \\
Yeast       & 0.4483  & 0.4366  & 0.4571 & 0.4324  & 0.4095  & 0.4464  & 0.479    & 0.4503      & 0.5076  & 0.4458 & 0.4259 & 0.4691 & 0.512  & 0.744        & 0.73         & 0.7674    & 0.4905           \\ \midrule
Average     & 0.7311  & 0.8045  & 0.7974 & 0.7481  & 0.7766  & 0.7655  & 0.7448   & 0.7418      & 0.8028  & 0.8174 & 0.8263 & 0.8665 & 0.8792 & 0.8801       & 0.8620       & 0.7766    & \textbf{0.9032}  \\
Rank        & 11.8095 & 11.3810 & 8.8333 & 11.5952 & 11.2857 & 12.7143 & 10.3095  & 11.2857     & 10.4762 & 9.3810 & 7.8571 & 6.8571 & 3.7143 & 6.2381       & 7.1905       & 8.6667    & \textbf{3.4048}  \\
Win         & 0       & 0       & 1      & 2       & 0       & 0       & 1        & 1           & 1       & 1      & 0      & 3      & 6      & 3            & 2            & 6         & \textbf{7}   \\ \bottomrule    
\end{tabular}}
\end{sidewaystable}

\clearpage
\subsection{Varying LLM backbones}
\label{appendix:various LLM}
To investigate the impact of using different LLMs in our framework, given their distinct prior knowledge and reasoning abilities from being trained on various text corpora, we measured performance using not only  Gemini-2.5-Pro but also the GPT-4o and Qwen3 as a backbone for comparison. 
Though the performance of LLM-DAS on Qwen3 slightly drops compared to Gemini-2.5-pro and GPT-4o, it still surpasses the base detector by a large margin. The results clearly show that LLM-DAS maintains its strong performance across all three LLMs.

\begin{table}[h!]
\centering
\caption{Effects of various LLM backbones. The AUC-PR results are averaged across all datasets. Here, we use OCSVM as the base detector.}
\label{appendix:tab_variousllm}
\scriptsize
\begin{tabular}{ccccc}
\toprule
               & OCSVM  & +LLM-DAS (Qwen3) & +LLM-DAS (GPT-4o) & +LLM-DAS (Gemini-2.5-Pro)  \\ \midrule
Abalone          & 0.8459 & 0.8460                              & 0.8465           & 0.8487                    \\
Amazon           & 0.1050 & 0.1048                              & 0.1058           & 0.1061                    \\
Annthyroid       & 0.1831 & 0.4490                              & 0.4532           & 0.2483                    \\
Arrhythmia       & 0.5339 & 0.5898                              & 0.5965           & 0.6086                    \\
Breastw          & 0.9934 & 0.9966                              & 0.9969           & 0.9954                    \\
Campaign         & 0.4749 & 0.4417                              & 0.4409           & 0.4349                    \\
Cardio           & 0.8614 & 0.8625                              & 0.8551           & 0.8083                    \\
Cardiotocography & 0.6619 & 0.6244                              & 0.6773           & 0.7029                    \\
Comm.and.crime   & 0.8371 & 0.8360                              & 0.8394           & 0.8434                    \\
Fault            & 0.6062 & 0.5597                              & 0.5637           & 0.5417                    \\
Glass            & 0.0896 & 0.2215                              & 0.2204           & 0.2006                    \\
Hepatitis        & 0.2815 & 0.4196                              & 0.4481           & 0.3002                    \\
Imgseg           & 0.7883 & 0.7970                              & 0.8056           & 0.8592                    \\
Ionosphere       & 0.8969 & 0.8212                              & 0.8211           & 0.8384                    \\
Lympho           & 0.8107 & 0.6816                              & 0.7346           & 0.9337                    \\
Mammography      & 0.4178 & 0.4181                              & 0.4440           & 0.5359                    \\
Mnist            & 0.1686 & 0.4366                              & 0.4573           & 0.1806                    \\
Musk             & 0.0614 & 0.9605                              & 0.9561           & 0.8676                    \\
Optdigits        & 0.0692 & 0.8242                              & 0.8456           & 0.9281                    \\
Parkinson        & 0.8892 & 0.9381                              & 0.9420           & 0.9200                    \\
Pendigits        & 0.5178 & 0.5584                              & 0.5563           & 0.4912                    \\
Pima             & 0.7008 & 0.6090                              & 0.6377           & 0.6514                    \\
Satellite        & 0.7778 & 0.8225                              & 0.8326           & 0.8319                    \\
Satimage-2       & 0.9192 & 0.7065                              & 0.8207           & 0.8996                    \\
Shuttle          & 0.9488 & 0.9955                              & 0.9958           & 0.9906                    \\
SpamBase         & 0.8136 & 0.8331                              & 0.8305           & 0.8138                    \\
Speech           & 0.0279 & 0.0275                              & 0.0278           & 0.0294                    \\
Thyroid          & 0.8134 & 0.3506                              & 0.3505           & 0.6326                    \\
Vertebral        & 0.1517 & 0.2292                              & 0.2336           & 0.2733                    \\
Vowels           & 0.2969 & 0.1860                              & 0.1843           & 0.2098                    \\
Wbc              & 0.8391 & 0.6796                              & 0.6811           & 0.7928                    \\
WDBC             & 0.4348 & 0.8900                              & 0.8733           & 0.8740                    \\
Wilt             & 0.2254 & 0.1914                              & 0.2619           & 0.2193                    \\
Wine             & 0.1424 & 0.4390                              & 0.4679           & 0.3335                    \\
WPBC             & 0.3974 & 0.4010                              & 0.3978           & 0.4065                    \\
Yeast            & 0.4803 & 0.4880                              & 0.4872           & 0.5148                    \\ \midrule
Average          & 0.5295 & 0.5899                              & 0.6025           & 0.6019                                      \\ 
\bottomrule
\end{tabular}
\end{table}

\subsection{Computational Efficiency.}
\label{appendix: time}
As shown in Table~\ref{table:time}, LLM-DAS remains computationally efficient in both training and inference stages. This efficiency comes from two key properties: (i) LLM-DAS does not require repeatedly querying or fine-tuning LLMs; instead, the detector-specific synthesis code is generated once per anomaly type and reused across datasets, and (ii) inference only involves an additional lightweight binary classifier on top of the source detector, introducing negligible overhead.

\begin{table}
\scriptsize
\centering
\caption{Runtime in seconds of baseline $f_t$ and LLM-DAS $F_t$ for the training and inference, averaged over all datasets. The training time of $f_t$ is not included in the LLM-DAS training time. LLM-DAS training time includes anomalies synthesis and binary classifier $\tilde{f}_t$ training, while excluding the two-time LLM querying cost for code generation, as the resulting synthesis code is reusable across any dataset.}
\label{table:time}
\begin{tabular}{cccccc}
\toprule
Stage           & PCA       & IForest       & OCSVM       & ECOD       & DRL \\ \midrule
Baseline Train  & 0.2891    & 0.8091        & 4.4825      & 0.1313     & 23.6555    \\
LLM-DAS Train   & 0.5186    & 1.1474        & 6.4229      & 0.1890     & 1.5253    \\
Baseline Test   & 0.2402    & 0.2391        & 1.8455      & 0.2727     & 0.0777    \\ 
LLM-DAS Test    & 0.3389    & 0.3089        & 2.5990      & 0.3207     & 0.2390    \\ \bottomrule
\end{tabular}
\end{table}

\subsection{Visualization and KDE Analysis of Synthetic Anomalies}
\label{appendix:visualization}
To further examine the quality of anomalies synthesized by LLM-DAS, we utilize T-SNE and Kernel Density Estimation (KDE) plots of anomaly scores on the Thyroid and Vertebral datasets.
\begin{figure}[th!]
    \vspace{-1em}
    \centering
    \includegraphics[width=\linewidth]{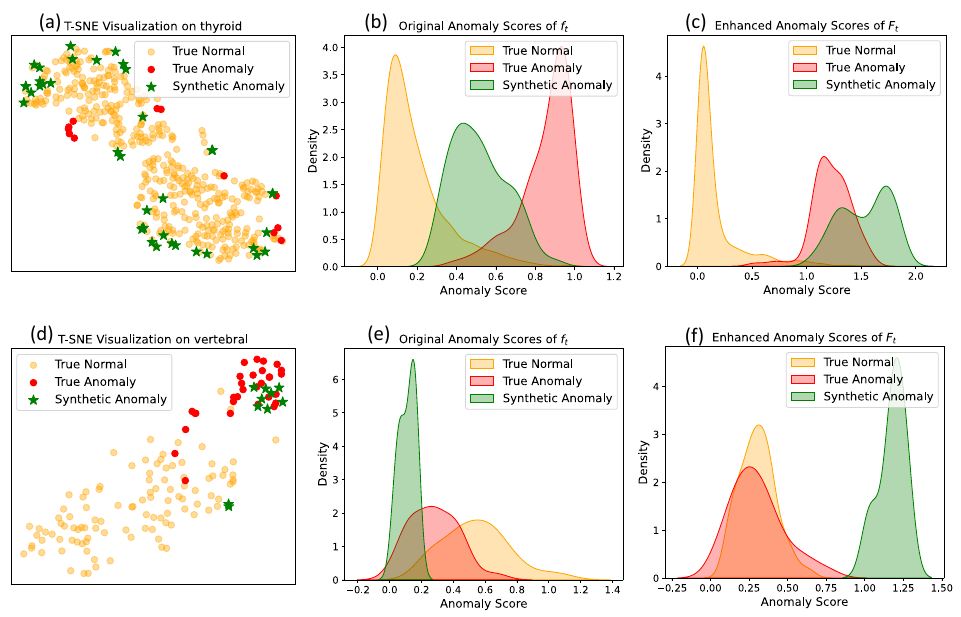}
    \caption{\small{Visualization of synthetic hard anomalies and score distributions  on Thyroid and Vertebral test dataset. (a, d) T-SNE plots of normal, real anomaly, and LLM-DAS–generated anomaly samples. (b, e) Kernel density estimation (KDE) of anomaly scores from the source detector $f_t$. (c, f) KDE of anomaly scores from the enhanced detector $F_t$.}}
    \label{fig:visualization_all}
    \vspace{-1em}
\end{figure}

\clearpage
\subsection{Sensitivity Analysis}
\label{appendix:sensitivity analysis}
We incorporate the sensitivity analysis for the number of synthetic samples (set to the ratio of training set size), performance robustness across different queries and the type of binary classifiers. We also verify the effectiveness of summing $f_t$ and $\tilde{f}_t$.

\begin{figure}[h!]
     \includegraphics[width=\linewidth]{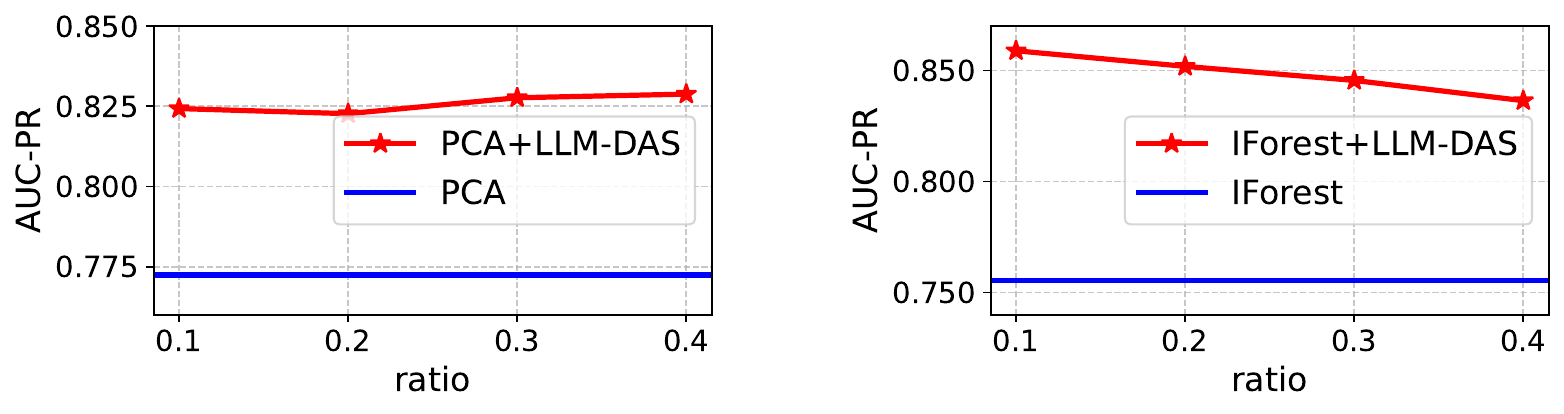}
    \captionsetup{font=small}
    \captionof{figure}{\label{fig:methods1} \small{AUC-PR-based performance comparison of different anomalous samples injection ratios on the Imgseg dataset, with results for PCA (left) and IForest (right).}}
    \vspace{-1em}
\end{figure}

\begin{table}[h!]
\centering
\caption{AUC-PR performance of PCA and PCA+LLM-DAS under different anomaly synthesis ratios.}
\scriptsize
\begin{tabular}{c|c|cccccccccc}
\toprule
                 & \multirow{2}{*}{\centering PCA}   & \multicolumn{10}{c}{LLM-DAS with various anomaly synthesis ratios}\\ \cline{3-12}
                 &       & 0.1   & 0.2   & 0.3   & 0.4   & 0.5   & 0.6   & 0.7   & 0.8   & 0.9   & 1     \\ \midrule
Abalone          & 0.839 & 0.841 & 0.841 & 0.841 & 0.841 & 0.842 & 0.841 & 0.842 & 0.841 & 0.841 & 0.841 \\
Amazon           & 0.107 & 0.108 & 0.108 & 0.108 & 0.107 & 0.108 & 0.108 & 0.107 & 0.108 & 0.108 & 0.108 \\
Annthyroid       & 0.566 & 0.660 & 0.701 & 0.690 & 0.698 & 0.698 & 0.695 & 0.712 & 0.692 & 0.691 & 0.688 \\
Arrhythmia       & 0.534 & 0.551 & 0.545 & 0.551 & 0.553 & 0.555 & 0.551 & 0.549 & 0.553 & 0.554 & 0.557 \\
Breastw          & 0.993 & 0.996 & 0.996 & 0.995 & 0.995 & 0.995 & 0.995 & 0.995 & 0.996 & 0.995 & 0.996 \\
Campaign         & 0.488 & 0.488 & 0.488 & 0.488 & 0.488 & 0.488 & 0.488 & 0.488 & 0.488 & 0.488 & 0.489 \\
Cardio           & 0.863 & 0.824 & 0.828 & 0.850 & 0.826 & 0.812 & 0.826 & 0.826 & 0.828 & 0.830 & 0.846 \\
Cardiotocography & 0.697 & 0.734 & 0.738 & 0.741 & 0.726 & 0.734 & 0.735 & 0.741 & 0.737 & 0.738 & 0.750 \\
Comm.and.crime   & 0.889 & 0.881 & 0.875 & 0.872 & 0.876 & 0.873 & 0.872 & 0.876 & 0.878 & 0.875 & 0.877 \\
Fault            & 0.604 & 0.603 & 0.607 & 0.589 & 0.591 & 0.583 & 0.591 & 0.597 & 0.583 & 0.600 & 0.576 \\
Glass            & 0.090 & 0.128 & 0.111 & 0.114 & 0.111 & 0.105 & 0.106 & 0.106 & 0.103 & 0.103 & 0.101 \\
Hepatitis        & 0.583 & 0.626 & 0.611 & 0.555 & 0.584 & 0.559 & 0.563 & 0.615 & 0.571 & 0.564 & 0.564 \\
Imgseg           & 0.772 & 0.824 & 0.823 & 0.828 & 0.829 & 0.825 & 0.840 & 0.832 & 0.838 & 0.831 & 0.831 \\
Ionosphere       & 0.897 & 0.926 & 0.919 & 0.944 & 0.939 & 0.902 & 0.912 & 0.926 & 0.922 & 0.910 & 0.914 \\
Lympho           & 1.000 & 0.989 & 0.969 & 0.968 & 0.986 & 0.963 & 0.962 & 0.968 & 0.969 & 0.990 & 0.995 \\
Mammography      & 0.417 & 0.579 & 0.580 & 0.579 & 0.574 & 0.578 & 0.573 & 0.577 & 0.571 & 0.578 & 0.577 \\
Mnist            & 0.650 & 0.650 & 0.650 & 0.650 & 0.650 & 0.662 & 0.651 & 0.650 & 0.650 & 0.650 & 0.650 \\
Musk             & 1.000 & 0.997 & 1.000 & 1.000 & 1.000 & 1.000 & 1.000 & 1.000 & 1.000 & 1.000 & 1.000 \\
Optdigits        & 0.060 & 0.060 & 0.060 & 0.060 & 0.060 & 0.060 & 0.060 & 0.060 & 0.060 & 0.060 & 0.060 \\
Parkinson        & 0.930 & 0.944 & 0.946 & 0.944 & 0.943 & 0.944 & 0.943 & 0.941 & 0.942 & 0.941 & 0.941 \\
Pendigits        & 0.386 & 0.485 & 0.467 & 0.485 & 0.478 & 0.550 & 0.544 & 0.548 & 0.585 & 0.558 & 0.546 \\
Pima             & 0.701 & 0.751 & 0.751 & 0.751 & 0.751 & 0.750 & 0.748 & 0.748 & 0.744 & 0.742 & 0.739 \\
Satellite        & 0.778 & 0.792 & 0.793 & 0.790 & 0.789 & 0.790 & 0.790 & 0.789 & 0.788 & 0.788 & 0.787 \\
Satimage-2       & 0.919 & 0.931 & 0.929 & 0.928 & 0.935 & 0.931 & 0.929 & 0.931 & 0.931 & 0.931 & 0.933 \\
Shuttle          & 0.963 & 0.991 & 0.991 & 0.991 & 0.991 & 0.991 & 0.990 & 0.991 & 0.991 & 0.991 & 0.991 \\
SpamBase         & 0.818 & 0.823 & 0.825 & 0.824 & 0.824 & 0.823 & 0.823 & 0.825 & 0.826 & 0.826 & 0.825 \\
Speech           & 0.028 & 0.029 & 0.031 & 0.033 & 0.031 & 0.032 & 0.035 & 0.029 & 0.029 & 0.030 & 0.030 \\
Thyroid          & 0.813 & 0.815 & 0.832 & 0.852 & 0.863 & 0.869 & 0.879 & 0.875 & 0.882 & 0.881 & 0.877 \\
Vertebral        & 0.138 & 0.143 & 0.139 & 0.142 & 0.142 & 0.141 & 0.142 & 0.143 & 0.140 & 0.140 & 0.141 \\
Vowels           & 0.105 & 0.109 & 0.123 & 0.132 & 0.131 & 0.133 & 0.137 & 0.156 & 0.128 & 0.137 & 0.151 \\
Wbc              & 0.839 & 0.840 & 0.834 & 0.840 & 0.827 & 0.818 & 0.831 & 0.846 & 0.825 & 0.826 & 0.826 \\
WDBC             & 0.983 & 0.980 & 0.992 & 0.989 & 0.974 & 0.964 & 0.977 & 0.974 & 0.972 & 0.972 & 0.970 \\
Wilt             & 0.064 & 0.097 & 0.108 & 0.106 & 0.120 & 0.141 & 0.132 & 0.144 & 0.123 & 0.135 & 0.137 \\
Wine             & 0.133 & 0.846 & 0.801 & 0.801 & 0.782 & 0.741 & 0.757 & 0.730 & 0.714 & 0.726 & 0.732 \\
WPBC             & 0.394 & 0.408 & 0.406 & 0.415 & 0.398 & 0.413 & 0.415 & 0.416 & 0.409 & 0.400 & 0.404 \\
Yeast            & 0.468 & 0.505 & 0.508 & 0.508 & 0.511 & 0.513 & 0.512 & 0.514 & 0.513 & 0.511 & 0.514 \\ \midrule
Average          & 0.597 & 0.638 & 0.637 & 0.638 & 0.637 & 0.636 & 0.638 & 0.641 & 0.637 & 0.637 & 0.638 \\ 
\bottomrule
\end{tabular}
\end{table}

\begin{table}[h!]
\centering
\caption{AUC-PR performance of IForest and IForest+LLM-DAS under different anomaly synthesis ratios.}
\scriptsize
\begin{tabular}{c|c|cccccccccc}
\toprule
                 & \multirow{2}{*}{\centering IForest}   & \multicolumn{10}{c}{LLM-DAS with various anomaly synthesis ratios}\\ \cline{3-12}
                 &       & 0.1   & 0.2   & 0.3   & 0.4   & 0.5   & 0.6   & 0.7   & 0.8   & 0.9   & 1     \\ \midrule
Abalone          & 0.848   & 0.855 & 0.856 & 0.857 & 0.859 & 0.858 & 0.858 & 0.858 & 0.858 & 0.859 & 0.858  \\
Amazon           & 0.109   & 0.109 & 0.109 & 0.109 & 0.109 & 0.109 & 0.109 & 0.109 & 0.109 & 0.109 & 0.109  \\
Annthyroid       & 0.615   & 0.620 & 0.628 & 0.628 & 0.630 & 0.647 & 0.634 & 0.643 & 0.636 & 0.633 & 0.647  \\
Arrhythmia       & 0.510   & 0.602 & 0.590 & 0.586 & 0.581 & 0.571 & 0.566 & 0.556 & 0.556 & 0.562 & 0.571  \\
Breastw          & 0.945   & 0.995 & 0.994 & 0.995 & 0.995 & 0.996 & 0.996 & 0.996 & 0.996 & 0.997 & 0.996  \\
Campaign         & 0.461   & 0.452 & 0.452 & 0.452 & 0.452 & 0.452 & 0.452 & 0.452 & 0.452 & 0.452 & 0.452  \\
Cardio           & 0.702   & 0.622 & 0.674 & 0.634 & 0.640 & 0.647 & 0.675 & 0.671 & 0.670 & 0.676 & 0.676  \\
Cardiotocography & 0.604   & 0.638 & 0.666 & 0.669 & 0.667 & 0.684 & 0.669 & 0.682 & 0.684 & 0.681 & 0.681  \\
Comm.and.crime   & 0.894   & 0.896 & 0.895 & 0.894 & 0.894 & 0.894 & 0.894 & 0.893 & 0.893 & 0.894 & 0.891  \\
Fault            & 0.595   & 0.577 & 0.574 & 0.568 & 0.562 & 0.564 & 0.563 & 0.557 & 0.555 & 0.557 & 0.555  \\
Glass            & 0.095   & 0.120 & 0.119 & 0.124 & 0.121 & 0.119 & 0.106 & 0.119 & 0.098 & 0.094 & 0.096  \\
Hepatitis        & 0.418   & 0.566 & 0.587 & 0.556 & 0.596 & 0.673 & 0.652 & 0.655 & 0.619 & 0.598 & 0.642  \\
Imgseg           & 0.756   & 0.859 & 0.852 & 0.846 & 0.836 & 0.824 & 0.829 & 0.836 & 0.829 & 0.829 & 0.829  \\
Ionosphere       & 0.977   & 0.869 & 0.818 & 0.763 & 0.746 & 0.699 & 0.675 & 0.670 & 0.654 & 0.641 & 0.631  \\
Lympho           & 0.959   & 0.979 & 0.973 & 0.965 & 0.943 & 0.924 & 0.956 & 0.955 & 0.932 & 0.925 & 0.963  \\
Mammography      & 0.333   & 0.571 & 0.584 & 0.575 & 0.576 & 0.572 & 0.576 & 0.572 & 0.567 & 0.565 & 0.567  \\
Mnist            & 0.535   & 0.601 & 0.610 & 0.610 & 0.610 & 0.610 & 0.610 & 0.610 & 0.610 & 0.610 & 0.610  \\
Musk             & 0.528   & 1.000 & 1.000 & 0.999 & 0.999 & 0.989 & 0.996 & 0.990 & 0.962 & 0.959 & 0.933  \\
Optdigits        & 0.157   & 0.179 & 0.161 & 0.161 & 0.161 & 0.161 & 0.161 & 0.161 & 0.161 & 0.161 & 0.161  \\
Parkinson        & 0.960   & 0.951 & 0.955 & 0.954 & 0.958 & 0.962 & 0.963 & 0.963 & 0.963 & 0.962 & 0.964  \\
Pendigits        & 0.513   & 0.602 & 0.604 & 0.608 & 0.596 & 0.626 & 0.630 & 0.639 & 0.652 & 0.642 & 0.643  \\
Pima             & 0.666   & 0.740 & 0.738 & 0.735 & 0.740 & 0.736 & 0.739 & 0.743 & 0.741 & 0.742 & 0.744  \\
Satellite        & 0.858   & 0.808 & 0.803 & 0.804 & 0.801 & 0.799 & 0.796 & 0.795 & 0.792 & 0.794 & 0.794  \\
Satimage-2       & 0.885   & 0.871 & 0.844 & 0.830 & 0.823 & 0.771 & 0.737 & 0.783 & 0.760 & 0.800 & 0.787  \\
Shuttle          & 0.917   & 0.995 & 0.995 & 0.995 & 0.995 & 0.994 & 0.994 & 0.994 & 0.994 & 0.994 & 0.995  \\
SpamBase         & 0.890   & 0.889 & 0.890 & 0.890 & 0.890 & 0.889 & 0.890 & 0.889 & 0.889 & 0.889 & 0.888  \\
Speech           & 0.035   & 0.032 & 0.037 & 0.037 & 0.034 & 0.033 & 0.036 & 0.035 & 0.032 & 0.039 & 0.037  \\
Thyroid          & 0.606   & 0.886 & 0.895 & 0.891 & 0.945 & 0.912 & 0.914 & 0.914 & 0.936 & 0.933 & 0.935  \\
Vertebral        & 0.134   & 0.269 & 0.304 & 0.315 & 0.320 & 0.335 & 0.385 & 0.384 & 0.373 & 0.389 & 0.382  \\
Vowels           & 0.098   & 0.178 & 0.150 & 0.161 & 0.151 & 0.172 & 0.186 & 0.160 & 0.182 & 0.165 & 0.190  \\
Wbc              & 0.857   & 0.823 & 0.844 & 0.842 & 0.846 & 0.840 & 0.841 & 0.830 & 0.832 & 0.827 & 0.831  \\
WDBC             & 0.975   & 0.994 & 0.987 & 0.989 & 0.976 & 0.986 & 0.984 & 0.972 & 0.973 & 0.944 & 0.962  \\
Wilt             & 0.085   & 0.278 & 0.263 & 0.238 & 0.207 & 0.187 & 0.181 & 0.179 & 0.165 & 0.157 & 0.154  \\
Wine             & 0.246   & 0.456 & 0.307 & 0.361 & 0.352 & 0.414 & 0.449 & 0.472 & 0.483 & 0.518 & 0.514  \\
WPBC             & 0.376   & 0.399 & 0.396 & 0.395 & 0.390 & 0.405 & 0.406 & 0.400 & 0.405 & 0.408 & 0.412  \\
Yeast            & 0.465   & 0.546 & 0.574 & 0.600 & 0.597 & 0.603 & 0.603 & 0.603 & 0.610 & 0.604 & 0.607  \\ \midrule
Average          & 0.572   & 0.634 & 0.631 & 0.629 & 0.628 & 0.629 & 0.631 & 0.632 & 0.628 & 0.628 & 0.631  \\
\bottomrule
\end{tabular}
\end{table}

\begin{figure}[h!]
     \includegraphics[width=\linewidth]{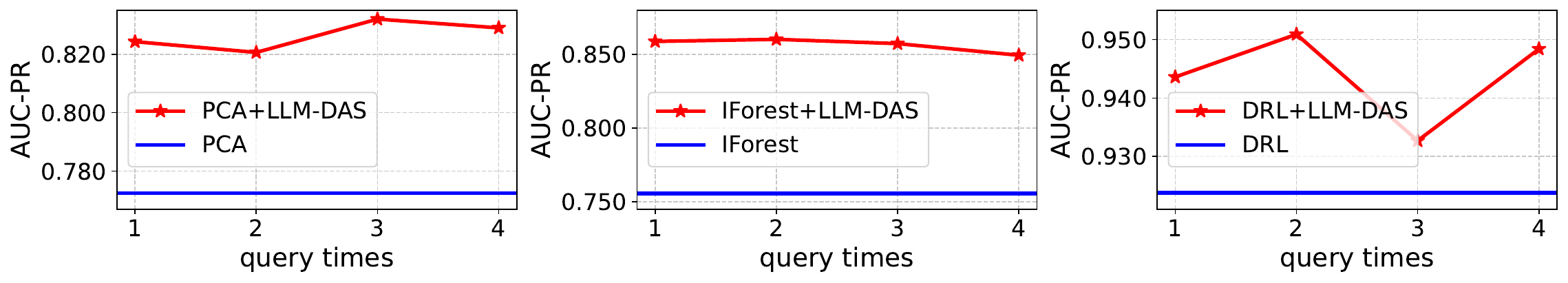}
    \captionsetup{font=small}
    \captionof{figure}{\label{fig:methods2} \small{AUC-PR-based performance comparison of different query times on the Imgseg dataset, with results for PCA (left), IForest (middle) and DRL(right).}}
\end{figure}

\begin{figure}[h!]
     \includegraphics[width=\linewidth]{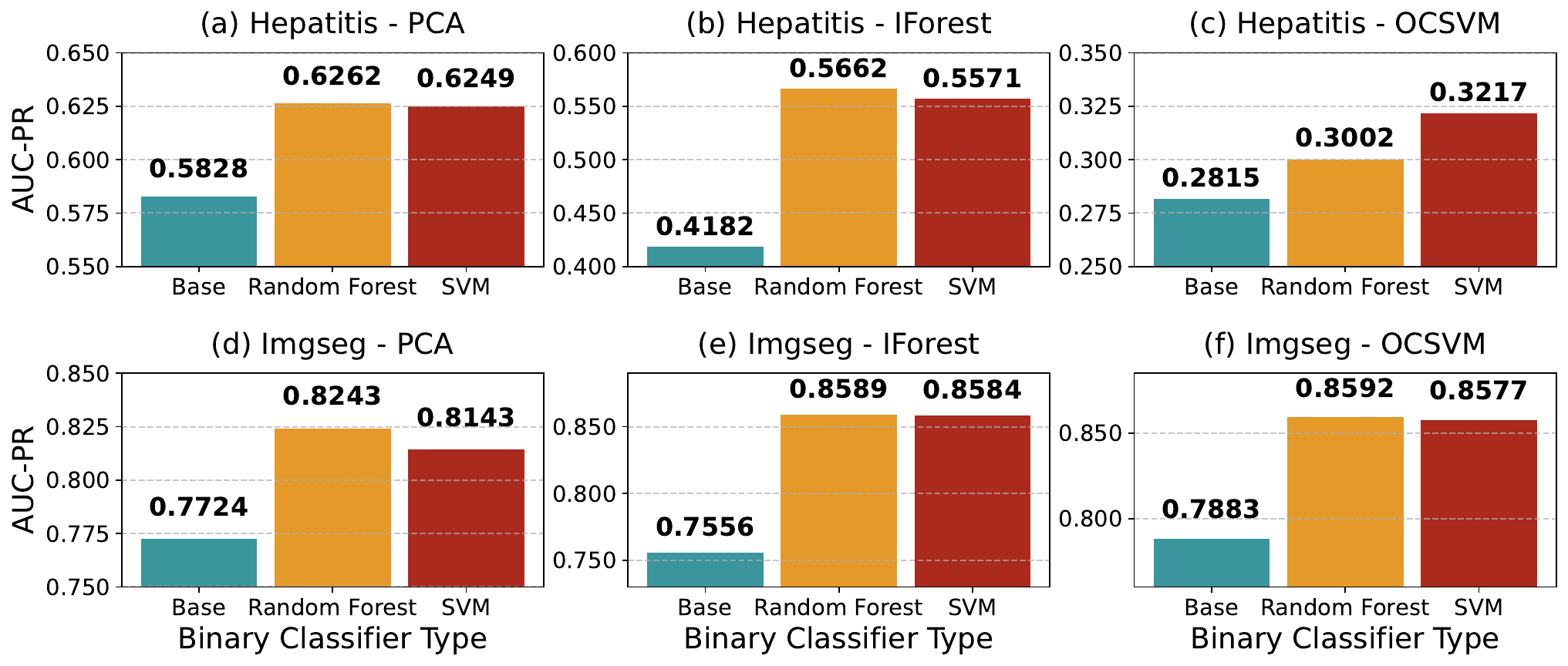}
    \captionsetup{font=small}
    \captionof{figure}{\label{fig:methods3} \small{AUC-PR-based performance comparison of different binary classifier choices for LLM-DAS on the Hepatitis and Imgseg datasets,  where PCA (left), IForest (middle), and OCSVM (right) serve as the base detectors. ``Base'' corresponds to the base detector performance. ``Random Forest'' and ``SVM'' correspond to LLM-DAS's performance with different binary classifier types.}}
\end{figure}

\begin{figure}[h!]
     \includegraphics[width=\linewidth]{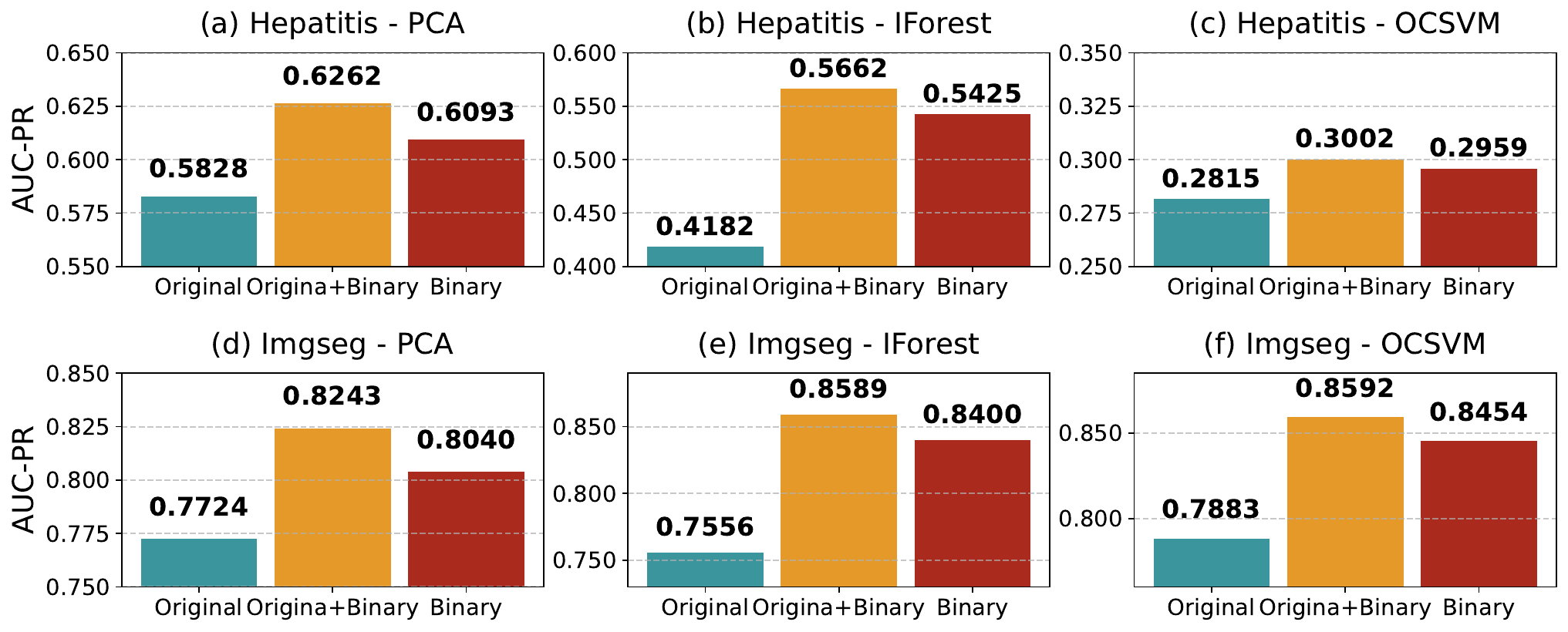}
    \captionsetup{font=small}
    \captionof{figure}{\label{fig:methods4} \small{AUC-PR-based performance comparison of ensemble strategies on the Hepatitis and Imgseg datasets, where PCA (left), IForest (middle), and OCSVM (right) serve as the base detectors. ``Original'' denotes the base detector performance. ``Origina+Binary''denotes the performance of LLM-DAS (our proposed method, combining binary classifiers with original detectors), while ``Binary'' denotes the scheme using binary classifiers directly without integrating original detectors.}}
\end{figure}

\clearpage
\subsection{Additional results}
\label{appendix:additional results}

\subsubsection{Comparison with Alternative Anomaly Synthesis Methods}
\label{appendix: Alternative Anomaly Synthesis}

To further verify the effectiveness of LLM-DAS, we compared several representative anomaly synthesis strategies, including \textit{Gaussian noise injection}, \textit{random outliers}, and \textit{SMOTE-based sample generation}, which are denoted as \textbf{Gauss}, \textbf{Random}, and \textbf{Smote} respectively. The results are in Table~\ref{appendix:alternative anomaly synthesis PCA} to Table~\ref{appendix:alternative anomaly synthesis DRL}.

Three Common Anomalies synthesis strategies:
\begin{itemize}
    \item \textit{Gaussian noise injection} is a method for generating anomalous samples by introducing Gaussian-distributed noise to normal training samples.The process works as follows: first, randomly select a specified number of base samples from the normal training data. Then, generate noise that follows a Gaussian distribution with a predefined mean and standard deviation. This noise is then added to the selected base samples, resulting in new samples that deviate from the original normal data.
    \item \textit{Random outliers} refer to anomalous samples generated by randomly sampling points within an extended feature space derived from the range of normal training samples. This method first expands the boundaries of each feature dimension beyond the minimum and maximum values of normal data, then randomly distributes the generated anomalies within this extended range, thereby ensuring the anomalies deviate from the distribution of normal samples.
    \item \textit{SMOTE-based sample generation} draws on the SMOTE algorithm, which is originally used to address data imbalance issues. It can generate boundary anomalous samples near normal training samples.The method first randomly selects base samples from the normal training data and finds k nearest neighbors for each base sample. Then, it calculates the difference between each base sample and a randomly chosen neighbor, and extrapolates along the line connecting them to generate new samples.To create boundary anomalous samples that are closer to or slightly beyond the boundary of the normal data distribution, the interpolation factor is set within the range of 0.5 to 1.5. This ensures that the generated samples either lie near the boundary of the normal data distribution or just outside it, thus effectively serving as boundary anomalous samples.
\end{itemize}

\begin{table}[h!]
\centering
\caption{AUC-PR ($\uparrow$) performance comparison of PCA under different alternative anomaly synthesis methods across 36 datasets.}
\label{appendix:alternative anomaly synthesis PCA}
\begin{tabular}{lccccc}
\toprule
                 & PCA    & Gauss  & Random & Smote  & LLM-DAS          \\ \midrule
Abalone          & 0.8393 & 0.8307 & 0.8418 & 0.7683 & 0.8411           \\
Amazon           & 0.1072 & 0.1079 & 0.1077 & 0.1116 & 0.1077           \\
Annthyroid       & 0.5657 & 0.6143 & 0.5667 & 0.2574 & 0.6602           \\
Arrhythmia       & 0.5336 & 0.4718 & 0.5406 & 0.5216 & 0.5508           \\
Breastw          & 0.9934 & 0.9881 & 0.9939 & 0.9872 & 0.9956           \\
Campaign         & 0.4884 & 0.4860 & 0.4845 & 0.3820 & 0.4884           \\
Cardio           & 0.8628 & 0.7322 & 0.9025 & 0.7755 & 0.8243           \\
Cardiotocography & 0.6969 & 0.6646 & 0.7145 & 0.5911 & 0.7337           \\
Comm.and.crime   & 0.8892 & 0.8628 & 0.8829 & 0.7624 & 0.8806           \\
Fault            & 0.6035 & 0.5131 & 0.5451 & 0.5475 & 0.6034           \\
Glass            & 0.0896 & 0.0975 & 0.1055 & 0.0937 & 0.1278           \\
Hepatitis        & 0.5828 & 0.4790 & 0.4874 & 0.5052 & 0.6262           \\
Imgseg           & 0.7724 & 0.7722 & 0.7793 & 0.8002 & 0.8243           \\
Ionosphere       & 0.8969 & 0.9480 & 0.9372 & 0.9295 & 0.9260           \\
Lympho           & 1.0000 & 0.8056 & 0.9754 & 0.9440 & 0.9889           \\
Mammography      & 0.4165 & 0.4184 & 0.4990 & 0.4003 & 0.5793           \\
Mnist            & 0.6499 & 0.6584 & 0.6445 & 0.5175 & 0.6499           \\
Musk             & 1.0000 & 0.9981 & 0.9990 & 1.0000 & 0.9969           \\
Optdigits        & 0.0602 & 0.0601 & 0.0604 & 0.0724 & 0.0602           \\
Parkinson        & 0.9297 & 0.9372 & 0.9442 & 0.9300 & 0.9436           \\
Pendigits        & 0.3863 & 0.6063 & 0.4464 & 0.3828 & 0.4847           \\
Pima             & 0.7008 & 0.6723 & 0.7054 & 0.6732 & 0.7509           \\
Satellite        & 0.7778 & 0.8137 & 0.7897 & 0.8035 & 0.7921           \\
Satimage-2       & 0.9192 & 0.9163 & 0.9517 & 0.8954 & 0.9311           \\
Shuttle          & 0.9627 & 0.9845 & 0.9891 & 0.9726 & 0.9906           \\
SpamBase         & 0.8184 & 0.8230 & 0.8384 & 0.7829 & 0.8228           \\
Speech           & 0.0277 & 0.0295 & 0.0250 & 0.0275 & 0.0293           \\
Thyroid          & 0.8134 & 0.8486 & 0.8230 & 0.6885 & 0.8146           \\
Vertebral        & 0.1381 & 0.1933 & 0.1424 & 0.1771 & 0.1433           \\
Vowels           & 0.1051 & 0.1570 & 0.1680 & 0.1958 & 0.1090           \\
Wbc              & 0.8391 & 0.7876 & 0.8159 & 0.7477 & 0.8400           \\
WDBC             & 0.9833 & 0.8165 & 0.9912 & 0.6980 & 0.9802           \\
Wilt             & 0.0641 & 0.1281 & 0.0642 & 0.1334 & 0.0972           \\
Wine             & 0.1325 & 0.1859 & 0.1772 & 0.1401 & 0.8459           \\
WPBC             & 0.3940 & 0.3522 & 0.3871 & 0.4112 & 0.4085           \\
Yeast            & 0.4678 & 0.5049 & 0.4825 & 0.5126 & 0.5053           \\ \midrule
Average          & 0.5975 & 0.5907 & 0.6058 & 0.5594 & \textbf{0.6376}\\ \bottomrule   
\end{tabular}
\end{table}

\begin{table}[h!]
\centering
\caption{AUC-PR ($\uparrow$) performance comparison of IForest under different alternative anomaly synthesis methods across 36 datasets.}
\label{appendix:alternative anomaly synthesis IForest}
\begin{tabular}{lccccc}
\toprule
                  & IForest & Gauss  & Random & Smote  & LLM-DAS          \\ \midrule
Abalone          & 0.8481  & 0.8177 & 0.8523 & 0.8056 & 0.8555           \\
Amazon           & 0.1091  & 0.1094 & 0.1093 & 0.1097 & 0.1093           \\
Annthyroid       & 0.6149  & 0.6642 & 0.6365 & 0.1515 & 0.6200           \\
Arrhythmia       & 0.5097  & 0.5044 & 0.5848 & 0.5562 & 0.6021           \\
Breastw          & 0.9449  & 0.9817 & 0.9949 & 0.9839 & 0.9947           \\
Campaign         & 0.4608  & 0.4396 & 0.4421 & 0.3518 & 0.4517           \\
Cardio           & 0.7018  & 0.6447 & 0.8473 & 0.6018 & 0.6220           \\
Cardiotocography & 0.6036  & 0.6355 & 0.6835 & 0.5607 & 0.6382           \\
Comm.and.crime   & 0.8940  & 0.8577 & 0.8503 & 0.7641 & 0.8958           \\
Fault            & 0.5948  & 0.4997 & 0.5393 & 0.5677 & 0.5774           \\
Glass            & 0.0952  & 0.1162 & 0.1495 & 0.1218 & 0.1205           \\
Hepatitis        & 0.4182  & 0.5088 & 0.4368 & 0.4816 & 0.5662           \\
Imgseg           & 0.7556  & 0.7894 & 0.7987 & 0.8346 & 0.8589           \\
Ionosphere       & 0.9768  & 0.9374 & 0.9237 & 0.9006 & 0.8685           \\
Lympho           & 0.9593  & 0.6675 & 0.9642 & 0.9503 & 0.9794           \\
Mammography      & 0.3334  & 0.3690 & 0.4970 & 0.3024 & 0.5712           \\
Mnist            & 0.5349  & 0.6283 & 0.5880 & 0.4381 & 0.6011           \\
Musk             & 0.5279  & 0.9168 & 1.0000 & 1.0000 & 1.0000           \\
Optdigits        & 0.1570  & 0.1458 & 0.1417 & 0.1555 & 0.1790           \\
Parkinson        & 0.9595  & 0.9639 & 0.9583 & 0.9175 & 0.9514           \\
Pendigits        & 0.5133  & 0.6070 & 0.5365 & 0.4698 & 0.6020           \\
Pima             & 0.6662  & 0.6985 & 0.7074 & 0.6792 & 0.7397           \\
Satellite        & 0.8583  & 0.8206 & 0.8281 & 0.8118 & 0.8079           \\
Satimage-2       & 0.8846  & 0.9063 & 0.9495 & 0.9119 & 0.8706           \\
Shuttle          & 0.9172  & 0.9870 & 0.9937 & 0.9813 & 0.9954           \\
SpamBase         & 0.8902  & 0.8707 & 0.8809 & 0.8227 & 0.8892           \\
Speech           & 0.0353  & 0.0331 & 0.0279 & 0.0301 & 0.0318           \\
Thyroid          & 0.6055  & 0.7906 & 0.8381 & 0.5110 & 0.8856           \\
Vertebral        & 0.1342  & 0.1557 & 0.1354 & 0.2320 & 0.2686           \\
Vowels           & 0.0984  & 0.1729 & 0.1436 & 0.1676 & 0.1779           \\
Wbc              & 0.8573  & 0.8015 & 0.8271 & 0.7239 & 0.8233           \\
WDBC             & 0.9749  & 0.7128 & 0.9625 & 0.6298 & 0.9936           \\
Wilt             & 0.0848  & 0.1246 & 0.0871 & 0.1531 & 0.2784           \\
Wine             & 0.2458  & 0.3130 & 0.1687 & 0.2132 & 0.4559           \\
WPBC             & 0.3760  & 0.3829 & 0.3998 & 0.4076 & 0.3993           \\
Yeast            & 0.4654  & 0.4964 & 0.4717 & 0.4933 & 0.5462           \\ \midrule
Average          & 0.5724  & 0.5853 & 0.6099 & 0.5498 & \textbf{0.6341}\\ \bottomrule   
\end{tabular}
\end{table}

\begin{table}[h!]
\centering
\caption{AUC-PR ($\uparrow$) performance comparison of DRL under different alternative anomaly synthesis methods across 36 datasets.}
\label{appendix:alternative anomaly synthesis DRL}
\begin{tabular}{lccccc}
\toprule
                 & DRL    & Gauss  & Random & Smote  & LLM-DAS          \\ \midrule
Abalone          & 0.8850 & 0.8394 & 0.8747 & 0.8202 & 0.9021           \\
Amazon           & 0.1206 & 0.0972 & 0.1150 & 0.1021 & 0.1197           \\
Annthyroid       & 0.6761 & 0.1731 & 0.6310 & 0.1727 & 0.7618           \\
Arrhythmia       & 0.6270 & 0.6273 & 0.6518 & 0.6992 & 0.7298           \\
Breastw          & 0.9966 & 0.9857 & 0.9942 & 0.9639 & 0.9972           \\
Campaign         & 0.5013 & 0.4854 & 0.4898 & 0.4625 & 0.4945           \\
Cardio           & 0.8325 & 0.8480 & 0.9142 & 0.8427 & 0.8861           \\
Cardiotocography & 0.7540 & 0.7323 & 0.7148 & 0.7252 & 0.8046           \\
Comm.and.crime   & 0.9164 & 0.9421 & 0.9455 & 0.9386 & 0.9470           \\
Fault            & 0.6649 & 0.5765 & 0.6417 & 0.5880 & 0.7053           \\
Glass            & 0.1670 & 0.3107 & 0.2948 & 0.2273 & 0.4934           \\
Hepatitis        & 0.6627 & 0.6880 & 0.6856 & 0.6870 & 0.6957           \\
Imgseg           & 0.9238 & 0.9239 & 0.9300 & 0.9363 & 0.9436           \\
Ionosphere       & 0.9895 & 0.9530 & 0.9636 & 0.9539 & 0.9846           \\
Lympho           & 1.0000 & 0.6193 & 0.7607 & 0.7348 & 0.8706           \\
Mammography      & 0.8406 & 0.3719 & 0.4496 & 0.2873 & 0.5288           \\
Mnist            & 0.8870 & 0.9067 & 0.9049 & 0.8693 & 0.9109           \\
Musk             & 1.0000 & 0.8481 & 1.0000 & 0.9998 & 1.0000           \\
Optdigits        & 0.9356 & 0.2927 & 0.2713 & 0.3133 & 0.9566           \\
Parkinson        & 0.9210 & 0.9649 & 0.9629 & 0.9636 & 0.9659           \\
Pendigits        & 0.9360 & 0.0688 & 0.2553 & 0.2345 & 0.8848           \\
Pima             & 0.7449 & 0.6972 & 0.7100 & 0.7083 & 0.7912           \\
Satellite        & 0.8692 & 0.8408 & 0.8857 & 0.7997 & 0.8942           \\
Satimage-2       & 0.9703 & 0.9160 & 0.9666 & 0.9248 & 0.9804           \\
Shuttle          & 0.9819 & 0.0884 & 0.9924 & 0.0875 & 0.9991           \\
SpamBase         & 0.8413 & 0.6783 & 0.8975 & 0.8278 & 0.9335           \\
Speech           & 0.0584 & 0.0516 & 0.0511 & 0.0476 & 0.1189           \\
Thyroid          & 0.8626 & 0.2007 & 0.7836 & 0.1660 & 0.8770           \\
Vertebral        & 0.2854 & 0.2369 & 0.2592 & 0.2277 & 0.5657           \\
Vowels           & 0.4506 & 0.8461 & 0.8216 & 0.8381 & 0.8918           \\
Wbc              & 0.9742 & 0.8381 & 0.7973 & 0.7393 & 0.9341           \\
WDBC             & 1.0000 & 0.9164 & 0.9363 & 0.8364 & 1.0000           \\
Wilt             & 0.4543 & 0.6363 & 0.4872 & 0.3760 & 0.5926           \\
Wine             & 1.0000 & 0.9564 & 0.9292 & 0.9423 & 1.0000           \\
WPBC             & 0.5017 & 0.4507 & 0.4409 & 0.4720 & 0.5671           \\
Yeast            & 0.5416 & 0.5263 & 0.5143 & 0.6136 & 0.5281           \\ \midrule
Average          & 0.7437 & 0.6149 & 0.6923 & 0.6147 & \textbf{0.7849}  \\ \bottomrule   
\end{tabular}
\end{table}

\clearpage
\subsubsection{Ablation Studies on LLM-DAS Design}
\label{appendix: Ablation Studies on LLM-DAS Design}
To further validate the effectiveness of the key components in LLM-DAS, we conduct an ablation study by systematically removing distinct components of our framework. The results are in Table~\ref{appendix: ablation PCA} to Table~\ref{appendix: ablation DRL}. Specifically, we compare the full LLM-DAS model with its variants that:

\begin{itemize}
    \item \textit{Remove detector-awareness and use generic prompts} (\textbf{Generic}): modify the prompt by removing descriptions of the specific detector's principles, and instruct the LLM to generate a ``general-purpose'' synthesis code. By comparing its performance with that of LLM-DAS, we can demonstrate that ``awareness'' the detector's underlying principles and designing targeted anomaly sample generation strategies are crucial.
    
    \item \textit{Replace ``hard'' anomalies with ``simple'' ones} (\textbf{Simple}): modify  the prompt such that the LLM generates only ``simple'' anomalies that are far from the normal cluster. We remove the availability of model.predict\_score() for evaluating the anomaly degree of generated samples, thereby eliminating the capability to generate more difficult anomalies based on model.predict\_score() values. By comparing the performance of this setup against that of LLM-DAS, we can verify that the ``hard'' requirement (for generating hard-to-detect anomalies) is critical.

    \item \textit{Replace borderline synthesis with random selection} (\textbf{Random}):  generates anomalies by first transforming ``randomly selected'' normal samples into anomalies through prompt modification, unlike LLM-DAS, which emphasizes selecting or transforming ``borderline normal samples'' from the training set into anomalies. By comparing the performance of this ``randomly selected normal sample-based transformation'' method against that of LLM-DAS, we can verify that the ``borderline'' heuristic is critical.
\end{itemize}

\begin{table}[h!]
\centering
\caption{Ablation Studies on the Different Components of LLM-DAS Design}
\begin{tabular}{lccc}
\toprule
Variants & Awareness & Hard & Borderline  \\ \midrule 
Generic  & -         & \checkmark    &\checkmark           \\
Simple   & \checkmark         & -    & -           \\
Random   & \checkmark         & \checkmark    & -           \\
LLM-DAS  & \checkmark         &\checkmark    & \checkmark       \\  \bottomrule    
\end{tabular}
\end{table}

\begin{table}[h!]
\centering
\caption{AUC-PR ($\uparrow$) performance comparison of PCA under different LLM prompt designs across 36 datasets.}
\label{appendix: ablation PCA}
\begin{tabular}{lccccc}
\toprule
                  & PCA    & Generic & Simple & Random & LLM-DAS  \\ \midrule 
Abalone          & 0.8393 & 0.8290 & 0.8252 & 0.8415 & 0.8411   \\
Amazon           & 0.1072 & 0.1079 & 0.1055 & 0.1078 & 0.1077   \\
Annthyroid       & 0.5657 & 0.5614 & 0.5329 & 0.5418 & 0.6602   \\
Arrhythmia       & 0.5336 & 0.5497 & 0.4569 & 0.5442 & 0.5508   \\
Breastw          & 0.9934 & 0.9944 & 0.9916 & 0.9937 & 0.9956   \\
Campaign         & 0.4884 & 0.4784 & 0.4894 & 0.4881 & 0.4884   \\
Cardio           & 0.8628 & 0.7465 & 0.7576 & 0.7830 & 0.8243   \\
Cardiotocography & 0.6969 & 0.6627 & 0.7612 & 0.6716 & 0.7337   \\
Comm.and.crime   & 0.8892 & 0.8909 & 0.8854 & 0.8312 & 0.8806   \\
Fault            & 0.6035 & 0.5320 & 0.5045 & 0.5421 & 0.6034   \\
Glass            & 0.0896 & 0.0968 & 0.0829 & 0.0999 & 0.1278   \\
Hepatitis        & 0.5828 & 0.5799 & 0.5948 & 0.4836 & 0.6262   \\
Imgseg           & 0.7724 & 0.8150 & 0.8054 & 0.8124 & 0.8243   \\
Ionosphere       & 0.8969 & 0.8291 & 0.9571 & 0.9246 & 0.9260   \\
Lympho           & 1.0000 & 0.9549 & 0.8556 & 0.9718 & 0.9889   \\
Mammography      & 0.4165 & 0.4808 & 0.3893 & 0.4725 & 0.5793   \\
Mnist            & 0.6499 & 0.6423 & 0.6549 & 0.6543 & 0.6499   \\
Musk             & 1.0000 & 0.9855 & 0.9981 & 0.9585 & 0.9969   \\
Optdigits        & 0.0602 & 0.0677 & 0.0602 & 0.0602 & 0.0602   \\
Parkinson        & 0.9297 & 0.9341 & 0.9323 & 0.9343 & 0.9436   \\
Pendigits        & 0.3863 & 0.5422 & 0.6302 & 0.4028 & 0.4847   \\
Pima             & 0.7008 & 0.7064 & 0.6244 & 0.7034 & 0.7509   \\
Satellite        & 0.7778 & 0.7923 & 0.7697 & 0.7886 & 0.7921   \\
Satimage-2       & 0.9192 & 0.8532 & 0.2350 & 0.8704 & 0.9311   \\
Shuttle          & 0.9627 & 0.9862 & 0.9791 & 0.9745 & 0.9906   \\
SpamBase         & 0.8184 & 0.8237 & 0.8289 & 0.8312 & 0.8228   \\
Speech           & 0.0277 & 0.0279 & 0.0311 & 0.0280 & 0.0293   \\
Thyroid          & 0.8134 & 0.8244 & 0.7363 & 0.8364 & 0.8146   \\
Vertebral        & 0.1381 & 0.1426 & 0.1391 & 0.1426 & 0.1433   \\
Vowels           & 0.1051 & 0.0930 & 0.2136 & 0.1203 & 0.1090   \\
Wbc              & 0.8391 & 0.7705 & 0.7304 & 0.8249 & 0.8400   \\
WDBC             & 0.9833 & 0.8506 & 0.9124 & 0.9226 & 0.9802   \\
Wilt             & 0.0641 & 0.0658 & 0.0642 & 0.0641 & 0.0972   \\
Wine             & 0.1325 & 0.2685 & 0.1809 & 0.7635 & 0.8459   \\
WPBC             & 0.3940 & 0.3872 & 0.4341 & 0.3855 & 0.4085   \\
Yeast            & 0.4678 & 0.4723 & 0.4789 & 0.4964 & 0.5053   \\ \midrule 
Average          & 0.5975 & 0.5929 & 0.5730 & 0.6076 & \textbf{0.6376} \\ \bottomrule   
\end{tabular}
\end{table}

\begin{table}[h!]
\centering
\caption{AUC-PR ($\uparrow$) performance comparison of IForest under different LLM prompt designs across 36 datasets.}
\label{appendix: ablation IForest}
\begin{tabular}{lccccc}
\toprule
                 & IForest     & Generic & Simple & Random & LLM-DAS  \\  \midrule 
Abalone          & 0.8481  & 0.8390 & 0.8492 & 0.7452 & 0.8555   \\
Amazon           & 0.1091  & 0.1107 & 0.1099 & 0.1102 & 0.1093   \\
Annthyroid       & 0.6149  & 0.5911 & 0.6243 & 0.2804 & 0.6200   \\
Arrhythmia       & 0.5097  & 0.5794 & 0.5951 & 0.4967 & 0.6021   \\
Breastw          & 0.9449  & 0.9966 & 0.9921 & 0.9266 & 0.9947   \\
Campaign         & 0.4608  & 0.4332 & 0.4517 & 0.4330 & 0.4517   \\
Cardio           & 0.7018  & 0.6611 & 0.7771 & 0.5234 & 0.6220   \\
Cardiotocography & 0.6036  & 0.6570 & 0.6513 & 0.5278 & 0.6382   \\
Comm.and.crime   & 0.8940  & 0.8794 & 0.7943 & 0.8661 & 0.8958   \\
Fault            & 0.5948  & 0.5347 & 0.5894 & 0.5602 & 0.5774   \\
Glass            & 0.0952  & 0.1004 & 0.1063 & 0.0890 & 0.1205   \\
Hepatitis        & 0.4182  & 0.5292 & 0.4908 & 0.3799 & 0.5662   \\
Imgseg           & 0.7556  & 0.8238 & 0.8268 & 0.6780 & 0.8589   \\
Ionosphere       & 0.9768  & 0.7474 & 0.9119 & 0.7488 & 0.8685   \\
Lympho           & 0.9593  & 0.9577 & 0.9213 & 0.8498 & 0.9794   \\
Mammography      & 0.3334  & 0.4366 & 0.4321 & 0.1642 & 0.5712   \\
Mnist            & 0.5349  & 0.5728 & 0.6154 & 0.6294 & 0.6011   \\
Musk             & 0.5279  & 0.8382 & 1.0000 & 0.8611 & 1.0000   \\
Optdigits        & 0.1570  & 0.1339 & 0.1579 & 0.1423 & 0.1790   \\
Parkinson        & 0.9595  & 0.9615 & 0.9557 & 0.9156 & 0.9514   \\
Pendigits        & 0.5133  & 0.6743 & 0.5427 & 0.5022 & 0.6020   \\
Pima             & 0.6662  & 0.7244 & 0.7110 & 0.6084 & 0.7397   \\
Satellite        & 0.8583  & 0.8203 & 0.8217 & 0.8158 & 0.8079   \\
Satimage-2       & 0.8846  & 0.5511 & 0.9515 & 0.1444 & 0.8706   \\
Shuttle          & 0.9172  & 0.9956 & 0.9912 & 0.9978 & 0.9954   \\
SpamBase         & 0.8902  & 0.8832 & 0.8175 & 0.8648 & 0.8892   \\
Speech           & 0.0353  & 0.0267 & 0.0296 & 0.0351 & 0.0318   \\
Thyroid          & 0.6055  & 0.7622 & 0.8280 & 0.1924 & 0.8856   \\
Vertebral        & 0.1342  & 0.1385 & 0.1339 & 0.1814 & 0.2686   \\
Vowels           & 0.0984  & 0.1127 & 0.1054 & 0.0940 & 0.1779   \\
Wbc              & 0.8573  & 0.7837 & 0.8531 & 0.2100 & 0.8233   \\
WDBC             & 0.9749  & 0.8893 & 0.9687 & 0.1164 & 0.9936   \\
Wilt             & 0.0848  & 0.0955 & 0.0871 & 0.1913 & 0.2784   \\
Wine             & 0.2458  & 0.1551 & 0.1886 & 0.5410 & 0.4559   \\
WPBC             & 0.3760  & 0.3752 & 0.3990 & 0.4400 & 0.3993   \\
Yeast            & 0.4654  & 0.4798 & 0.4668 & 0.4708 & 0.5462   \\  \midrule 
Average          & 0.5724  & 0.5792 & 0.6041 & 0.4815 & \textbf{0.6341} \\ \bottomrule   
\end{tabular}
\end{table}

\begin{table}[h!]
\centering
\caption{AUC-PR ($\uparrow$) performance comparison of DRL under different LLM prompt designs across 36 datasets.}
\label{appendix: ablation DRL}
\begin{tabular}{lccccc}
\toprule
                & DRL     & Generic & Simple & Random & LLM-DAS  \\ \midrule
Abalone          & 0.8850 & 0.8813 & 0.8716 & 0.8625 & 0.9021   \\
Amazon           & 0.1206 & 0.1023 & 0.1011 & 0.0958 & 0.1197   \\
Annthyroid       & 0.6761 & 0.3477 & 0.5409 & 0.3578 & 0.7618   \\
Arrhythmia       & 0.6270 & 0.6657 & 0.6138 & 0.6380 & 0.7298   \\
Breastw          & 0.9966 & 0.9954 & 0.9934 & 0.9879 & 0.9972   \\
Campaign         & 0.5013 & 0.4957 & 0.4857 & 0.4852 & 0.4945   \\
Cardio           & 0.8325 & 0.8978 & 0.9116 & 0.8597 & 0.8861   \\
Cardiotocography & 0.7540 & 0.6607 & 0.7109 & 0.7084 & 0.8046   \\
Comm.and.crime   & 0.9164 & 0.9460 & 0.9442 & 0.9438 & 0.9470   \\
Fault            & 0.6649 & 0.5967 & 0.6044 & 0.6537 & 0.7053   \\
Glass            & 0.1670 & 0.3493 & 0.3166 & 0.3036 & 0.4934   \\
Hepatitis        & 0.6627 & 0.6876 & 0.6880 & 0.6808 & 0.6957   \\
Imgseg           & 0.9238 & 0.9301 & 0.9308 & 0.9283 & 0.9436   \\
Ionosphere       & 0.9895 & 0.9486 & 0.9770 & 0.9514 & 0.9846   \\
Lympho           & 1.0000 & 0.7374 & 0.7339 & 0.7490 & 0.8706   \\
Mammography      & 0.8406 & 0.4620 & 0.4819 & 0.3158 & 0.5288   \\
Mnist            & 0.8870 & 0.9165 & 0.9083 & 0.9106 & 0.9109   \\
Musk             & 1.0000 & 0.9925 & 1.0000 & 1.0000 & 1.0000   \\
Optdigits        & 0.9356 & 0.3192 & 0.2817 & 0.2909 & 0.9566   \\
Parkinson        & 0.9210 & 0.9643 & 0.9628 & 0.9659 & 0.9659   \\
Pendigits        & 0.9360 & 0.0538 & 0.3199 & 0.0467 & 0.8848   \\
Pima             & 0.7449 & 0.7499 & 0.6457 & 0.6541 & 0.7912   \\
Satellite        & 0.8692 & 0.8659 & 0.8843 & 0.8770 & 0.8942   \\
Satimage-2       & 0.9703 & 0.9384 & 0.9641 & 0.9353 & 0.9804   \\
Shuttle          & 0.9819 & 0.1962 & 0.9934 & 0.8961 & 0.9991   \\
SpamBase         & 0.8413 & 0.8272 & 0.7837 & 0.7212 & 0.9335   \\
Speech           & 0.0584 & 0.0478 & 0.0502 & 0.0494 & 0.1189   \\
Thyroid          & 0.8626 & 0.2966 & 0.6301 & 0.3089 & 0.8770   \\
Vertebral        & 0.2854 & 0.2329 & 0.2564 & 0.2389 & 0.5657   \\
Vowels           & 0.4506 & 0.8592 & 0.8147 & 0.8985 & 0.8918   \\
Wbc              & 0.9742 & 0.7369 & 0.8001 & 0.7317 & 0.9341   \\
WDBC             & 1.0000 & 0.8423 & 0.9250 & 0.7377 & 1.0000   \\
Wilt             & 0.4543 & 0.6526 & 0.4637 & 0.6250 & 0.5926   \\
Wine             & 1.0000 & 0.9591 & 0.9577 & 0.9635 & 1.0000   \\
WPBC             & 0.5017 & 0.4675 & 0.4550 & 0.4361 & 0.5671   \\
Yeast            & 0.5416 & 0.5457 & 0.5195 & 0.5373 & 0.5281   \\ \midrule
Average          & 0.7437 & 0.6436 & 0.6812 & 0.6485 & \textbf{0.7849} \\ \bottomrule   
\end{tabular}
\end{table}

\clearpage
\subsection{A Simple Theoretical Illustration of Boundary Refinement}
\label{appendix:theoretical_analysis}

This appendix provides a stylized theoretical perspective on why detector-aware synthetic 
hard anomalies can improve the detection of true hard anomalies. While the main paper 
focuses on empirical validation across five detector families, a lightweight theoretical model 
helps clarify the geometric mechanism underlying the improvements observed in Fig.~5 
and Tables~1, 4–5.

\subsubsection{Motivation and Theoretical Abstraction}

\paragraph{Motivation.}
Empirically, we observe that most detection errors arise from \emph{hard anomalies}—
true anomalies that lie close to the decision boundary of the base detector \(f_t\). In contrast,
\emph{easy anomalies} that lie far from this boundary are already well-detected by \(f_t\)
without augmentation.  
LLM-DAS is explicitly designed to target this near-boundary region: it identifies borderline 
normal samples and transforms them into synthetic hard anomalies that reflect the 
failure modes of \(f_t\). These synthetic anomalies provide precisely the type of 
near-boundary supervision that one-class detectors inherently lack. The enhancement 
classifier \(\tilde f_t\) is trained on these hard synthetic samples, and the final fused detector
\[
F_t(x) = \mathrm{Norm}(f_t(x)) + \mathrm{Norm}(\tilde f_t(x))
\]
combines the strengths of both models.

\paragraph{Theoretical abstraction.}
To formalize this mechanism, we adopt the following stylized model:

- The set of true anomalies is conceptually decomposed into  
  \emph{easy anomalies} (far from the decision boundary of \(f_t\)) and  
  \emph{hard anomalies} (near the boundary, where most errors occur).

- LLM-DAS generates \emph{detector-aware synthetic hard anomalies} by transforming 
  normal samples that lie near the decision boundary of \(f_t\).

- If these synthetic anomalies geometrically cover the true hard anomalies within a
  small neighborhood, then any margin learned by the enhancement classifier \(\tilde f\)
  on the synthetic anomalies should transfer to the nearby true hard anomalies.

This abstraction leads to the proposition below, which formalizes this margin-transfer
intuition under a mild smoothness assumption on \(\tilde f\).

\subsubsection{Margin Transfer from Synthetic Hard Anomalies}

\begin{proposition}[Margin transfer via detector-aware synthetic anomalies]
\label{prop:margin_transfer}
Let $\mathcal{X}\subset\mathbb{R}^d$ be the input space and $\tau\in\mathbb{R}$ a decision 
threshold. Let $\tilde f:\mathcal{X}\to\mathbb{R}$ be a $B$-Lipschitz scoring function:
\[
|\tilde f(x)-\tilde f(x')|\le B\|x-x'\|_2 \quad \forall x,x'.
\]
Suppose a set of detector-aware synthetic anomalies 
$\mathcal{S}_{\mathrm{syn}} = \{x_{\mathrm{syn}}\}$ satisfies
\[
\tilde f(x_{\mathrm{syn}}) \ge \tau + \gamma, \qquad \gamma>0.
\]
Assume the set of true hard anomalies $\mathcal{D}_A$ is $\varepsilon$-covered by 
$\mathcal{S}_{\mathrm{syn}}$, i.e., for each $x_a\in\mathcal{D}_A$ there exists 
$x_{\mathrm{syn}}\in\mathcal{S}_{\mathrm{syn}}$ with
$\|x_a-x_{\mathrm{syn}}\|_2\le\varepsilon$.
Then each true hard anomaly satisfies
\[
\tilde f(x_a)\ge \tau + \gamma - B\varepsilon.
\]
In particular, if $B\varepsilon<\gamma$, then $\tilde f(x_a)>\tau$, and $x_a$ is
correctly detected under threshold $\tau$.
\end{proposition}

\textit{Proof.}
Fix any $x_a\in\mathcal{D}_A$. By $\varepsilon$-covering, choose 
$x_{\mathrm{syn}}\in\mathcal{S}_{\mathrm{syn}}$ with 
$\|x_a-x_{\mathrm{syn}}\|_2\le\varepsilon$.  
By Lipschitzness,
\[
|\tilde f(x_a)-\tilde f(x_{\mathrm{syn}})|\le B\varepsilon.
\]
Thus,
\[
\tilde f(x_a)\ge \tilde f(x_{\mathrm{syn}}) - B\varepsilon
\ge (\tau+\gamma) - B\varepsilon.
\]
If $B\varepsilon<\gamma$, then $\tilde f(x_a)>\tau$, completing the proof.

\paragraph{Interpretation.}
The proposition states that if synthetic hard anomalies are close (in geometry) to the true
hard anomalies ($\downarrow \varepsilon$), and if the enhancement classifier can enforce a margin ($\uparrow \gamma$) on the synthetic 
samples, then the true hard anomalies inherit a positive margin up to a Lipschitz slack ($B\varepsilon<\gamma$).  
This formalizes how near-boundary synthetic anomalies help the enhanced classifier 
learn a tighter decision boundary exactly where the base detector is weak.

\subsubsection{Remarks on Applicability and Model Classes}

\paragraph{Smooth vs.\ non-smooth enhancement classifiers.}
The margin-transfer proposition relies on a Lipschitz regularity condition on \(\tilde f\), 
which holds for smooth discriminative models such as linear or kernel SVMs.  
As shown in Fig.~\ref{fig:methods3} of Appendix~\ref{appendix:sensitivity analysis}, LLM-DAS yields clear performance improvements 
when \(\tilde f\) is instantiated as an SVM, consistent with the proposition.

Although tree ensembles are not Lipschitz in the Euclidean sense, they follow similar qualitative 
geometry: synthetic anomalies near the boundary tend to fall into similar leaves or 
partitions as true hard anomalies, providing comparable refinement. Extending the formal 
analysis to non-smooth models is an interesting direction for future work.

\subsubsection{Convergence and Stability of the Fused Detector}

The final fused detector is
\[
F_t(x)=\mathrm{Norm}(f_t(x))+\mathrm{Norm}(\tilde f_t(x)).
\]
Both components are bounded and monotone, so the enhancement step preserves the 
stability and asymptotic behavior of the base detector \(f_t\), while adding a bounded 
corrective term concentrated around the decision boundary. Thus the enhancement 
procedure behaves as a stable additive refinement:  
it reduces bias near hard anomalies while introducing only controlled variance determined 
by the number and quality of the synthetic samples and the complexity of \(\tilde f_t\).

\subsubsection{Justification for Local Boundary Refinement Mechanism in LLM-DAS}

A complex, global decision boundary may not be uniformly "tightened". The effectiveness of LLM-DAS does not rely on such global coverage but on a targeted, local refinement mechanism. Our conceptual justification is twofold and is strongly supported by our anomaly-type analysis.

\textbf{(1) Local Patching of Algorithmic Weak Spots:} The LLM in LLM-DAS acts as an automatic analyst that identifies a detector's structural weaknesses from its algorithm description. It then generates a program to create anomalies that specifically target these weaknesses. The goal is not to resample the entire boundary but to patch critical, high-error regions where the base detector $f_t$ is known to fail. The performance improvement stems from precisely this targeted effort.

\textbf{(2) Empirical Evidence via Anomaly-Type Alignment (Connecting to Appendix 6.15):} Our experiments in Appendix 6.15 (Table 19, Fig. 11-13) provide direct, empirical validation of this local-patching principle. They demonstrate that LLM-DAS automatically synthesizes anomalies whose distribution is closest to the specific anomaly types (e.g., dependency and local anomalies) that each base detector (e.g., IForest, OCSVM) struggles with most, as per established benchmarks like ADBench.

\begin{itemize}
    \item This is crucial: Different anomaly types challenge different segments of the decision boundary. By generating anomalies that align with a detector's known algorithmic blind spots, LLM-DAS proves it can automatically locate and target the most impactful "weak regions" of the boundary, even if they constitute only a portion of the whole.
    \item The significant performance gains we observe on these specific anomaly types confirm that patching these localized regions is both necessary and sufficient for overall improvement.
\end{itemize}

\textbf{(3) Preservation of Existing Strengths via Ensembling:} Furthermore, our framework is designed to be robust and non-destructive. As noted in our analysis, for anomaly types where the base detector already performs well (e.g., IForest on Cluster anomalies in Table 19), LLM-DAS maintains stable or slightly improved results. This indicates that our method enhances robustness without overfitting to the particular anomaly categories it uses for augmentation. This is a direct benefit of our ensemble design $F(x) = \text{Norm}(f_t(x)) + \text{Norm}(\tilde{f}_t(x))$. The original detector $f_t$ continues to dominate in regions it already models accurately, preserving its discriminative ability for "easy" anomalies, while the enhancement classifier $\tilde{f}_t$ extends the coverage toward harder cases targeted by the LLM.

Therefore, while a single set of synthetic anomalies may not cover the entire boundary, its coverage is highly strategic. It focuses on the regions that matter most for reducing the base detector's error, which is why the method delivers consistent and significant gains beyond purely empirical observations.

\subsubsection{Summary}

Together with Fig.~\ref{fig:visualization}, this stylized analysis provides a formal illustration of the empirical 
finding that LLM-DAS tightens the decision boundary around hard anomalies. The 
proposition explains why detector-aware synthetic anomalies bring improved separation 
precisely in the near-boundary region, while the ensemble structure ensures that easy 
anomalies remain correctly handled by the original detector.

\clearpage

\subsection{Detector-Aware Anomaly Synthesis Across Different Anomaly Types}
\label{appendix:anomaly_types}

\subsubsection{Motivation and Experimental Philosophy.}
A core challenge in tabular anomaly detection (TAD) is the diversity of anomaly types. The ground-truth anomaly type is not standardized across datasets, and
different datasets may correspond to different or mixed anomaly mechanisms, making real-world anomalies inherently complex and heterogeneous. Different detectors inherently excel at detecting different kinds of anomalies due to their underlying algorithms' assumptions. According to ADBench~\citep{han2022adbench}, many tabular detectors—such as IForest, OCSVM—struggle to detect ``Dependency'' and ``Local'' anomalies, while performing well on ``Clustered'' types.

The key innovation of LLM-DAS is that it does not require a priori specification of anomaly types. Instead, it positions the LLM as an ``algorithmist'' that reasons about a detector's intrinsic logic to generate code $\text{Code}^t$ for synthesizing anomalies that are specifically ``hard'' for that detector. A critical question thus arises: What types of anomalies does this detector-aware code actually produce?

To answer this scientifically, we need a controlled experimental setup where ground-truth anomaly types are known. We, therefore, adopt the established synthesis protocol from ADBench~\citep{han2022adbench} to create a benchmark where different, pure anomaly types are injected into the same normal data backbone. This allows us to rigorously analyze the relationship between a detector's weaknesses and the characteristics of the anomalies synthesized by its corresponding $\text{Code}^t$.

\subsubsection{Experimental Setup: Constructing Type-Specific Testbeds}
To analyze how LLM-DAS adapts to different anomaly mechanisms in a controlled and detector-aware manner, 
we follow the official ADBench~\citep{han2022adbench} protocol to construct four benchmark variants 
of the same real-world dataset (``Hepatitis''). 
The core idea is to build a generative model (e.g., Gaussian mixture model GMM) using the normal samples from the real-world
dataset (``Hepatitis'') and discard its original anomalies as we do not know their types. Then, We could generate normal samples and different types of anomalies based on their definitions by tweaking the generative model. The generation of normal samples is the same in all settings if not noted, and we provide the
generation process of four types of anomalies below. 

\paragraph{Unified anomaly-generation protocol.}
Each anomaly type is created using the following standardized procedures from ADBench~\citep{han2022adbench}:
\begin{itemize}
    \item \textbf{Local anomalies:} Generated by scaling the covariance matrix 
    of Gaussian mixture components ($\hat{\Sigma} \leftarrow \alpha \hat{\Sigma}$ with $\alpha=5$), 
    producing locally deviant samples within dense regions of the normal manifold.
    \item \textbf{Global anomalies:} They are more different from the normal data, drawn from a uniform distribution 
    $\mathrm{Unif}(\alpha \min(X_k), \alpha \max(X_k))$ with $\alpha=1.1$, where the boundaries are defined as the min and max
of an input feature, e.g., $k$-th feature $X_k$.
    \item \textbf{Dependency anomalies:} Produced by removing feature dependencies 
    modeled by Vine Copula, where the joint density is replaced by independent marginals, 
    breaking the correlation structure among features.
    \item \textbf{Clustered anomalies:} Formed by scaling the mean vector 
    of the normal data ($\hat{\mu} \leftarrow \alpha \hat{\mu}$ with $\alpha=5$), 
    creating compact abnormal clusters separated from the normal distribution.
\end{itemize}

For each anomaly type~$\tau \in \{\text{Local}, \text{Global}, \text{Dependency}, \text{Clustered}\}$,  
we obtain a dataset consisting of: $\{
D_{\mathrm{train}}^{(\tau)} \quad \text{(normal only)}, 
D_{\mathrm{test}}^{(\tau)} \quad \text{(normal + anomalies of type $\tau$)}
\}$.
Importantly, the synthesized anomalies produced by LLM-DAS and normal samples are kept the same across all variants,  
ensuring that the only varying factor is the anomaly type in the test set.

\subsubsection{Analysis Procedure for LLM-DAS}

Given a detector $t$ (e.g., IForest),  
we examine how the LLM-generated synthesis code $\text{Code}^t$ adapts to different anomaly types:

\begin{enumerate}
    \item \textbf{Train the base detector.}  
    Train $f_t$ on the normal-only dataset $D_{\mathrm{train}}^{(\tau)}$.

    \item \textbf{Generate detector-specific synthetic anomalies.}  
    We instantiate the LLM-generated, detector-specific synthesis code $\text{Code}^t$ using $D_{\mathrm{train}}^{(\tau)}$ and $f_t$. This produces a set of synthetic anomalies $A(D_{\mathrm{train}}^{(\tau)}, t)$:

    \item \textbf{Measure distributional similarity. (Table~\ref{appendix:table_anomaly_type})}  
    Compute the Wasserstein distance between  
    $A(D_{\mathrm{train}}^{(\tau)}, t)$  
    and each of the four ground-truth anomaly-type distributions in $D_{\mathrm{test}}^{(\tau)}$.

    \item \textbf{Evaluate performance improvement. (Table~\ref{appendix:table_anomaly_type})}  
    We use $A(D_{\mathrm{train}}^{(\tau)}, t)$ to augment $D_{\mathrm{train}}^{(\tau)}$ and train a binary classifier $\tilde{f}_t$. We then fuse the $\tilde{f}_t$ and $f_t$ to obtain the final detector $F_t$, and evaluate performance  
    on $D_{\mathrm{test}}^{(\tau)}$ to obtain the improvement $\Delta$.

    \item \textbf{Kernel Density Estimation (KDE). (Fig.~\ref{appendix:anomaly_type_kde_IForest}–\ref{appendix:anomaly_type_kde_LOF})}
    We visualize the kernel density estimation (KDE) of anomaly scores before and after applying LLM-DAS, illustrating how the synthesized anomalies reshape the decision boundary.
\end{enumerate}

Together, these analyses enable a detailed examination of 
\emph{(i)} what types of anomalies LLM-DAS tends to generate, 
and \emph{(ii)} how such generated samples influence the detector’s behavior 
and decision-space separation under different anomaly mechanisms.

\subsubsection{Original detectors' priori on different types of anomalies.}
According to ADBench~\citep{han2022adbench}, many tabular detectors—such as IForest, 
OCSVM—struggle to detect ``Dependency'' and ``Local'' anomalies, 
while performing well on ``Clustered'' types.  
To further broaden our observations, we also evaluate LLM-DAS paired with LOF, 
since LOF shows the opposite pattern: it performs better on ``Dependency'' and ``Local'' anomalies 
but poorly on ``Clustered'' anomalies.  
Although LOF was not part of our main benchmark suite, 
it serves as a valuable complementary case for examining 
whether LLM-DAS adaptively targets detector’s weakest anomaly type.


\subsubsection{Key findings: LLM-DAS automatically targets the anomaly types that each detector performs worst on}
Table~\ref{appendix:table_anomaly_type} shows that the base detectors exhibit distinct weaknesses across different anomaly types: 
IForest and OCSVM perform poorly on \emph{Dependency}- and \emph{Local}-type anomalies, 
while LOF struggles on \emph{Cluster}-type anomalies.
These findings are consistent with the conclusions of ADBench~\citep{han2022adbench}, 
confirming that each detector has its own characteristic sensitivity to specific anomaly structures.

As shown in Table~\ref{appendix:table_anomaly_type}, the anomalies synthesized by LLM-DAS 
are distributionally closer to the anomaly types on which the corresponding detectors underperform—
\emph{Dependency} and \emph{Local} for IForest and OCSVM, and \emph{Cluster} for LOF.
This observation aligns with our core claim that LLM-DAS is not generating arbitrary or 
predefined anomaly types, but instead dynamically adapts its synthesis behavior 
based on the weaknesses of the base detector.
In other words, LLM-DAS learns to identify and reproduce the challenging regions 
within each detector’s decision space rather than following a fixed anomaly taxonomy.

\begin{figure}[h!]
    \centering
    \begin{subfigure}[b]{0.24\textwidth}
        \centering
        \includegraphics[width=\textwidth]{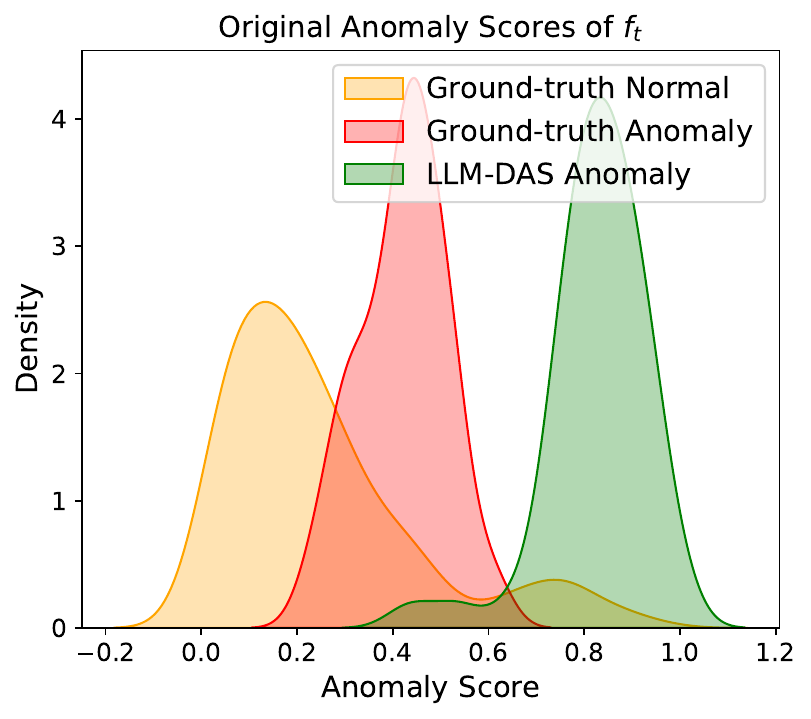}
        \caption{$f_t$ on Dependency}
    \end{subfigure}
    \begin{subfigure}[b]{0.24\textwidth}
        \centering
        \includegraphics[width=\textwidth]{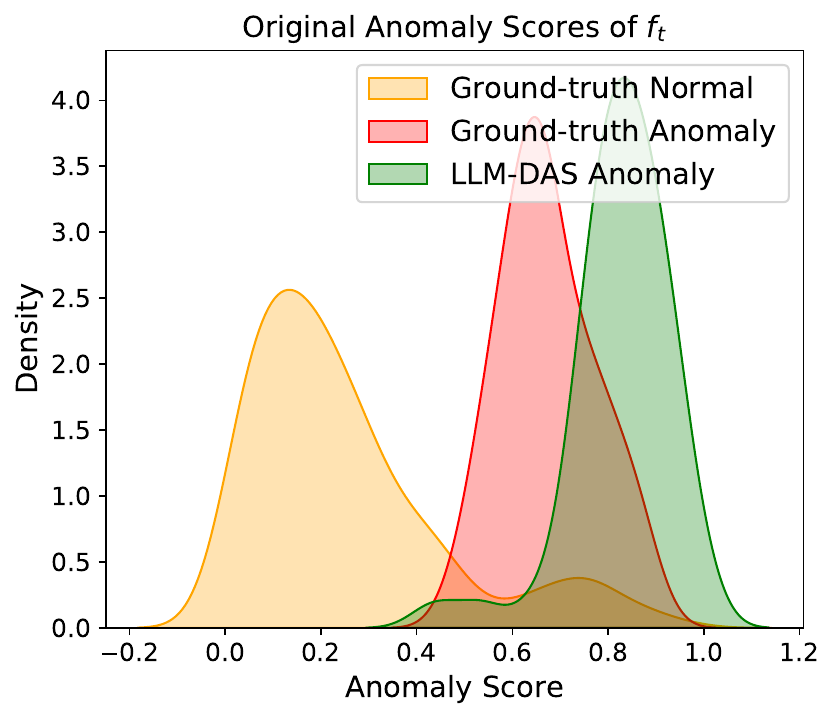}
        \caption{$f_t$ on Local}
    \end{subfigure}
    \begin{subfigure}[b]{0.24\textwidth}
        \centering
        \includegraphics[width=\textwidth]{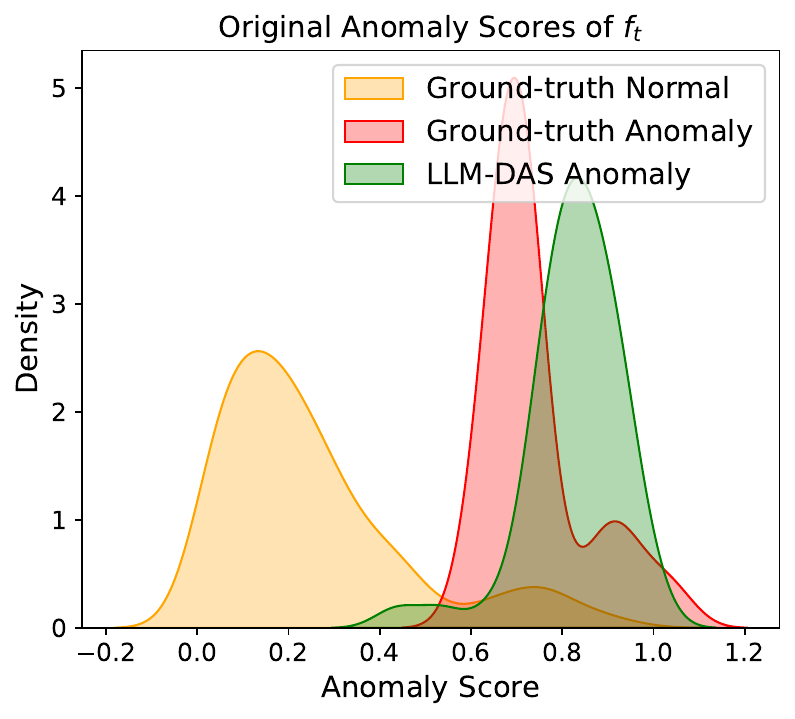}
        \caption{$f_t$ on Cluster}
    \end{subfigure}
    \begin{subfigure}[b]{0.24\textwidth}
        \centering
        \includegraphics[width=\textwidth]{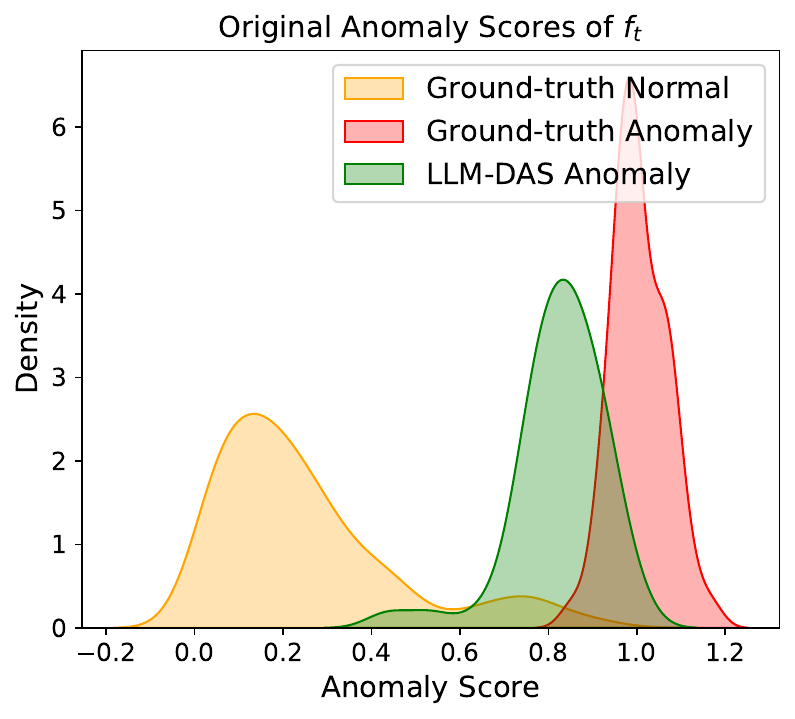}
        \caption{$f_t$ on Global}
    \end{subfigure}
    
    \begin{subfigure}[b]{0.24\textwidth}
        \centering
        \includegraphics[width=\textwidth]{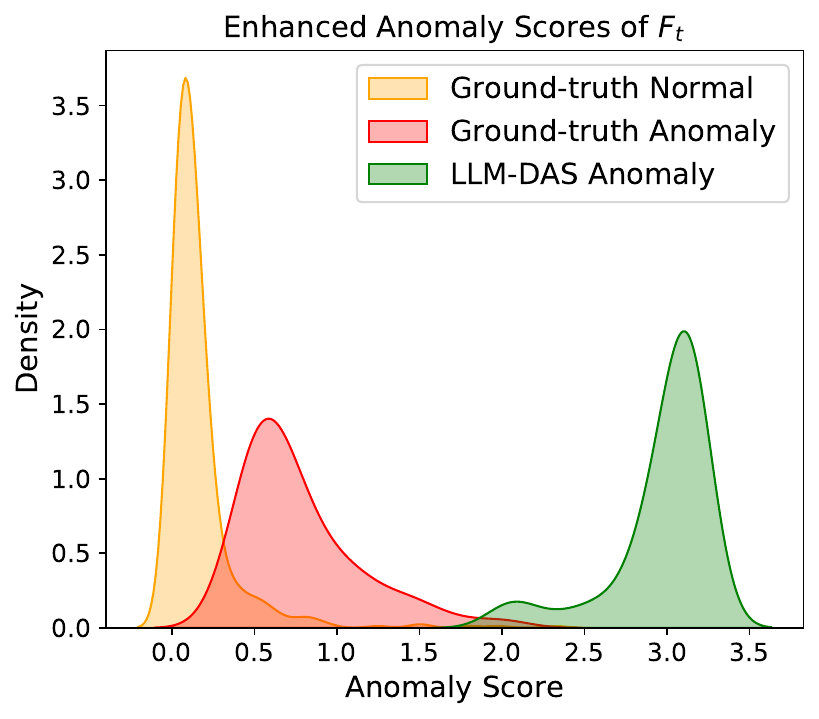}
        \caption{$F_t$ on Dependency}
    \end{subfigure}
    \begin{subfigure}[b]{0.24\textwidth}
        \centering
        \includegraphics[width=\textwidth]{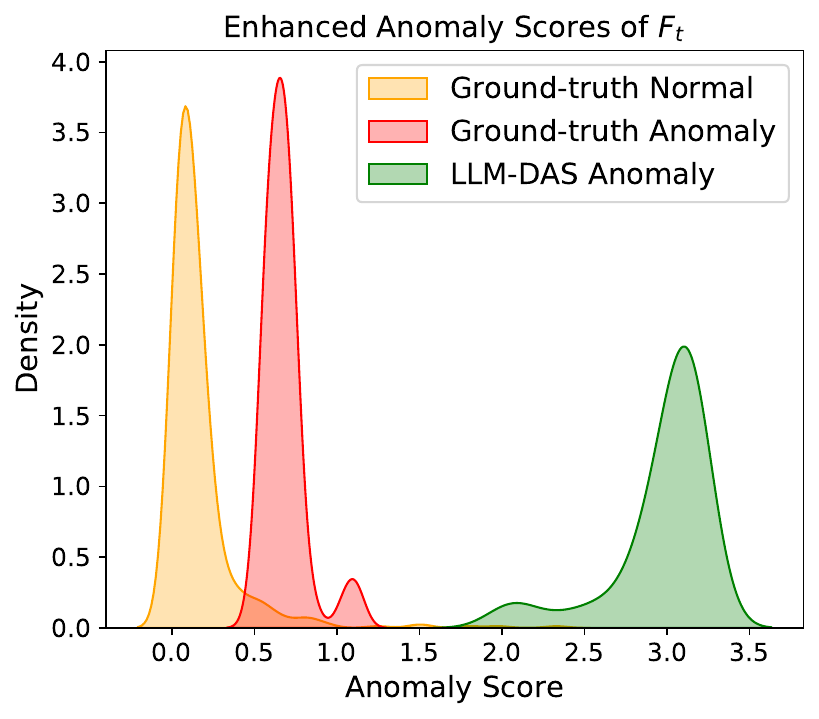}
        \caption{$F_t$ on Local}
    \end{subfigure}
    \begin{subfigure}[b]{0.24\textwidth}
        \centering
        \includegraphics[width=\textwidth]{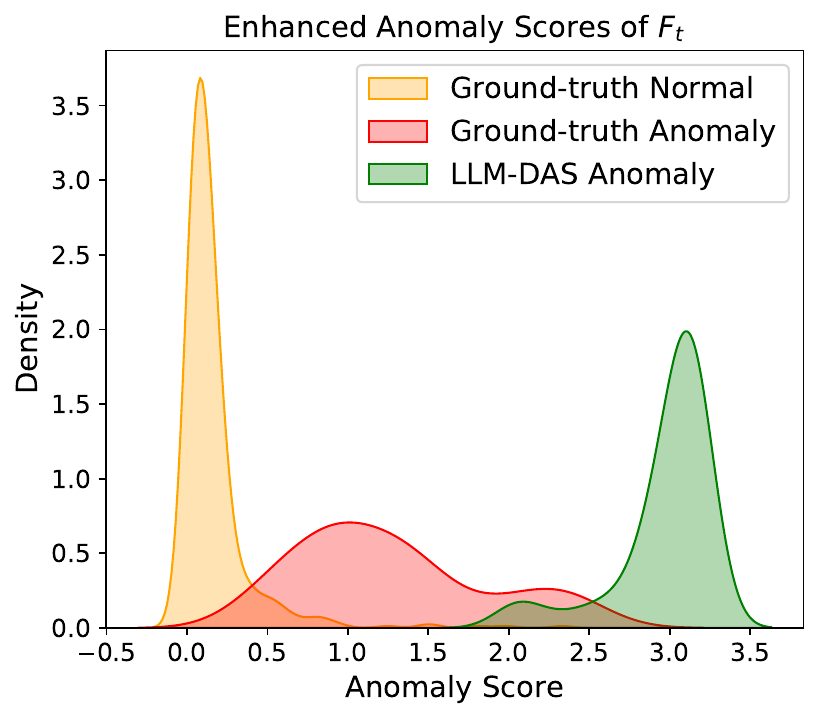}
        \caption{$F_t$ on Cluster}
    \end{subfigure}
    \begin{subfigure}[b]{0.24\textwidth}
        \centering
        \includegraphics[width=\textwidth]{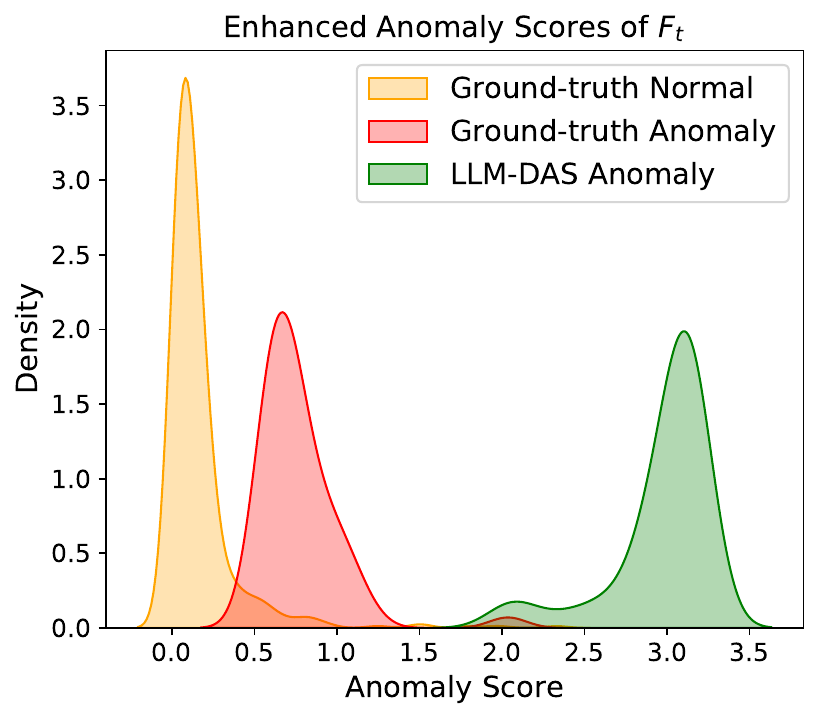}
        \caption{$F_t$ on Global}
    \end{subfigure}
    \caption{Kernel density estimation (KDE) of anomaly scores from the source IForest detector $f_t$ and the LLM-DAS–enhanced detector $F_t$.
    The first row (subplots a–d) shows the original detector $f_t$, and the second row (e–h) shows the corresponding results after applying LLM-DAS.
    Each column corresponds to one anomaly type (\emph{Dependency}, \emph{Local}, \emph{Cluster}, \emph{Global}).
    The enhanced detector $F_t$ exhibits improved boundary separation, especially for the \textbf{\emph{Dependency} and \emph{Local}} types, 
    whose characteristics are most similar to the synthesized anomalies.}
    \label{appendix:anomaly_type_kde_IForest}
\end{figure}

\begin{figure}[h!]
    \centering
    \begin{subfigure}[b]{0.24\textwidth}
        \centering
        \includegraphics[width=\textwidth]{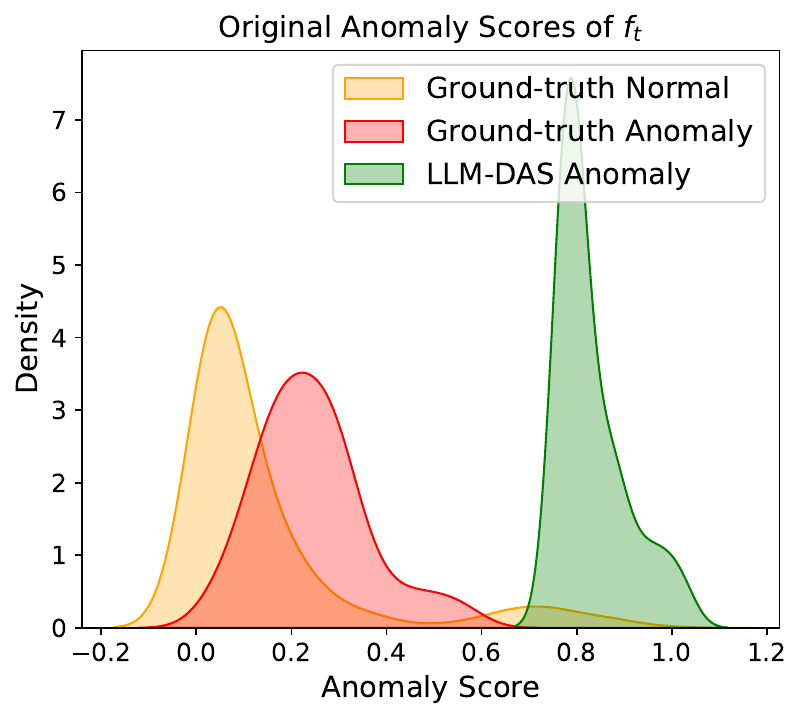}
        \caption{$f_t$ on Dependency}
    \end{subfigure}
    \begin{subfigure}[b]{0.24\textwidth}
        \centering
        \includegraphics[width=\textwidth]{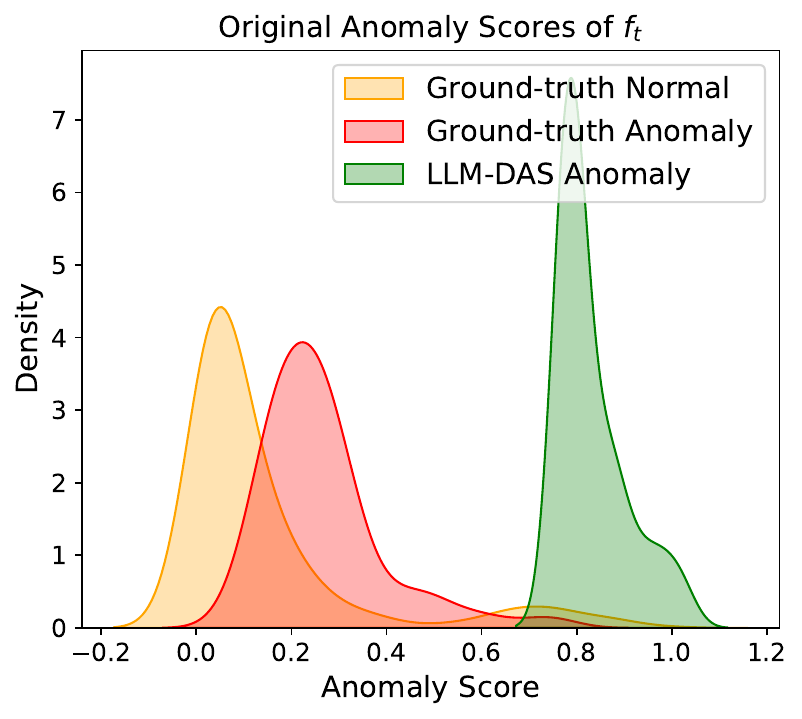}
        \caption{$f_t$ on Local}
    \end{subfigure}
    \begin{subfigure}[b]{0.24\textwidth}
        \centering
        \includegraphics[width=\textwidth]{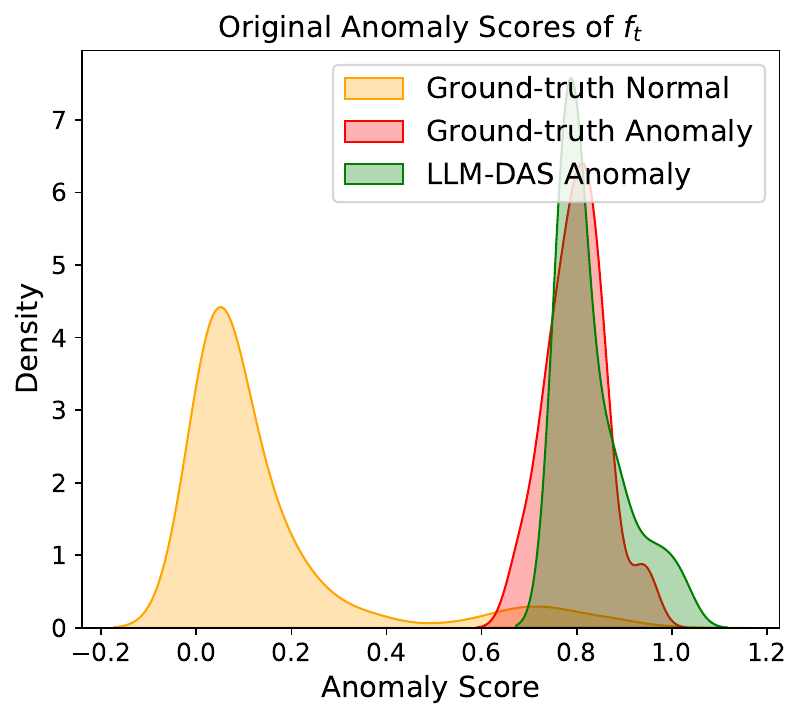}
        \caption{$f_t$ on Cluster}
    \end{subfigure}
    \begin{subfigure}[b]{0.24\textwidth}
        \centering
        \includegraphics[width=\textwidth]{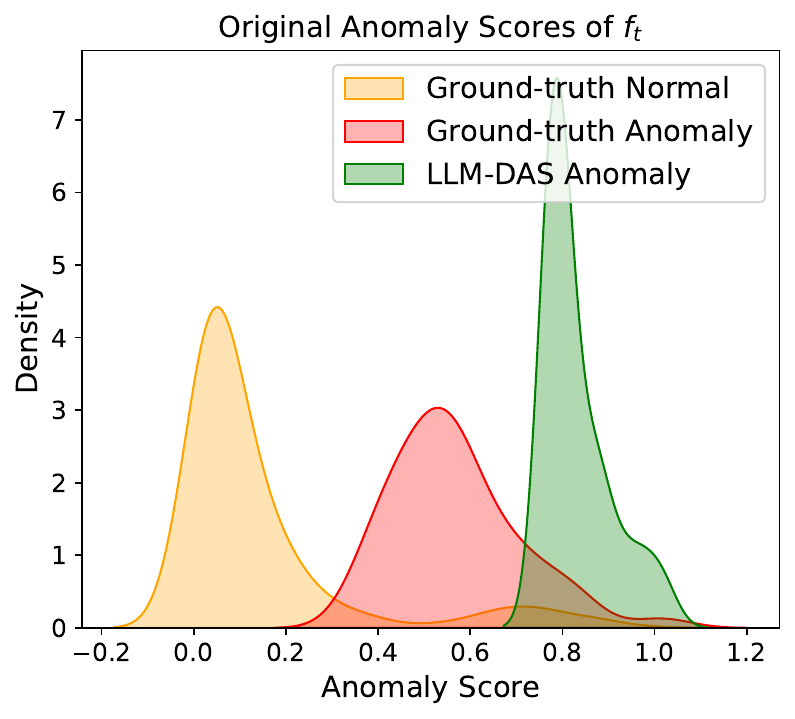}
        \caption{$f_t$ on Global}
    \end{subfigure}
    
    \begin{subfigure}[b]{0.24\textwidth}
        \centering
        \includegraphics[width=\textwidth]{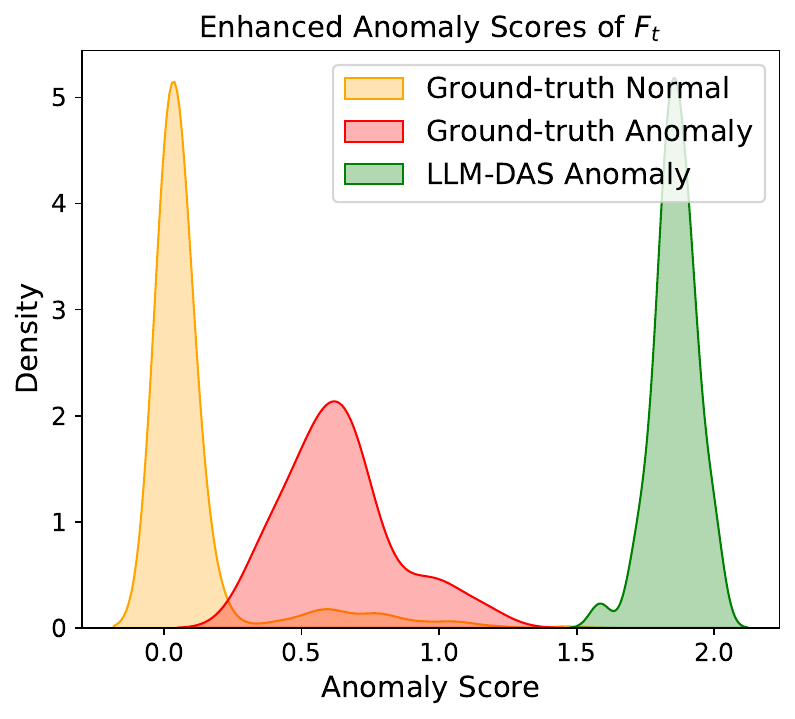}
        \caption{$F_t$ on Dependency}
    \end{subfigure}
    \begin{subfigure}[b]{0.24\textwidth}
        \centering
        \includegraphics[width=\textwidth]{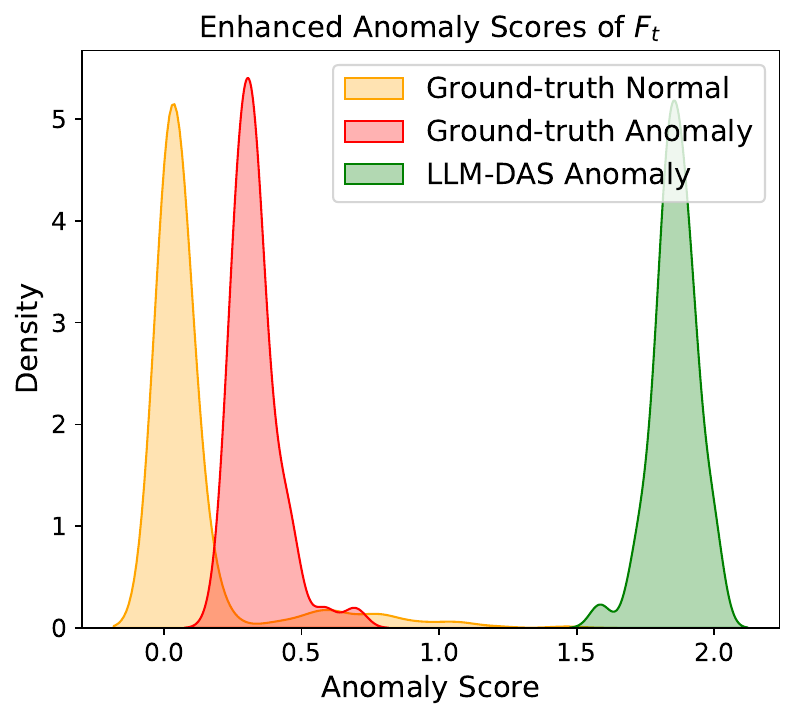}
        \caption{$F_t$ on Local}
    \end{subfigure}
    \begin{subfigure}[b]{0.24\textwidth}
        \centering
        \includegraphics[width=\textwidth]{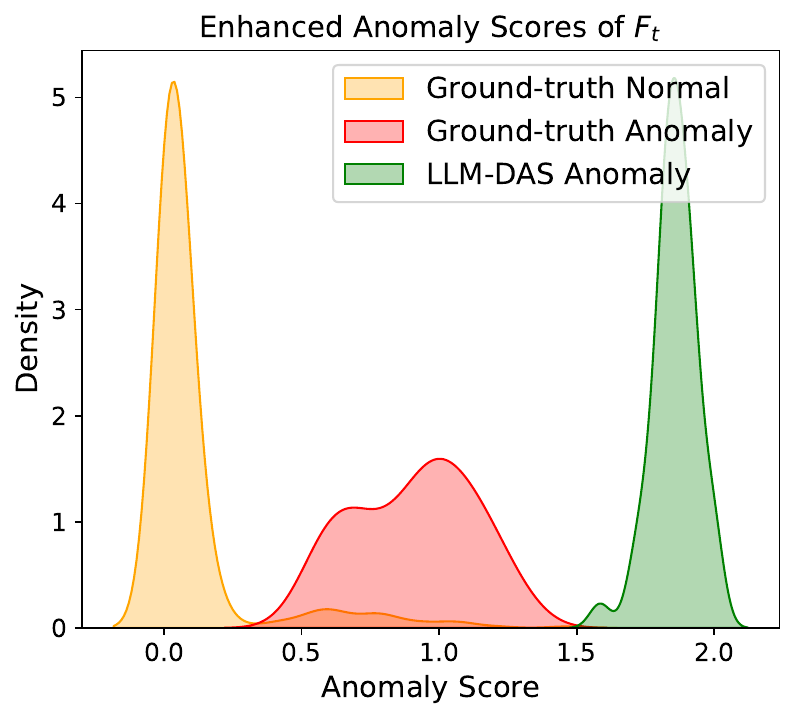}
        \caption{$F_t$ on Cluster}
    \end{subfigure}
    \begin{subfigure}[b]{0.24\textwidth}
        \centering
        \includegraphics[width=\textwidth]{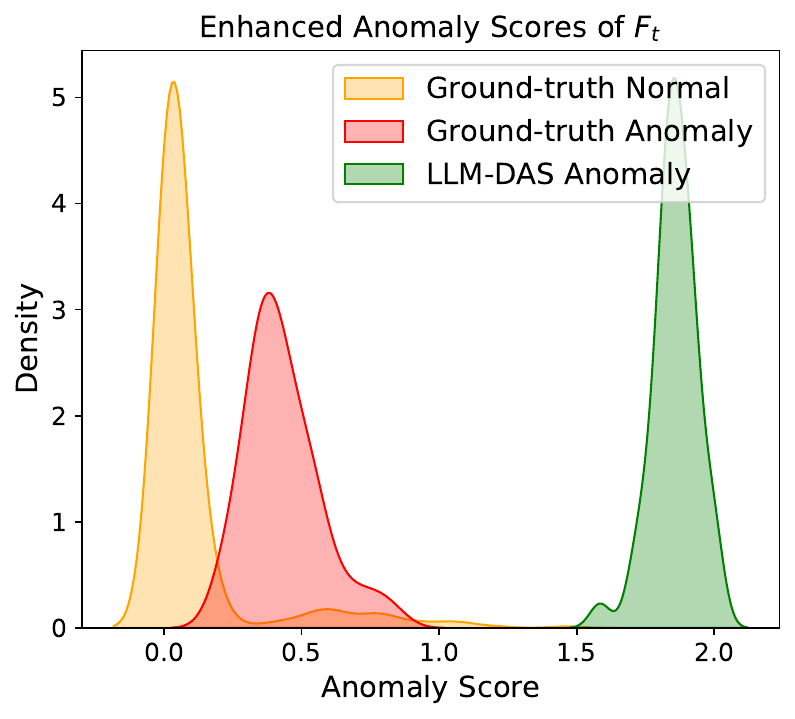}
        \caption{$F_t$ on Global}
    \end{subfigure}
    \caption{Kernel density estimation (KDE) of anomaly scores from the source OCSVM detector $f_t$ and the LLM-DAS–enhanced detector $F_t$.
    The first row (subplots a–d) shows the original detector $f_t$, and the second row (e–h) shows the corresponding results after applying LLM-DAS.
    Each column corresponds to one anomaly type (\emph{Dependency}, \emph{Local}, \emph{Cluster}, \emph{Global}).
    The enhanced detector $F_t$ exhibits improved boundary separation, especially for the \textbf{\emph{Dependency} and \emph{Local}} types, 
    whose characteristics are most similar to the synthesized anomalies.}
    \label{appendix:anomaly_type_kde_OCSVM}
\end{figure}

\begin{figure}[h!]
    \centering
    \begin{subfigure}[b]{0.24\textwidth}
        \centering
        \includegraphics[width=\textwidth]{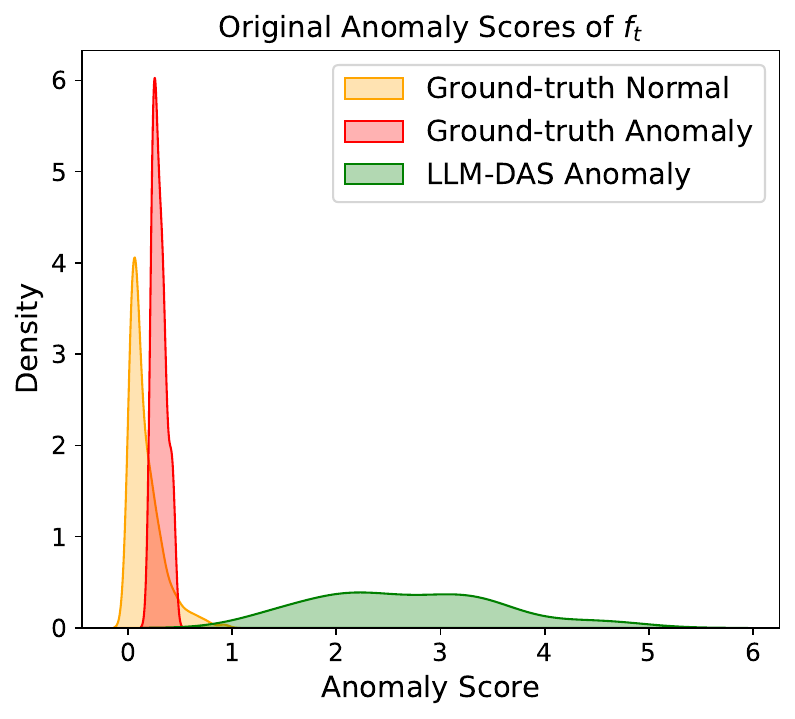}
        \caption{$f_t$ on Dependency}
    \end{subfigure}
    \begin{subfigure}[b]{0.24\textwidth}
        \centering
        \includegraphics[width=\textwidth]{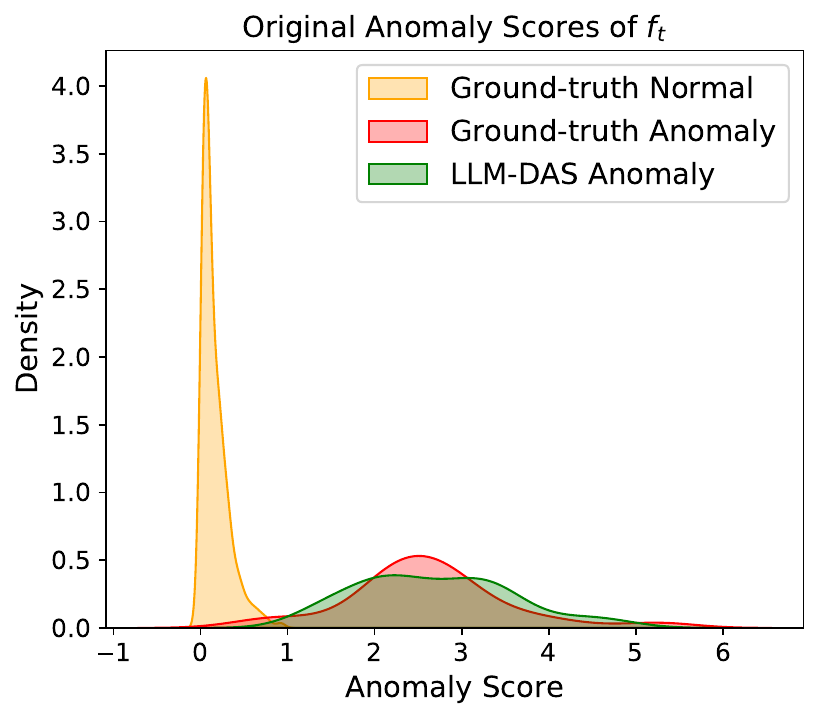}
        \caption{$f_t$ on Local}
    \end{subfigure}
    \begin{subfigure}[b]{0.24\textwidth}
        \centering
        \includegraphics[width=\textwidth]{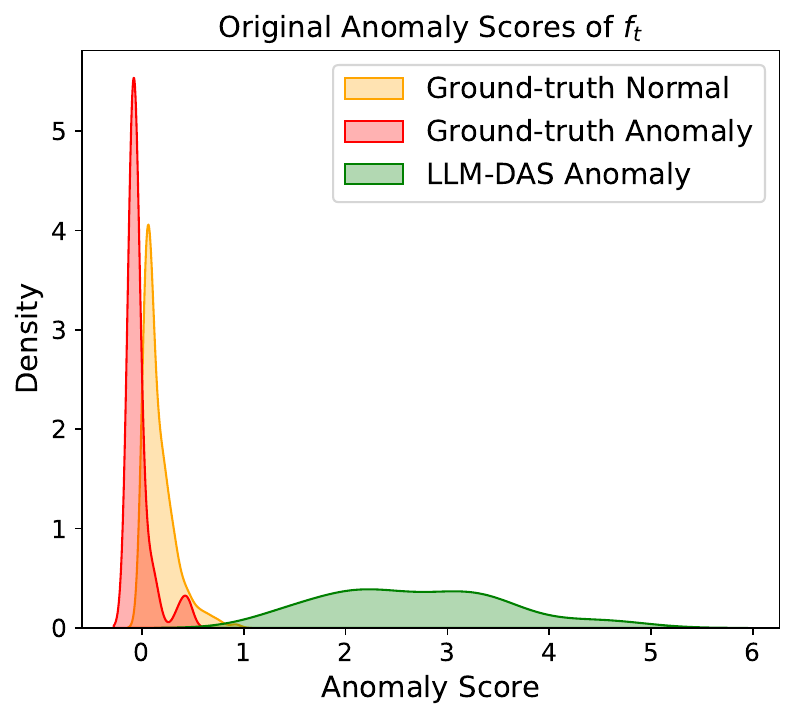}
        \caption{$f_t$ on Cluster}
    \end{subfigure}
    \begin{subfigure}[b]{0.24\textwidth}
        \centering
        \includegraphics[width=\textwidth]{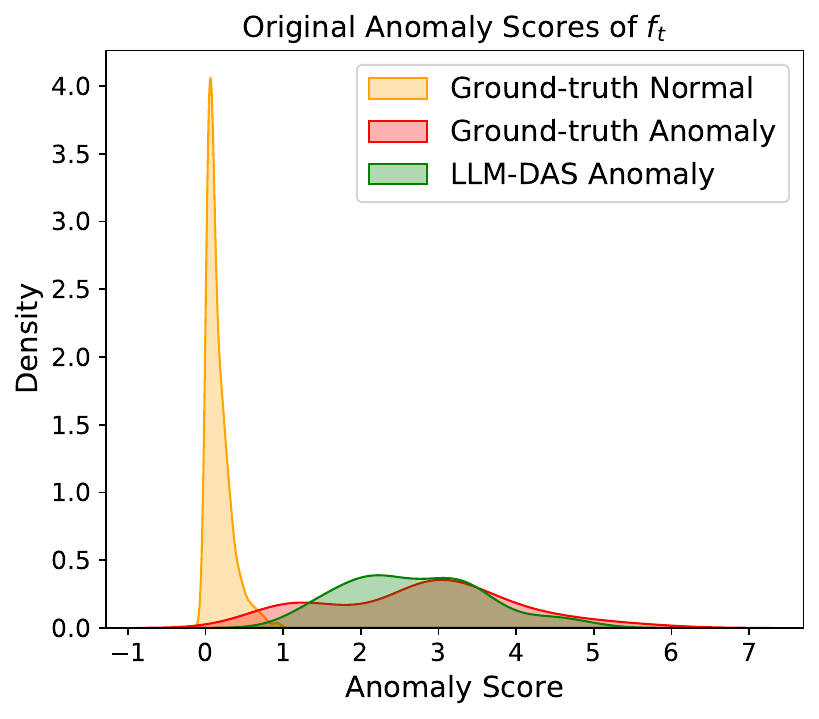}
        \caption{$f_t$ on Global}
    \end{subfigure}
    
    \begin{subfigure}[b]{0.24\textwidth}
        \centering
        \includegraphics[width=\textwidth]{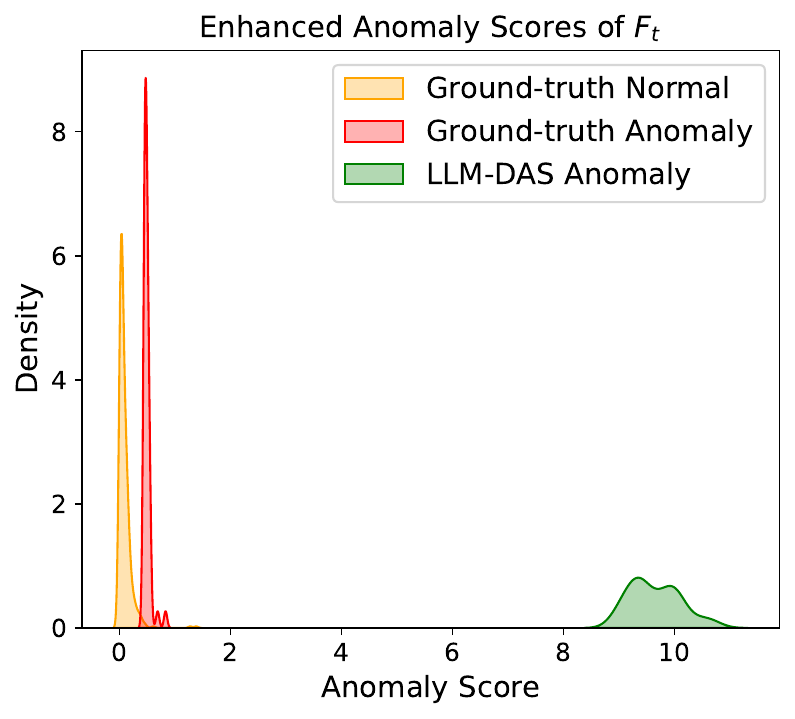}
        \caption{$F_t$ on Dependency}
    \end{subfigure}
    \begin{subfigure}[b]{0.24\textwidth}
        \centering
        \includegraphics[width=\textwidth]{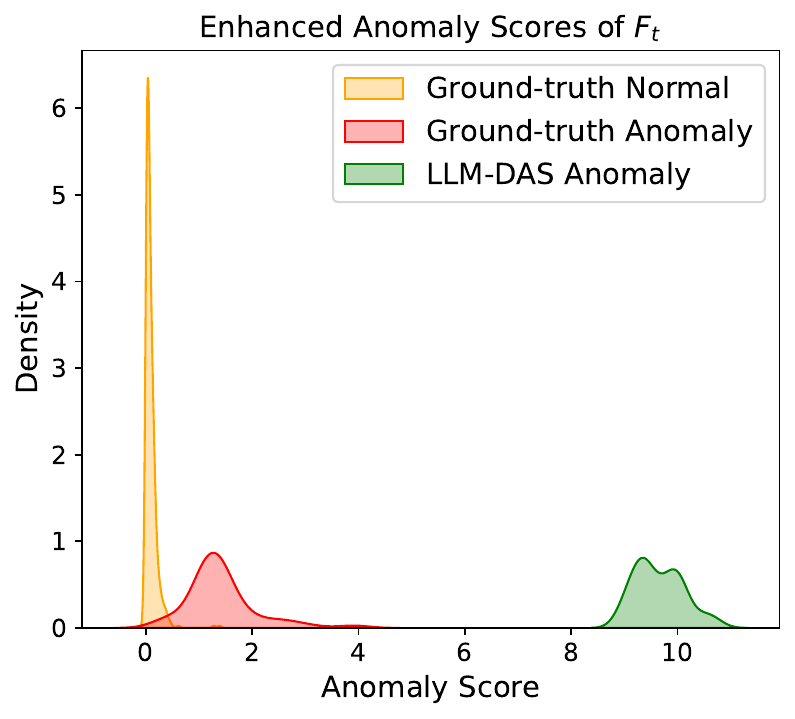}
        \caption{$F_t$ on Local}
    \end{subfigure}
    \begin{subfigure}[b]{0.24\textwidth}
        \centering
        \includegraphics[width=\textwidth]{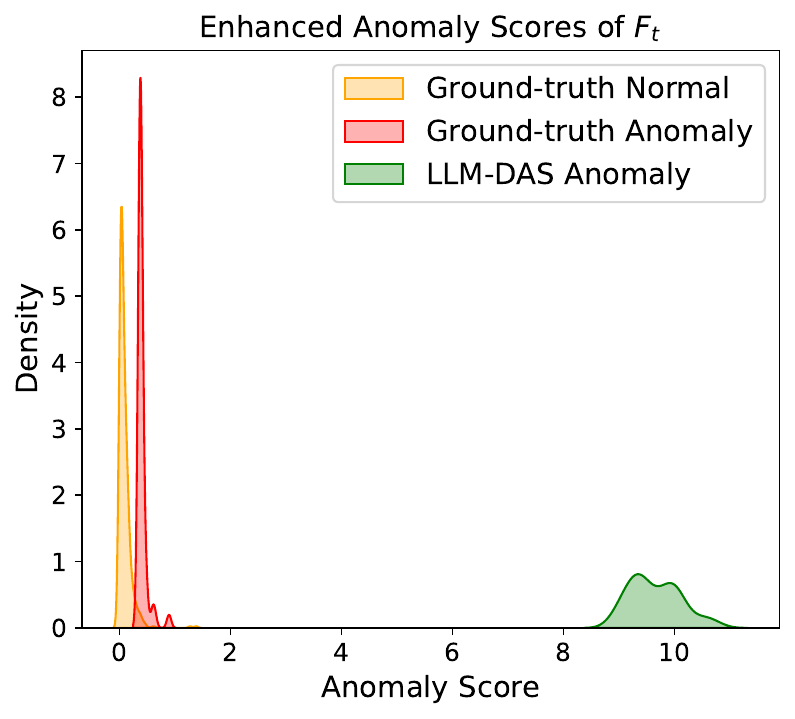}
        \caption{$F_t$ on Cluster}
    \end{subfigure}
    \begin{subfigure}[b]{0.24\textwidth}
        \centering
        \includegraphics[width=\textwidth]{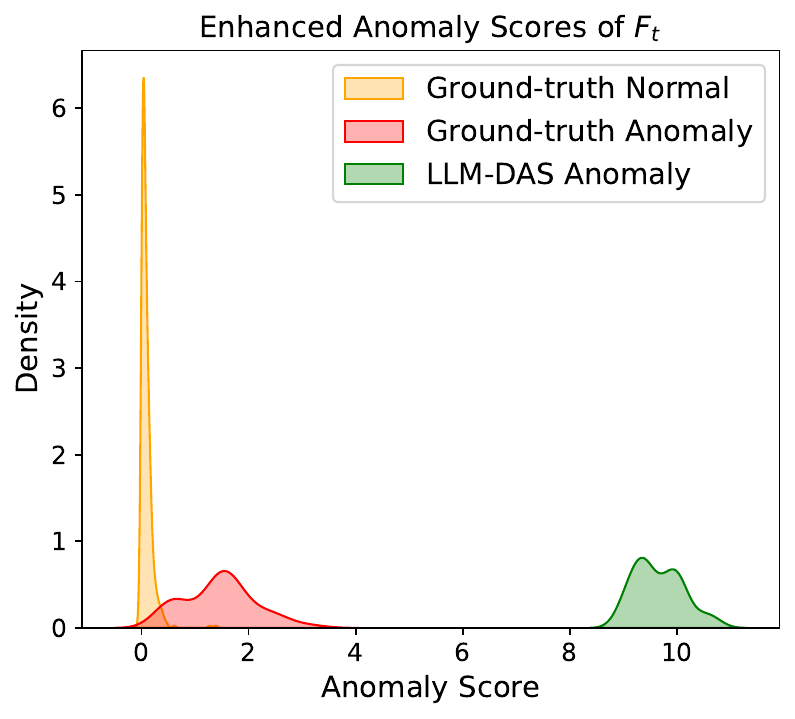}
        \caption{$F_t$ on Global}
    \end{subfigure}
    \caption{Kernel density estimation (KDE) of anomaly scores from the source LOF detector $f_t$ and the LLM-DAS–enhanced detector $F_t$.
    The first row (subplots a–d) shows the original detector $f_t$, and the second row (e–h) shows the corresponding results after applying LLM-DAS.
    Each column corresponds to one anomaly type (\emph{Dependency}, \emph{Local}, \emph{Cluster}, \emph{Global}).
    The enhanced detector $F_t$ exhibits improved boundary separation, especially for the \textbf{\emph{Cluster}} types, 
    whose characteristics are most similar to the synthesized anomalies.}
    \label{appendix:anomaly_type_kde_LOF}
\end{figure}

Furthermore, combining the quantitative and visual evidence from 
Table~\ref{appendix:table_anomaly_type} and Fig.~\ref{appendix:anomaly_type_kde_IForest}–\ref{appendix:anomaly_type_kde_LOF}, 
we observe that the anomalies synthesized by LLM-DAS 
significantly improve performance and enhance score separation between normal and abnormal samples 
on those hard anomaly types where the base detectors originally perform poorly.
This confirms that LLM-DAS effectively addresses the intrinsic blind spots of each detector.  
Meanwhile, for anomaly types that the base detectors already handle well, 
LLM-DAS maintains stable or slightly improved performance, 
demonstrating its robustness and balanced ensemble design.

Overall, these results validate that LLM-DAS adaptively captures 
the detector-specific difficulty patterns across heterogeneous anomaly types, 
thereby inherently accounting for anomaly diversity rather than ignoring it.

\paragraph{Discussion and Future Work.}
These findings highlight that LLM-DAS does not generate arbitrary anomalies, 
but adaptively learns to emphasize those challenging for the base detector.  
Meanwhile, our ensemble strategy ensures that 
the original detector $f_t$ continues to dominate detection of easy anomaly types 
it already handles well, preserving generalization (see Appendix~6.14).  
In future work, we plan to design a controllable synthesis module 
that allows users to explicitly specify target anomaly types 
(e.g., \emph{dependency} or \emph{clustered}) 
to further enhance interpretability and diagnostic use.

\begin{table}
\caption{
Detector-aware anomaly synthesis: Wasserstein distance versus performance improvement.
For each detector and each canonical anomaly type 
(\textit{Local}, \textit{Global}, \textit{Dependency}, \textit{Clustered}), we report:
(i) the Wasserstein distance between the LLM-DAS–generated anomalies and the corresponding ground-truth anomaly-type distribution,
(ii) the base detector performance~($f_t$), 
(iii) the enhanced performance after applying LLM-DAS~($F_t$), and
(iv) the improvement $\Delta$ of $F_t$ over $f_t$.
Across detectors, the anomaly types with the \emph{smallest} Wasserstein distances (highlighted in blue) 
consistently correspond to the \emph{largest} performance gains, demonstrating that LLM-DAS 
automatically synthesizes anomalies that resemble each detector's weakest anomaly mechanism.
}
\label{appendix:table_anomaly_type}
\centering
\scriptsize
\begin{tabular}{cccccc}
\toprule
Detector type & Anomaly type & Wasserstein distance $\downarrow$ & Base detector ($f_t$) $\uparrow$ & LLM-DAS ($F_t$) $\uparrow$ & Improv. ($\Delta$) $\uparrow$  \\ \midrule
\multirow{4}{*}{\centering IForest}  & Dependency   & \cellcolor{blue!30}\textbf{0.942}                & 0.419   & 0.701    & \cellcolor{blue!30}\textbf{0.282}    \\
         & Local        & \cellcolor{blue!30}\textbf{0.958}                & 0.575   & 0.717    & \cellcolor{blue!30}\textbf{0.142}    \\
         & Cluster      & 1.694                & 0.781   & 0.797    & 0.016    \\
         & Global       & 1.824                & 0.701   & 0.722    & 0.021    \\ \midrule
\multirow{4}{*}{\centering OCSVM}    & Dependency   & \cellcolor{blue!30}\textbf{0.651}                & 0.246   & 0.727    & \cellcolor{blue!30}\textbf{0.481}    \\
         & Local        & \cellcolor{blue!30}\textbf{0.681}                & 0.560   & 0.696    & \cellcolor{blue!30}\textbf{0.136}    \\
         & Cluster      & 1.630                & 0.789   & 0.803    & 0.014    \\
         & Global       & 1.779                & 0.719   & 0.722    & 0.002    \\ \midrule
\multirow{4}{*}{\centering LOF}      & Dependency   & 1.117                & 0.777   & 0.791    & 0.014    \\
         & Local        & 1.877                & 0.737   & 0.761    & 0.024    \\
         & Cluster      & \cellcolor{blue!30}\textbf{0.897}                & 0.141   & 0.757    & \cellcolor{blue!30}\textbf{0.616}    \\
         & Global       & 1.723                & 0.647   & 0.676    & 0.029    \\
\bottomrule
\end{tabular}
\end{table}

\clearpage

\subsection{Hardness-Controlled Synthetic Anomalies and Side-Effects Analysis}
\label{appendix:hardness_control}

\paragraph{Motivation.}
An important question is 
(i)~whether LLM-DAS can generate anomalies with \emph{different levels of hardness}, 
and (ii)~whether overly hard synthetic anomalies may negatively impact generalization in 
some datasets.
To systematically study these aspects, we extend the synthesis prompt with an explicit
\emph{hardness-level interface} (prompt is detailed in Prompt~\ref{prompt:varying_difficulty}) and introduce a quantitative metric for anomaly hardness
based on Wasserstein distances.
This appendix provides the experimental protocol and detailed results that support our claim.

\paragraph{Hardness-level interface and experimental setup.}
We modify the original LLM-DAS prompt so that the LLM is required to produce synthesis
code that accepts a discrete hardness level \texttt{"High"}, \texttt{"Medium"}, or \texttt{"Low"}.
Intuitively:
\begin{itemize}
    \item \textbf{High} aims to generate anomalies that are close to the normal-data manifold
    and are therefore \emph{hard to detect};
    \item \textbf{Medium} generates anomalies moderately separated from normal data;
    \item \textbf{Low} generates anomalies further away, which are \emph{easier} to detect.
\end{itemize}
The original LLM-DAS implementation in the main paper corresponds closely to the
\textbf{High-hardness} behavior, i.e., its synthesized anomalies typically lie near the
decision boundary and are thus challenging for the base detector.

We instantiate this interface on two representative base detectors:
PCA and IForest.
For each dataset, we compare:
\begin{itemize}
    \item the base detector alone (PCA / IForest),
    \item the original LLM-DAS (denoted ``LLM-DAS (Ours)''),
    \item LLM-DAS with hardness levels High / Medium / Low 
    (denoted ``LLM-DAS (Hardness Level)'').
\end{itemize}

\paragraph{Performance effect of hardness control.}
Table~\ref{tab:hardness_perf} reports the performance of PCA and IForest-based LLM-DAS
under different hardness settings.
Rows highlighted in \textcolor{blue!80}{blue} correspond to datasets where the original
LLM-DAS already improves over the base detector.
Rows highlighted in \textcolor{red!80}{red} correspond to \emph{fail cases} where the original
LLM-DAS underperforms the base detector.

Several consistent patterns emerge:
\begin{itemize}
    \item \textbf{Original $\approx$ High hardness.}
    For most datasets, the original LLM-DAS and the High-hardness variant achieve very similar performance.
    This confirms that the default policy learned by LLM-DAS indeed focuses on generating
    \emph{hard} anomalies.
    \item \textbf{Robust gains across hardness levels.}
    On the majority of datasets (blue rows), all hardness levels (High/Medium/Low) outperform the base detector,
    and the original (High) version is often the best or close to the best.
    This shows that LLM-DAS is not fragile with respect to the exact hardness level,
    as long as synthetic anomalies inhabit a reasonably informative region near the boundary.
    \item \textbf{Failure cases and recovery by lowering hardness.}
    In a few datasets (red rows), the original LLM-DAS underperforms the base detector.
    Crucially, these failures are \emph{reversible}:
    by lowering the hardness level (Medium or Low), performance recovers to the baseline or even exceeds it.
    For example, for PCA on \textsf{lympho}, the original LLM-DAS yields $0.989$,
    while Medium/Low hardness both achieve $1.000$;
    for IForest on \textsf{cardio}, the original LLM-DAS yields $0.622$, while Low hardness reaches $0.718$.
\end{itemize}
These results empirically validate that (i)~LLM-DAS can generate anomalies at different difficulty levels,
and (ii)~the hardness control interface provides a simple and effective knob to mitigate rare side effects.

\begin{table}[t!]
\centering
\scriptsize
\caption{
AUC-PR performance of LLM-DAS with different hardness levels of anomaly synthesis.
We extend the synthesis prompt (detailed in Prompt~\ref{prompt:varying_difficulty}) so that the LLM-generated code includes a discrete hardness parameter
(\texttt{High}, \texttt{Medium}, \texttt{Low}). 
Rows in \textcolor{blue!80}{blue} indicate datasets where the original LLM-DAS improves over the base detector.
Rows in \textcolor{red!80}{red} indicate cases where the original LLM-DAS underperforms the base detector,
but performance can be recovered or improved by lowering the hardness.
}
\label{tab:hardness_perf}
\begin{tabular}{cccccc}
\toprule
\multirow{2}{*}{\centering Datasets}    & Base detector & \multirow{2}{*}{\centering +LLM-DAS (Ours)} & \multicolumn{3}{c}{+LLM-DAS (Hardness Level)} \\ \cline{4-6}
            & PCA     &         & High  & Medium   & Low    \\ \midrule
\rowcolor{blue!20}
imgseg      & 0.772          & \textbf{0.824} & 0.820                        & 0.796          & 0.787          \\
\rowcolor{red!20}
lympho      & \textbf{1.000} & 0.989          & 0.987                        & \textbf{1.000} & \textbf{1.000} \\
\rowcolor{blue!20}
satellite   & 0.778          & 0.792          & \textbf{0.799}               & 0.796          & 0.790          \\
\rowcolor{blue!20}
satimage-2  & 0.919          & 0.931          & \textbf{0.933}               & \textbf{0.933} & 0.927          \\
\rowcolor{blue!20}
vertebral   & 0.138          & 0.143          & \textbf{0.144}               & 0.143          & 0.141          \\
\rowcolor{blue!20}
WPBC        & 0.394          & 0.408          & \textbf{0.411}               & 0.402          & 0.401          \\
\rowcolor{red!20}
musk        & 1.000          & 0.997          & 0.996                        & \textbf{1.000} & \textbf{1.000} \\
\rowcolor{red!20}
cadio       & 0.863          & 0.824          & 0.826                        & 0.847          & \textbf{0.864} \\
\rowcolor{red!20}
WDBC        & 0.983          & 0.980          & 0.981                        & 0.983          & \textbf{0.989} \\
\rowcolor{blue!20}
mammography & 0.417          & \textbf{0.579} & 0.575                        & 0.563          & 0.561          \\
\rowcolor{blue!20}
pima        & 0.701          & \textbf{0.751} & 0.749                        & 0.740          & 0.734          \\
\bottomrule
\toprule
\multirow{2}{*}{\centering Datasets}    & Base detector & \multirow{2}{*}{\centering +LLM-DAS} & \multicolumn{3}{c}{+LLM-DAS (Hardness Level)} \\ \cline{4-6}
            & IForest     &         & High  & Medium   & Low    \\ \midrule
\rowcolor{blue!20}
imgseg      & 0.756          & \textbf{0.859} & 0.854                        & 0.843          & 0.847          \\
\rowcolor{blue!20}
lympho      & 0.959          & \textbf{0.979} & 0.976                        & 0.976          & 0.968          \\
\rowcolor{red!20}
satellite   & 0.858          & 0.808          & 0.811                        & 0.849          & \textbf{0.862} \\
\rowcolor{red!20}
satimage-2  & 0.885          & 0.871          & 0.870                        & 0.892          & \textbf{0.906} \\
\rowcolor{blue!20}
vertebral   & 0.134          & \textbf{0.269} & 0.249                        & 0.237          & 0.231          \\
\rowcolor{blue!20}
WPBC        & 0.376          & \textbf{0.399} & 0.396                        & 0.386          & 0.384          \\
\rowcolor{blue!20}
musk        & 0.528          & \textbf{1.000} & \textbf{1.000}               & \textbf{1.000} & \textbf{1.000} \\
\rowcolor{red!20}
cadio       & 0.702          & 0.622          & 0.624                        & 0.687          & \textbf{0.718} \\
\rowcolor{blue!20}
WDBC        & 0.975          & 0.994          & \textbf{0.997}               & 0.991          & 0.983          \\
\rowcolor{blue!20}
mammography & 0.333          & \textbf{0.571} & \textbf{0.571}               & 0.565          & 0.554          \\
\rowcolor{blue!20}
pima        & 0.666          & 0.740          & \textbf{0.747}               & 0.728          & 0.721          \\
\bottomrule
\end{tabular}
\end{table}

\begin{table}[t!]
\centering
\scriptsize
\caption{
Wasserstein distance metrics for quantifying anomaly hardness.
$d(\mathcal{A}, \mathcal{N})$ is the multivariate Wasserstein distance between anomalies $\mathcal{A}$ and normal samples $\mathcal{N}$ via optimal-transport in the feature space.
Lower distance indicates higher hardness (more overlap with normal).
The ratio $d(\mathcal{A}_{\text{Syn}},\mathcal{N}) / d(\mathcal{A}_{\text{True}},\mathcal{N})$ is the \emph{Relative Hardness Ratio}.
}
\label{tab:hardness_wdist}
\resizebox{\linewidth}{!}{
\begin{tabular}{cccc|ccc}
\toprule
\multicolumn{7}{c}{Base detector: PCA} \\ \midrule
Datasets    & $d(\mathcal{A}_{\text{Syn}}, \mathcal{N})$ / $d(\mathcal{A}_{\text{True}}, \mathcal{N})$    &  $d(\mathcal{A}_{\text{True}}, \mathcal{N})$ & $d(\mathcal{A}_{\text{Syn}}, \mathcal{N})$      & $d(\mathcal{A}_{\text{Syn}_\text{High}}, \mathcal{N})$ & $d(\mathcal{A}_{\text{Syn}_\text{Medium}}, \mathcal{N})$ & $d(\mathcal{A}_{\text{Syn}_\text{Low}}, \mathcal{N})$  \\ \midrule
\rowcolor{blue!20}
imgseg      & 1.870 & 4.963         & 9.279  & 9.148              & 10.534            & 11.953             \\
\rowcolor{red!20}
lympho      & 0.540 & 17.529        & 9.474  & 9.504              & 18.017            & 58.429             \\
\rowcolor{blue!20}
satellite   & 1.431 & 7.828         & 11.198 & 11.078             & 11.386            & 13.125             \\
\rowcolor{blue!20}
satimage-2  & 0.457 & 14.716        & 6.728  & 6.734              & 7.194             & 8.066              \\
\rowcolor{blue!20}
vertebral   & 8.141 & 1.691         & 13.765 & 13.895             & 14.911            & 16.732             \\
\rowcolor{blue!20}
WPBC        & 0.915 & 5.407         & 4.947  & 4.962              & 5.237             & 5.875              \\
\rowcolor{red!20}
musk        & 0.261 & 31.207        & 8.157  & 8.179              & 9.249             & 12.112             \\
\rowcolor{red!20}
cardio      & 0.724 & 12.054        & 8.726  & 8.736              & 9.088             & 9.769              \\
\rowcolor{red!20}
WDBC        & 0.322 & 16.559        & 5.340  & 5.724              & 8.706             & 9.159              \\
\rowcolor{blue!20}
mammography & 1.424 & 5.439         & 7.743  & 7.746              & 7.903             & 8.813              \\
\rowcolor{blue!20}
pima        & 5.944 & 2.376         & 14.124 & 14.173             & 17.924            & 20.202             \\
\bottomrule
\toprule
\multicolumn{7}{c}{Base detector: IForest} \\ \midrule
Datasets    & $d(\mathcal{A}_{\text{Syn}}, \mathcal{N})$ / $d(\mathcal{A}_{\text{True}}, \mathcal{N})$    &  $d(\mathcal{A}_{\text{True}}, \mathcal{N})$ & $d(\mathcal{A}_{\text{Syn}}, \mathcal{N})$      & $d(\mathcal{A}_{\text{Syn}_\text{High}}, \mathcal{N})$ & $d(\mathcal{A}_{\text{Syn}_\text{Medium}}, \mathcal{N})$ & $d(\mathcal{A}_{\text{Syn}_\text{Low}}, \mathcal{N})$  \\ \midrule
\rowcolor{blue!20}
imgseg      & 1.050 & 4.963         & 5.213  & 5.214              & 5.678             & 6.168              \\
\rowcolor{blue!20}
lympho      & 0.224 & 17.529        & 3.920  & 3.919              & 4.050             & 4.263              \\
\rowcolor{red!20}
satellite   & 0.772 & 7.828         & 6.042  & 6.056              & 6.773             & 7.922              \\
\rowcolor{red!20}
satimage-2  & 0.222 & 14.716        & 3.262  & 3.258              & 4.910             & 7.228              \\
\rowcolor{blue!20}
vertebral   & 6.440 & 1.691         & 10.889 & 10.768             & 10.910            & 11.729             \\
\rowcolor{blue!20}
WPBC        & 0.685 & 5.407         & 3.705  & 3.718              & 4.057             & 4.763              \\
\rowcolor{blue!20}
musk        & 0.125 & 31.207        & 3.906  & 3.917              & 4.917             & 5.196              \\
\rowcolor{red!20}
cardio      & 0.379 & 12.054        & 4.573  & 4.588              & 7.112             & 8.878              \\
\rowcolor{blue!20}
WDBC        & 0.174 & 16.559        & 2.888  & 2.919              & 3.071             & 4.471              \\
\rowcolor{blue!20}
mammography & 1.518 & 5.439         & 8.257  & 8.276              & 8.377             & 8.658              \\
\rowcolor{blue!20}
pima        & 3.557 & 2.376         & 8.453  & 8.438              & 8.950             & 9.052              \\
\bottomrule
\end{tabular}}
\end{table}

\paragraph{Quantifying hardness via Wasserstein distance.}
To make hardness evaluation model-agnostic, we compute the multivariate Wasserstein distance between the synthetic anomalies 
and the normal set using an optimal-transport formulation EMD (Earth Mover's Distance), where the two 
point clouds are treated as empirical distributions in the feature space.
Unlike the classical 1D Wasserstein distance computed over scalar score 
distributions, our implementation uses the full multivariate point sets 
(normal vs. synthetic anomalies) and measures their geometric discrepancy 
through an optimal transport cost.

We denote $d(\mathcal{A}, \mathcal{N})$ as the multivariate Wasserstein distance between anomalies $\mathcal{A}$ and normal samples $\mathcal{N}$ via optimal-transport in the feature space.
A lower value of $d(\mathcal{A},\mathcal{N})$ indicates that anomalies lie closer to
the normal manifold and are therefore \emph{harder} to detect.

Table~\ref{tab:hardness_wdist} reports:
\begin{itemize}
    \item $d(\mathcal{A}_{\text{True}},\mathcal{N})$: the inherent difficulty of the dataset
    (how far true anomalies are from normal samples),
    \item $d(\mathcal{A}_{\text{Syn}},\mathcal{N})$: distance for anomalies synthesized by the original LLM-DAS,
    \item $d(\mathcal{A}_{\text{Syn}_{\text{High}}},\mathcal{N})$,
          $d(\mathcal{A}_{\text{Syn}_{\text{Medium}}},\mathcal{N})$,
          $d(\mathcal{A}_{\text{Syn}_{\text{Low}}},\mathcal{N})$:
    distances under each hardness level,
    \item the ratio 
    $\frac{d(\mathcal{A}_{\text{Syn}},\mathcal{N})}{d(\mathcal{A}_{\text{True}},\mathcal{N})}$,
    which we refer to as the \emph{Relative Hardness Ratio}.
\end{itemize}

Two observations are particularly important:
\begin{itemize}
    \item \textbf{Monotonic trend across hardness levels.}
    For both PCA and IForest, $d(\mathcal{A}_{\text{Syn}_{\text{High}}},\mathcal{N})$
    is consistently smaller than or comparable to the Medium and Low settings,
    and the distances increase from High $\to$ Medium $\to$ Low.
    This confirms that the hardness control interface is effective:
    High hardness produces anomalies closer to normal data, while Low hardness produces
    more easily separable anomalies.
    \item \textbf{Failure cases correspond to ``overly hard'' synthetic anomalies.}
    In the failure cases highlighted in red in Table~\ref{tab:hardness_perf}, 
    the ratio $\frac{d(\mathcal{A}_{\text{Syn}},\mathcal{N})}{d(\mathcal{A}_{\text{True}},\mathcal{N})}$ in Table~\ref{tab:hardness_wdist}
    tends to be comparatively small.
    This means that the synthesized anomalies are much closer to normal samples than the true anomalies,
    even though the true anomalies are already far from normal
    (large $d(\mathcal{A}_{\text{True}},\mathcal{N})$).
    In these failure cases, the problem is inherently easy, and generating very hard synthetic anomalies
    can over-complicate the learned boundary and hurt performance.
\end{itemize}

\paragraph{Practical takeaway.}
Taken together, Table~\ref{tab:hardness_perf} and Table~\ref{tab:hardness_wdist} show that:
\begin{itemize}
    \item LLM-DAS can synthesize anomalies at controllable hardness levels;
    \item the original implementation naturally produces hard anomalies (small Wasserstein distance);
    \item rare side effects arise when the dataset is inherently easy but the synthetic anomalies are ``too hard'';
    \item these side effects can be mitigated in practice by lowering the hardness level,
    which increases $d(\mathcal{A}_{\text{Syn}},\mathcal{N})$ and restores performance to baseline or better.
\end{itemize}

\textbf{Prompt used in LLM-DAS variant with controlled hardness.}
To further analyze the impact of anomalous samples with varying difficulty levels, we modified the prompt in the ``Requirements'' section by introducing a variable, the difficulty coefficient ``degree'', enabling LLMs to generate anomalies with varying degrees of difficulty. The specific prompt is as follows. 

\begin{tcolorbox}[promptstyle, title={Example Prompt for Generating Varying Difficulty Synthesis Code}, label={prompt:varying_difficulty}]
You are an expert in anomaly detection systems. The training set contains only normal samples. We use a IForest detector, where the anomaly score is computed using model.predict\_score(). The higher the score, the more anomalous the sample.  

\textbf{The description of IForest.}

*XXX.

*XXX.

*XXX.

...(Same as original prompt)

\textbf{Your task} is ...

*XXX.

*XXX.

*XXX.

...(Same as original prompt)

\textbf{Requirements:}
Your should strictly follow below requirements:

...
    
  Thus the function format is generate\_hard\_anomalies(n\_samples: int, model, X\_train: np.ndarray, degree) \\[5pt]
  \textbf{Supplementary explanation:}  The parameter degree is used to generate anomalies with varying levels of detection difficulty, featuring three optional values: \\
  - \texttt{low} (low difficulty)  \\
  - \texttt{med} (medium difficulty) \\
  - \texttt{high} (high difficulty) \\
   These correspond to generating anomalies that are hard, moderately hard, and extremely hard to detect, respectively. You need to generate anomalies of the corresponding difficulty level based on the specific value of degree.

Return only the complete Python function generate\_hard\_anomalies(...), with policy you used for genenrating anomalies and clear comments explaining key steps.
\end{tcolorbox}

\subsection{Analysis of Generating Detector-Specific Binary Classifiers}
\label{appendix:model_direct}
To further explore the potential of LLMs within the proposed framework, we redesigned the prompts in the ``Objective'' and ``Requirements'' sections to enable LLMs to develop detector-specific binary classifiers. These classifiers are then directly integrated with the original ones to enhance the anomaly detection performance of the baseline models. The specific prompt can be found in prompt~\ref{prompt:binary_direct}.

\begin{tcolorbox}[promptstyle, title={Example Prompt for Generating Detector-Specific Binary Classifiers}, label={prompt:binary_direct}]
You are an expert in anomaly detection systems. The training set contains only normal samples. We use a IForest detector, where the anomaly score is computed using model.predict\_score(). The higher the score, the more anomalous the sample.  

\textbf{The description of IForest.}

*XXX.

*XXX.

*XXX.

...(Same as original prompt)

\textbf{Your task} is to write a Python function detector\_specific\_binary\_classifiers(...) tailored for PCA detectors. This function must complement PCA by focusing on capturing the aforementioned anomaly types that PCA fails to detect. Once the function is implemented, users will be able to provide it with the following inputs:

* A trained PCA model (model) that exposes predict\_score(),

* The training samples (X\_train)

\textbf{Requirements}:
Your should strictly follow below requirements:

1. You must use your expertise to design a binary classifier that is specifically tailored for PCA detectors, not a model-agnostic classifier. The design should leverage the inherent characteristics and limitations of PCA to achieve complementarity.

2. The classifier should focus on capturing anomaly types that PCA fails to detect, particularly those that result in low scores from model.predict\_score() (i.e., anomalies with low reconstruction error in PCA). To achieve this, you can first identify "borderline" normal training samples based on professional understanding of PCA's inherent characteristics (not solely relying on anomaly scores), then design the classifier to be sensitive to transformations of these samples that PCA cannot detect. The classifier's decision logic must be specific to PCA's weaknesses, making it ineffective as a general classifier for other detectors.

3. For the PCA model, you can only use the function model.predict\_score() to interact with it.

4. The classifier must be trained on the provided training data and should expose standard fit(X, y) and predict\_score(X) methods.

5. The function should allow setting:

    * the trained PCA model (model),
    
    * training samples (X\_train).

    Thus the function format is detector\_specific\_binary\_classifiers(model, X\_train: np.ndarray)

6. All package imports must be done inside the function.
    
Return only the complete Python function detector\_specific\_binary\_classifiers(...), with the policy used for designing the classifier and clear comments explaining key steps.

\end{tcolorbox}

\subsection{Analysis of Generating Synthetic Samples Targeting Multiple Weaknesses of Base Detector}
\label{appendix:multiple_weekness}

To explore whether LLMs can identify multiple weaknesses of detectors and whether a mixture of synthetic anomalies targeting all these weaknesses enhances robustness, we modified the prompt in the ``Objective'' and ``Requirements'' sections by instructing LLMs to first identify several distinct logical weaknesses of the detector and then generate a mixture of synthetic anomalies targeting each weakness. The specific prompt can be found in prompt~\ref{prompt:m_w}. And the example of the generation policy generated by the LLMs is available in Code\ref{code:IForest_w}. The results are provided in Table~\ref{appendix:table_multiple_weakness}.

\begin{table}[h!]
\centering
\caption{Performance comparison between the original LLM-DAS and the multi-weakness variant LLM-DAS (M-W), 
which explicitly prompts the LLM to identify and target several challenging anomaly patterns per detector.
}
\label{appendix:table_multiple_weakness}
\scriptsize
\begin{tabular}{cccc|ccc}
\toprule
            & PCA   & +LLM-DAS (Ours) & +LLM-DAS (M-W)      & IForest    & +LLM-DAS (Ours) & +LLM-DAS (M-W)       \\ \midrule
abalone     & 0.839 & 0.839   & 0.848 & 0.848 & 0.855   & 0.850  \\
annthyroid  & 0.566 & 0.660   & 0.582 & 0.615 & 0.620   & 0.623  \\
imgseg      & 0.772 & 0.824   & 0.799 & 0.756 & 0.859   & 0.822  \\
lympho      & 1.000 & 0.989   & 1.000 & 0.959 & 0.979   & 0.972  \\
mammography & 0.417 & 0.579   & 0.431 & 0.333 & 0.571   & 0.453  \\
pendigits   & 0.386 & 0.485   & 0.397 & 0.513 & 0.602   & 0.613  \\
pima        & 0.701 & 0.751   & 0.749 & 0.666 & 0.740   & 0.758  \\
satellite   & 0.778 & 0.792   & 0.794 & 0.858 & 0.808   & 0.867  \\
satimage-2  & 0.919 & 0.931   & 0.951 & 0.885 & 0.871   & 0.950  \\
shuttle     & 0.963 & 0.991   & 0.991 & 0.917 & 0.995   & 0.995  \\
vertebral   & 0.138 & 0.143   & 0.141 & 0.134 & 0.269   & 0.138  \\
vowels      & 0.105 & 0.109   & 0.228 & 0.098 & 0.178   & 0.190  \\
wilt        & 0.064 & 0.097   & 0.090 & 0.085 & 0.278   & 0.088  \\
WPBC        & 0.394 & 0.408   & 0.400 & 0.376 & 0.399   & 0.405  \\
breastw     & 0.993 & 0.996   & 0.994 & 0.945 & 0.995   & 0.994  \\
cardio      & 0.863 & 0.824   & 0.873 & 0.702 & 0.622   & 0.834  \\
WDBC        & 0.983 & 0.980   & 0.982 & 0.975 & 0.994   & 0.980  \\ \midrule
Average         & 0.640 & 0.670   & 0.662 & 0.627 & 0.684   & 0.678  \\
\bottomrule
\end{tabular}
\end{table}

\begin{tcolorbox}[promptstyle, title={Example Prompt  for Generating Synthetic Samples Targeting Multiple Weaknesses}, label={prompt:m_w}]
You are an expert in anomaly detection systems. The training set contains only normal samples. We use a IForest detector, where the anomaly score is computed using model.predict\_score(). The higher the score, the more anomalous the sample.  

\textbf{The description of IForest.}

*XXX.

*XXX.

*XXX.

...(Same as original prompt)

\textbf{Your task} is to write a Python function generate\_hard\_anomalies(...) that generates anomalies which are the most difficult for the IForest detector to detect. This means that you the generated anomalies should have relatively low anomaly score, thus they are hard to be detected. But these anomalies are helpful to build a more robust detector. \textbf{Critically, you must first identify multiple distinct logical weaknesses of IForest (not just one), then design targeted anomaly generation policies for each weakness. The final output should be a mixture of synthetic anomalies targeting all identified weaknesses to maximize robustness.} After the function is completed, users can provide it with:

* A trained IForest model (model) that exposes predict\_score(),

* The training samples (X\_train)

\textbf{Requirements}:
Your should strictly follow below requirements:

1. \textbf{You must first identify more than one distinct, logical weaknesses of IForest.} For each weakness, clearly explain the logical basis (how it arises from IForest's design) in the function’s comments.

2. \textbf{For each identified weakness, design a unique, IForest-specific anomaly generation policy (not model-agnostic).} For each policy , you can first find the set of ‘borderline’ normal training samples based on your unique and professional understanding to IForest, not only based on the anomaly score. Then transform them to anomalies that is tailor-designed for IForest. Please note that the transformation function should be specific for IForest, which means that it is not a general transformation for other detectors.  

*XXX.

*XXX.

*XXX.

...(Same as original prompt)
\end{tcolorbox}

\definecolor{lightgreen}{RGB}{184, 219, 179}  
\definecolor{darkgreen}{RGB}{114, 176, 99}  
\definecolor{grayblue}{RGB}{113, 154, 172}  
\definecolor{darkblue}{RGB}{62, 96, 141}  
\definecolor{blue2}{RGB}{68, 133, 199}  
\definecolor{gray2}{RGB}{120,120,120}  
\renewcommand{\lstlistingname}{Code}
\lstset{
    basicstyle = \fontfamily{qcr}\selectfont\small, 
     keywordstyle = \color{magenta}, 
    commentstyle = \color{darkgreen},
    stringstyle = \color{gray2},   
    numbers = left,                  
    numberstyle=\tiny\color{red},
    firstnumber=1,
    showspaces = false,                 
    showstringspaces = false,                 
    frame = lrtb,   
    breaklines = true,                  
    columns = fixed,                 
    basewidth = 0.55em,                
    escapeinside = {},                    
    lineskip=0pt,
}

\lstdefinestyle{Python}{
    language = Python,
    basicstyle = \fontfamily{qcr}\selectfont\small, 
    numberstyle = \tiny\color{gray},
    breaklines = true,
}

\label{appendix:generated_multi_weekness}
\begin{lstlisting}[
    style = Python,
    caption = {Generated Code for Synthesizing Anomalies Targeting Multiple Weaknesses of IForest}, 
    label = {code:IForest_w}
]
import numpy as np

def generate_hard_anomalies(n_samples: int, model, X_train: np.ndarray):
    """
    Generates anomalies that are difficult for an Isolation Forest (IForest) model to detect.

    This function operates by first identifying three distinct logical weaknesses of the IForest
    algorithm, then implementing a targeted generation policy for each weakness. The final
    output is a mixture of synthetic anomalies from all policies.

    Args:
        n_samples (int): The total number of hard anomalies to generate.
        model: A trained IForest-like model object that exposes a `predict_score` method.
               The method should return a score where a higher value indicates a higher
               likelihood of being an anomaly.
        X_train (np.ndarray): The training data used to train the model, containing only
                              normal samples. Shape (n_train_samples, n_features).

    Returns:
        np.ndarray: An array of generated hard anomalies, with shape (n_samples, n_features).
    """
    # 1. Package Imports
    import numpy as np

    # 2. Pre-computation and Helper Functions
    if n_samples == 0:
        return np.array([])
        
    n_train, d = X_train.shape
    if d == 0:
        return np.zeros((n_samples, 0))

    # The user-provided `model.predict_score()` is assumed to return higher scores for
    # more anomalous samples. We use these scores to stratify the training data into
    # topological regions as perceived by the IForest model itself.
    train_scores = model.predict_score(X_train)

    # Define quantiles to identify different types of normal points. This is a nuanced
    # way to select 'borderline' samples based on the model's own perception of the data.
    # - Central points: Deep inside the data cloud (low scores).
    # - Borderline points: On the fringes of dense regions (medium scores).
    # - Edge points: Most isolated of the normal points (high scores for a normal sample).
    score_quantiles = np.percentile(train_scores, [25, 75, 95])
    central_mask = train_scores <= score_quantiles[0]
    borderline_mask = (train_scores > score_quantiles[0]) & (train_scores <= score_quantiles[1])
    edge_mask = train_scores >= score_quantiles[2]

    # Ensure we have points in each category to draw from.
    X_central = X_train[central_mask] if np.any(central_mask) else X_train
    X_borderline = X_train[borderline_mask] if np.any(borderline_mask) else X_train
    X_edge = X_train[edge_mask] if np.any(edge_mask) else X_train

    # Global statistics for transformations
    data_min = np.min(X_train, axis=0)
    data_max = np.max(X_train, axis=0)
    data_range = data_max - data_min
    data_std = np.std(X_train, axis=0)
    data_centroid = np.mean(X_train, axis=0)
    
    # Avoid division by zero for constant features
    data_range[data_range == 0] = 1.0
    data_std[data_std == 0] = 1.0

    # 3. IForest Weaknesses and Corresponding Anomaly Generation Policies

    generated_anomalies = []
    
    # Allocate samples to each policy
    n_policy1 = n_samples // 3
    n_policy2 = n_samples // 3
    n_policy3 = n_samples - n_policy1 - n_policy2
    policy_allocations = [n_policy1, n_policy2, n_policy3]

    # ---
    # Policy 1: Clustered Anomalies
    # ---
    # Weakness ID: "Masking" or "Crowding" Effect.
    # Logical Basis: IForest isolates points. If several anomalies are positioned closely
    # together, they form a dense cluster. The algorithm will then require many splits to
    # isolate any single point within this new cluster, making them appear "normal" (i.e.,
    # giving them longer path lengths and thus lower anomaly scores). This policy creates
    # such a dense micro-cluster in a sparse area of the feature space.
    if policy_allocations[0] > 0:
        # Step 1: Find a 'borderline' region by selecting a normal point from a sparse area.
        # We use 'edge' points as seeds because they are already in less dense regions,
        # making them good candidates for locating our new anomaly cluster far from the main data.
        seed_point = X_edge[np.random.randint(0, len(X_edge))]

        # Step 2: Define the center of the anomaly cluster by pushing the seed point
        # further away from the data's centroid.
        cluster_center = data_centroid + 2.0 * (seed_point - data_centroid)

        # Step 3: Transform by generating a very tight cluster around this new center.
        # The small covariance ensures the points are very close, maximizing the crowding effect.
        cluster_cov = np.diag((data_std / 50.0)**2) # Very small variance
        anomalies_p1 = np.random.multivariate_normal(
            mean=cluster_center, cov=cluster_cov, size=policy_allocations[0]
        )
        generated_anomalies.append(anomalies_p1)

    # ---
    # Policy 2: Diagonal Anomalies
    # ---
    # Weakness ID: Inefficiency of Axis-Parallel Splits for Diagonal Patterns.
    # Logical Basis: IForest builds trees by splitting on a single feature at a time, creating
    # axis-parallel decision boundaries. It is inefficient at isolating anomalies that only
    # become apparent through a combination of features (e.g., lying on a diagonal).
    # Such points require many "stair-step" splits to be isolated, increasing their path
    # length and lowering their anomaly score.
    if policy_allocations[1] > 0:
        anomalies_p2 = []
        # Step 1: Select 'borderline' normal samples. These are points on the edge of the main
        # data cloud, making them good candidates to be pushed slightly further out in a
        # non-obvious, diagonal direction to become hard-to-detect anomalies.
        base_points_indices = np.random.randint(0, len(X_borderline), size=policy_allocations[1])
        base_points = X_borderline[base_points_indices]
        
        for point in base_points:
            anomaly = point.copy()
            # Step 2: IForest-specific transformation. Choose two random features and push the point's
            # values along both dimensions simultaneously, creating a diagonal shift.
            if d > 1:
                f1, f2 = np.random.choice(d, 2, replace=False)
                
                # Perturb by a magnitude related to the standard deviation of each feature
                perturbation1 = np.random.choice([-1, 1]) * 2.5 * data_std[f1]
                perturbation2 = np.random.choice([-1, 1]) * 2.5 * data_std[f2]
                
                anomaly[f1] += perturbation1
                anomaly[f2] += perturbation2
            else: # Handle 1D case gracefully
                anomaly[0] += np.random.choice([-1, 1]) * 2.5 * data_std[0]
            
            anomalies_p2.append(anomaly)
        generated_anomalies.append(np.array(anomalies_p2))

    # ---
    # Policy 3: Subspace Anomalies
    # ---
    # Weakness ID: High-Dimensional Feature Sparsity.
    # Logical Basis: In high-dimensional spaces, IForest randomly selects a feature to split on
    # at each node. If a point is anomalous only within a small subset of features (a subspace),
    # the algorithm is unlikely to select these specific features for splitting early on. It will
    # waste many splits on irrelevant dimensions where the point looks normal, again increasing
    # its path length and making it appear less anomalous.
    if policy_allocations[2] > 0:
        anomalies_p3 = []
        # Step 1: Select 'central' normal samples. These points have the lowest anomaly scores and
        # are deep within the normal data distribution. Modifying them subtly makes them hard to
        # distinguish from their many legitimate neighbors.
        base_points_indices = np.random.randint(0, len(X_central), size=policy_allocations[2])
        base_points = X_central[base_points_indices]

        # Step 2: Define a small subspace (a few features) to perturb.
        num_perturbed_dims = max(1, int(d * 0.1)) # Perturb ~10% of features

        for point in base_points:
            anomaly = point.copy()
            # Step 3: Apply the IForest-specific subspace transformation.
            perturbed_dims = np.random.choice(d, num_perturbed_dims, replace=False)
            for dim in perturbed_dims:
                # Push the value just outside the observed range of normal data for that feature.
                epsilon = 0.1 * data_range[dim]
                if np.random.rand() > 0.5:
                    anomaly[dim] = data_max[dim] + epsilon
                else:
                    anomaly[dim] = data_min[dim] - epsilon
            anomalies_p3.append(anomaly)
        generated_anomalies.append(np.array(anomalies_p3))

    # 4. Final Combination and Output
    if not generated_anomalies:
        # Fallback in case n_samples was small and all policies got 0
        return np.empty((0, d))

    final_anomalies = np.vstack(generated_anomalies)
    
    # Shuffle to mix anomalies from different policies
    np.random.shuffle(final_anomalies)

    return final_anomalies[:n_samples] # Ensure exact number is returned
\end{lstlisting} 

\subsection{How we ensure the robustness of LLM-DAS?}
LLM-DAS relies on the LLM’s 
ability to interpret a detector’s algorithmic description. However, the framework is 
not brittle and is intentionally designed to remain robust even when the LLM's 
analysis is imperfect.

\textbf{(1) The LLM analyzes a clean, parameter-free, logical description—not raw tabular data.}
The LLM’s task is restricted to summarizing the detector’s high-level mechanism 
and reasoning about its structural weaknesses. These descriptions (e.g., PCA steps) are symbolic and algorithmic rather than data-dependent, 
making them significantly easier and more reliable for LLMs to understand.

\textbf{(2) The prompt imposes strict structural and functional constraints.}
As detailed in Eq.3 of Section 3.2, the LLM is required to produce:
(i) a detector-aware synthesis policy,
(ii) an executable, data-agnostic program,
(iii) an explanation of its reasoning.
These constraints substantially reduce the risk of superficial reasoning by enforcing 
explicit, verifiable logic.
For example, in Case Study of Section 3.2.2, the LLM's policy for IForest describes two core steps:
(i) selecting borderline normal samples using the detector’s own score function, and 
(ii) performing controlled extrapolation that preserves long path lengths.  
The accompanying explanation clarifies why these operations align with the 
algorithmic weakness of IForest (i.e., difficulty isolating deep points), making the 
LLM’s intentions transparent and allowing us to verify that the generated code faithfully 
implements this logic.  
This level of specificity prevents arbitrary code generation.

\textbf{(3) The final detector does not rely solely on the LLM-generated component.}
After generating synthetic anomalies, we train an enhancement classifier $\tilde{f}_t$ and 
combine it with the original detector $f_t$ using a normalized score-level ensemble. 
This fusion (Eq.7 of Section 3.2.2) ensures that even if the LLM's synthesis is occasionally 
suboptimal, the original detector’s behavior remains preserved, preventing degradation.

\textbf{(4) Extensive empirical evidence demonstrates robustness across detectors, datasets, and LLMs.}
The results in Table 1 of manuscript indicate that the improvements over the original source detectors across 36 benchmark datasets are statistically significant.
The detailed results in Table 4 and Table 5 of Appendix 6.7 also show that the suboptimal results are occasional.
In addition, our experiments in Table 8 of Appendix 6.9 further verify that LLM-DAS remains effective across different LLMs, including GPT-4o, Gemini-2.5-Pro, and Qwen3. 

\textbf{(5) Boundary refinement after enhancement.}
Fig. 5(b) of manuscript shows that the synthetic anomalies lie close to the normal manifold and receive scores similar to borderline normal samples under the original detector $f_t$.  
This confirms that the generated samples match the detector-aware policy and are indeed “hard” for the original detector.
Crucially, Fig. 5(c) of manuscript demonstrates that after training with LLM-DAS, the enhanced detector exhibits a clear rightward shift of both synthetic and real anomaly score  distributions relative to normal data.  This tightening of the decision boundary shows that the synthesized anomalies  not only challenge the detector but meaningfully reshape the classifier to produce  better separation. This is direct evidence that the LLM-generated samples are of sufficiently high quality to improve the detector's generalization.

\subsection{On Isolating the Contribution of LLM Reasoning}

What knowledge is LLM actually contributed? Our claim is not that the LLM invents entirely new mathematics, but that it performs a contextual, detector-specific reasoning and synthesis task that goes far beyond retrieving a single, static rule.

\textbf{The Ablations Test Specific, Necessary Components:} Our ablation studies (Generic/Simple/Random in Fig. 4(b)) are designed to remove specific, instructed capabilities from the LLM:

\begin{itemize}
    \item ``Generic'': Removes detector-awareness. The LLM cannot perform reasoning conditioned on the target detector's algorithm.
    \item ``Simple'': Removes the capability to generate hard anomalies by denying access to the score function.
    \item ``Random'': Removes the borderline heuristic, a key strategic element.
\end{itemize}

The significant performance drop in each variant confirms that each of these instructed reasoning components is critical. This shows that the LLM is not just executing a single built-in rule but is dynamically combining these elements based on the prompt.

\textbf{The Cross-Detector Experiment is the Key Evidence (Connecting to Table 2):} The most compelling evidence against a ``universal rule'' hypothesis comes from our cross-detector experiment (Table 2).

\begin{itemize}
    \item If the LLM were simply applying a generic, detector-agnostic heuristic (e.g., ``add small noise'' or ``extrapolate''), then the synthesis policy \( \mathcal{S}_{\text{program}}^A \) for Detector A should work reasonably well when transferred to Detector B.
    \item {The results clearly show the opposite:} Transferred strategies often perform poorly. This demonstrates that each LLM-generated policy is highly specialized and its effectiveness is contingent on a precise match between the synthesis logic and the target detector's internal mechanics. This specialization is the hallmark of genuine, detector-specific reasoning.
\end{itemize}

In summary, the ablations prove the necessity of key reasoning components, and the cross-detector experiment proves the specificity of the resulting strategies. Together, they robustly isolate and validate the contribution of the LLM's contextual reasoning capabilities.

\end{document}